\documentclass{article}

\usepackage{multicol}

\usepackage{arxiv}
\usepackage{longtable}
\usepackage[utf8]{inputenc} 
\usepackage[T1]{fontenc}    
\usepackage{hyperref}       
\usepackage{url}            
\usepackage{booktabs}       
\usepackage{amsfonts}       
\usepackage{nicefrac}       
\usepackage{microtype}      
\usepackage{lipsum}		
\usepackage{graphicx}
\usepackage{natbib}
\usepackage{doi}
\usepackage{tabularx}
\usepackage{setspace}
\usepackage{booktabs}
\usepackage{ltablex} 
\title{{\hspace{1mm}Benchmarking human face similarity using identical twins}\thanks{The paper is Accepted in IET Biometrics Journal in 5th August 2022 and DOI: 10.1049/bme2.12090}}

\author{ \href{https://orcid.org/0000-0002-0110-0949}{\includegraphics[scale=0.01]{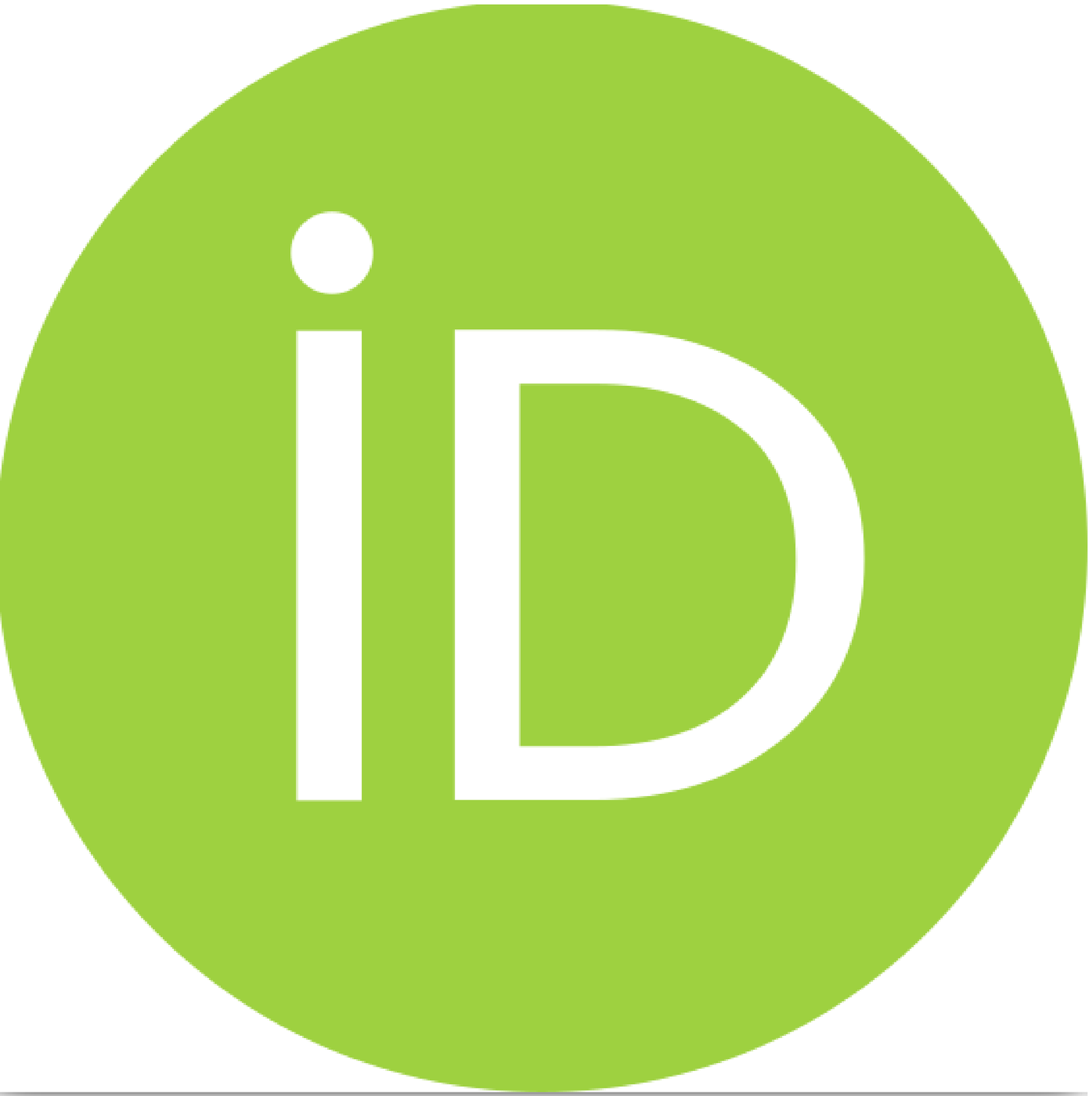}\hspace{1mm}Shoaib Meraj Sami}\\
	Lane Department of \\
	Computer Science and Electrical Engineering \\
	West Virginia University \\
	Morgantown WV, 26506 \\
	\texttt{sms00052@mix.wvu.edu} \\
	\And
	\href{https://orcid.org/0000-0002-2400-8619}{\includegraphics[scale=0.01]{orcid.eps}\hspace{1mm}John McCauley} \\
	Lane Department of\\
	Computer Science and Electrical Engineering \\
	West Virginia University \\
	Morgantown WV, 26506 \\
	\texttt{jamccauley@mix.wvu.edu} \\
	\And
	\href{https://orcid.org/0000-0003-3541-0918}{\includegraphics[scale=0.01]{orcid.eps}\hspace{1mm}Sobhan Soleymani} \\
	Lane Department of\\
	Computer Science and Electrical Engineering \\
	West Virginia University \\
	Morgantown WV, 26506 \\
	\texttt{ssoleyma@mix.wvu.edu} \\
	\And
	\href{https://orcid.org/0000-0001-8730-627X}{\includegraphics[scale=0.01]{orcid.eps}\hspace{1mm}Nasser Nasrabadi} \\
	Lane Department of\\
	Computer Science and Electrical Engineering \\
	West Virginia University \\
	Morgantown WV, 26506 \\
	\texttt{nasser.nasrabadi@mail.wvu.edu} \\
	\And
	\href{https://orcid.org/0000-0002-4539-7588}{\includegraphics[scale=0.01]{orcid.eps}\hspace{1mm}Jeremy Dawson}\thanks{Corresponding Author} \\
	Lane Department of\\
	Computer Science and Electrical Engineering \\
	West Virginia University \\
	Morgantown WV, 26506 \\
	\texttt{Jeremy.Dawson@mail.wvu.edu} \\
}

\renewcommand{\shorttitle}{\textit{IET} Biometrics}

\begin{document}
\maketitle

\begin{abstract}
The problem of distinguishing identical twins and non-twin look-alikes in automated facial recognition (FR) applications has become increasingly important with the widespread adoption of facial biometrics. Due to the high facial similarity of both identical twins and look-alikes, these face pairs represent the hardest cases presented to facial recognition tools. This work presents an application of one of the largest twin datasets compiled to date to address two FR challenges: 1) determining a baseline measure of facial similarity between identical twins and 2) applying this similarity measure to determine the impact of doppelgangers, or look-alikes, on FR performance for large face datasets. The facial similarity measure is determined via a deep convolutional neural network. This network is trained on a tailored verification task designed to encourage the network to group together highly similar face pairs in the embedding space and achieves a test AUC of 0.9799. The proposed network provides a quantitative similarity score for any two given faces and has been applied to large-scale face datasets to identify similar face pairs. An additional analysis which correlates the comparison score returned by a facial recognition tool and the similarity score returned by the proposed network has also been performed. 
\end{abstract}


\section{Introduction}
\subsection{Motivation }
\begin{figure}
	\centering
	\includegraphics{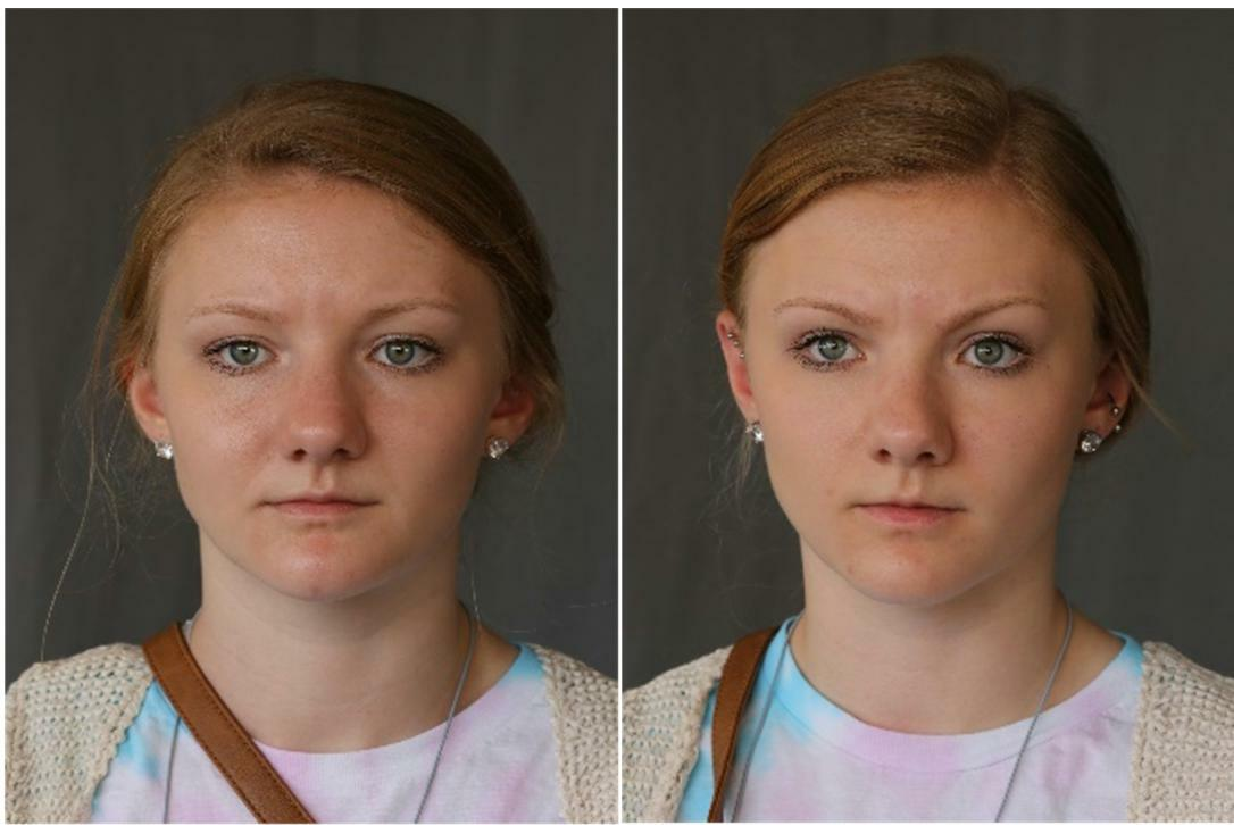}
	\caption{An example of an identical twin pair.}
		\label{fig:fig0}
		\includegraphics{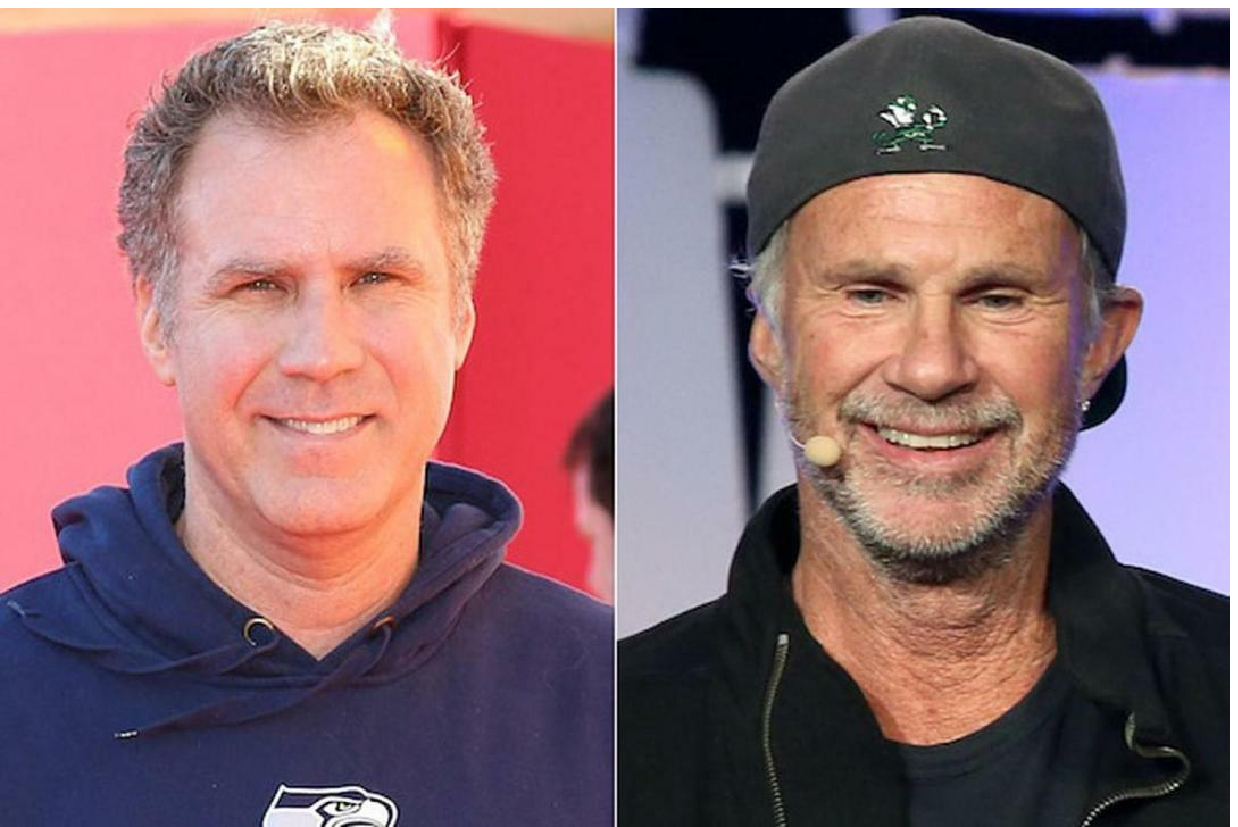}
	\caption{ Celebrity look-alikes, Will Ferrell and Chad Smith.}
	\label{fig:fig1}
\end{figure}

Identical or mono-zygotic twins pose an important and interesting problem to facial recognition (FR) systems. Due to the high level of facial similarity exhibited between identical twin pairs, these individuals are often mis-identified by automatic facial recognition systems, an example identical twin pair is shown in Figure \ref{fig:fig0}. Several studies have shown this to be case, with one study from \citep{Paone2014} showing that the average facial recognition system has a significantly higher equal error rate (EER) when presented with a population of identical twins versus a non-twin population, even under ideal imaging conditions. While identical twins present the worst case of facial similarity, this problem is also extended to non-related individuals with high facial similarity, often known as look-alikes or doppelgangers. A well-known celebrity look-alike pair, Will Ferrell and Chad Smith, is shown in Figure \ref{fig:fig1}. A study of look-alike recognition by \citep{Chen_2018-bt} found that even deep-learning based approaches to facial recognition have a difficult time accurately identifying look-alike pairs. Look-alike impacts on face recognition is especially relevant due to the rapid increase in the size of face datasets currently being seen. As these face datasets increase in size, they include a larger number of distinct identities, and with each added identity, the chance of encountering a look-alike pair within the dataset increases. Each of these problems represent some of the hardest cases presented to facial recognition systems. As such, these problems must be studied and understood to ensure the successful implementation of facial recognition technology in the future. 


The challenges caused by identical twins and look-alikes are rooted in the fact that the face comparisons in question happen between identities with high facial similarity. In both cases, two identities are matched against one another in a non-mated comparison in which a comparison or match score is returned for the pair. Due to the high facial similarity of the pair, errors and misidentifications are more likely to occur in these cases. Here, it becomes important to make a distinction between the comparison (or match) score generated by a face matcher and the facial similarity of the identities under comparison. In both cases, face images are being evaluated against one another to return some comparison result. In the case of the comparison score, the intent is to return an accurate identification of the individual(s) in the images in question, whereas the similarity measure seeks to quantify the facial similarity of the identities within the images. This distinction is important, as the comparison score returned by a facial recognition tool may not be directly related or determined by the facial similarity of the individuals. This work seeks to provide greater understanding of the relationship between facial similarity and the comparison score returned by a facial recognition tool. 

The research efforts presented in this work are important to facial recognition at large because they establish a worst-case baseline of facial similarity in all face comparisons from face images with known high similarity: those of identical twins. This result represents the most challenging case of non-mated face comparison in general and quantifies the average facial similarity of identical twin pairs. In addition to this baseline measurement, this work also has value in its method of determining facial similarity. This method utilizes one of the largest databases of identical twin images to train a deep convolutional neural network to quantify facial similarity. This measure has further importance in its use in determining highly similar faces in any face dataset. Identifying highly similar faces has several pertinent applications, two of which being the selection of appropriately similar faces for morphed face pair generation, and the evaluation of the difficulty of face datasets. Furthermore, this measure of facial similarity can provide key insights into the relationship between facial similarity and the comparison score returned by a facial recognition tool.

\section{Previous Work}
\subsection{Biometric Identification of Identical Twins}
Biometric identification of identical twins has been studied extensively in many different modalities. In one of the earliest works concerning the biometric identification of identical twins, \citep{Jain2002} used the fingerprints of twin pairs to accurately differentiate between the twin individuals. This study employed a dataset of 94 identical twin fingerprints and found that the tested fingerprint recognition tool saw only small performance decreases when presented with identical twins. In a study from  \citep{Sun2010}, multiple modalities of biometric identification were explored. This work examined three biometric modalities, fingerprint, iris, and face. This work found iris to perform the best out of the three modalities, with results for identical twins being just as good as those for non-twin individuals. Facial recognition was the worst performing of the three, showing significantly higher error rate for twins as compared to that of non-twin individuals. In the first study of ear recognition of identical twins, \citep{Nejati} . showed that ear recognition techniques can be applied successfully to identical twin datasets. This study developed a new ear recognition algorithm and applied the algorithm to the largest available dataset of identical twins at the time. Their results found recognition accuracy up to 92\% on the twin dataset, confirming that ear recognition is a viable tool for the biometric recognition of twins. Multiple surveys of identical twin recognition have confirmed the challenges, and successes, of twin recognition. A survey from  \citep{ref6_Ricanek2013} discusses the differences seen in identical twins as the twins age. This survey compiles research that indicates identical twins become easier to differentiate as they get older due to age-based changes in their biometric features. The conclusions of this survey indicate that experiments attempting to differentiate between two identical twins will perform better on older twin pairs due to epigenetic changes in the twins’ biometrics. In a later survey of the field,  \citep{Bowyer2016} discuss a multitude of biometric recognition techniques used to differentiate twins. They highlight the difficulty of facial recognition of twins through multiple studies, discuss the minor accuracy degradation seen in twin fingerprint recognition as compared to non-twin fingerprint recognition, and show the success of several works exploring iris recognition of identical twins. The survey also discusses the need for larger and more robust datasets of identical twins, as well as the need for further research in speaker recognition and handwriting recognition of identical twins. 

\subsection{Facial Recognition of Identical Twins}

Facial recognition is quickly becoming one of the most widely used biometric modalities, due to its high acceptance among the general public, ease of data capture, and accuracy. This wide scale implementation makes it more important than ever to evaluate the hardest cases presented to facial recognition systems, such as identical twin pairs. One of the earliest studies on the face recognition of twins employed the Notre Dame/West Virginia University Twins Days dataset, and found that COTS face matchers could identify twins when imaging conditions were ideal (i.e., studio lighting, neutral expression), but performance was measurably decreased when the test images were captured under ambient lighting or had expression variation \citep{Phillips2011}. A second study analyzing the performance of FR algorithms using the Twins Days dataset showed that significant work is still needed to improve the performance of face recognition of identical twins \citep{Pruitt2011}. This study tested the performance of both COTS and academic matchers on the twin dataset, and found that, under ideal imaging conditions, twins could generally be identified well, with one matcher obtaining a EER of only 0.01. However, when testing these matchers under non-ideal imaging conditions including expression variation, lighting variation, and facial occlusion from glasses, all the matchers saw significant performance decreases. A similar study from  \citep{Paone2014} further examined the performance of several face recognition algorithms on the Twins Days dataset. This work examined the effects of illumination variation, expression variation, age, and gender on twin face recognition, for images taken on the same day as well as cross year matches in which data from one year was used as the gallery, and the other year the probe. This study produced similar results to others \citep{Phillips2011, Pruitt2011} in that, under ideal imaging conditions, twins can be recognized, but imaging non-idealities significantly reduce performance. The worst results in this study were found in the cross-year experiments in which every algorithm performed worse than when presented with images taken on the same day. These experiments showed notable difficulty in differentiating between identical twins using images taken in two separate years than those taken on the same day. This is a particularly important result because, in many real-world FR applications, a face image could potentially be used as a gallery image for the recognition of an identity for several years. 

Multiple works have also explored using low-level facial features to identify identical twins. In an early work from \citep{Klare2011}, authors identify three levels of facial features that, when fused, can improve the facial recognition performance of identical twins. The work extracted multi-scale local binary patterns, scale-invariant feature transform features, and manually annotated facial marks from identical twin face images which were broken into face components (eyes, eyebrows, mouth, and nose). The results of the work indicate that the extraction and fusion of these features and face components led to better recognition performance of identical twins. A later study of component-based facial recognition of identical twins from \citep{ref_11_Mahalingam2013} confirmed the results of previous work by again showing that the fusion of facial components of identical twins improves facial recognition performance. Additionally, this study analyzed the impact of age and gender-based differences on facial recognition. Using the framework developed in the study, the authors found that gender made no significant impact to recognition performance, while age showed a positive impact on performance as the twin pairs age. 

\subsection{Facial Recognition of Look-alikes} 

Another of the most challenging cases presented to facial recognition systems is that of look-alike faces. An early study from \citep{Kosmerlj2005} on the robustness of a facial recognition tool to the occurrence of look-alikes in the general public found that the tested tool would not be robust to look-alikes. This study also reported an estimate of the frequency of look-alikes in the general public via the number of false accepts returned for each subject in the two tested datasets (labeled as dataset II and dataset III in the work). In the experiments performed the authors found that 97 \% and 99.99\% of the subjects in datasets II and III respectively had one or more false accepts at a false accept rate of 1\%. The experiments were repeated at a false accept rate of 0.1\% and found that dataset II showed almost no subjects with one or more false accepts, but dataset III showed 92\% of subjects having one or more false accepts. The authors admit that this evaluation may overestimate the true occurrence of look-alikes due to the overall performance of the facial recognition tool used in the study. In a preliminary study of look-alike facial recognition \citep{Lamba2011}, researchers developed a dataset of carefully constructed look-alike pairs. The results of this work showed that several facial recognition tools that were state-of-the-art at the time of publication performed poorly on this look-alike dataset. In a second study of look-alike recognition using the dataset constructed in \citep{Lamba2011}, \citep{Sun2010} showed that even neural-network-based FR approaches struggle with look-alike recognition. This work developed a deep Siamese convolutional neural network for twin recognition and found that both twins and look-alikes continue to pose significant challenges to the most advanced facial recognition techniques. A very recent work from \citep{Rathgeb2021} confirmed the conclusions of previous work in this sphere by showing the negative effect of look-alikes on even the most advanced facial recognition tools. This work developed a handpicked dataset of tailored look-alike pairs and tested the effectiveness of several facial recognition tools on this dataset. The results of this study show that four of the current state of the art facial recognition tools have very high imposter attack presentation match rate (IAPMR) when presented with a dataset containing look-alikes. This led the authors to the conclusion that these tools would not be robust to look-alike occurrences in general. 

\subsection{Facial Similarity}

A final related topic is that of facial similarity. Previous works have identified facial similarity as a topic that is distinctly different from facial recognition. Wherein facial recognition is used to differentiate between distinct identities, facial similarity is designed to calculate the similarity of the faces of two identities. One of the earliest works on facial similarity \citep{ref_15_Ramanathan15} laid the foundation of similarity as distinct from facial recognition, and developed a methodology to determine facial similarity using an eigenfaces framework. This study investigated facial similarity in a dataset of face images with illumination variation, occlusion, and pose variation, and ultimately used the similarity score returned by the eigenfaces tool to rank the most similar face images to a probe image via the generated similarity score. A more recent work from \citep{Sadovnik_2018_CVPR_Workshops} further developed the concept of facial similarity, and trained a deep neural network to rank similar faces in large scale datasets. This work used a dataset of human-chosen similar face images to train a deep neural network to return high similarity scores for visually similar faces via triplet loss. This work also studied the difference between facial similarity and facial recognition by testing the developed similarity network as a facial recognition tool. The authors found that their look-alike network performed worse at facial recognition than the VGG network the look-alike network was based on. The authors used this fact as evidence that their look-alike network had indeed been trained to complete a distinct task, that of determining facial similarity. Additionally, the determination of facial similarity has been shown to have practical applications in other use cases. A recent work from \citep{Rottcher2020} used the concept of facial similarity to find appropriate look-alike pairs for the generation of morphed faces. This work employed a variety of distinct features to determine the similarity of the faces used in the study and proved that their intelligently selected morphed faces out-performed randomly selected morphed pairs. This work is an extended version of \citep{McCauley2021}, which explored the concept of facial similarity as applied to identical twin pairs. This work has been expanded to include a more robust literature review and additional experimentation which explores the relationship between the comparison score returned by a facial recognition tool and the similarity score returned by the proposed similarity network. These experiments further support the claim that determining facial similarity is a distinct task from that of facial recognition and provide evidence that perceived facial similarity is a factor in the determination of comparison score but may not be the chief factor in the determination of such a score. These results highlight the need for further exploration of the concept of facial similarity and its relationship to the comparison score returned by any facial recognition tool. A review of previous works relating to biometric identification of identical twins, facial recognition of identical twins, facial recognition of look-a-like and facial similarity are summarized in Tables  \ref{tab:1} \ref{tab:2} \ref{tab:3} \ref{tab:4}.

\begin{tabularx}{\linewidth}{ c  X  X  X  X }

    \caption{Biometric Identification of Identical Twins}\label{tab:1}\\
    \hline
    Author’s Name & Feature Extraction & Algorithm & Datasets & Findings \\
    \\ \hline

    1. \citep{Le2015} & Aging features from nine region of the face image. & Local Fisher Discriminant Analysis (LFDA) & Notre Dame (ND) twin database & Identification rates of LFDA + Aging Features is approximately 10\% more accurate than LFDA. \\
    
    2.\citep{Nejati} & Exception Report Model (ERM) that prioritizes abnormal features & KNN classifierusing Mahalanobis distance & 39 pairs of identical twins from Sixth Mojiang International Twins Festival, China, 2010 &Recognition accuracy more than 90\%. \\

    3. \citep{mohammed2019overview} & ------ & PCA, LDA, Local Binary Patterns (LBP) &  Twin datasets (1,902 pairs), AR and FERET databases. & LBP outperforms PCA and LDA. \\
    
    4. \citep{Sun2010} & Multimodalusing fingerprints, iris and facial images. & Commercial matcher (FaceVACS) & CASIA Multimodal Biometrics Database of Twins (66 pairs) collected in the Fourth Annual Festival of Beijing Twins Day & Commercial face recognition system encounters a major challenge with considering iris matcher. To identify the identical twins, face based identification method EER is 13.18 \% more than two iris based fusion method. \\ 
    
    5. \citep{Juefei-Xu_2013_CVPR_Workshops} & Facial asymmetry features & Augmented Linear Discriminant Analysis & Notre Dame TWINS (2009-2010 database) & By using asymmetry features the identical twin verification rate is 48.50\% at 0.1\% false accept rate. \\
    
    6. \citep{Le2012} & Dense facial landmarks & Local Fisher Discriminant Analysis & Notre Dame TWINS (2009-2010 database) & By using aging features the identical twin verification rate is 48.50\% at 0.1\% false accept rate. \\
    
    7. \citep{Li2012} & Normal Local Binary Pattern, Normalized Gabor Filter and Normalized Local Gabor Binary Pattern & N-LBPs, N-GFs and N-LGBPs are fused at feature and score level then classify by Sparse representation classifier. &  Twins Days 2010 dataset Twinsburg, Ohio & Feature level fusion provided better recognition rate (96.73\%) than score level fusion. \\ 
    
    8. \citep{Priya2017} & Face, Fingerprint and Lip print based multi-modal verification & Kernel similarity with Euclidean distance for face matching, Possibilistic Fuzzy C-means clustering (PFCM) for fingerprint, fixed K-means Clustering features for lip print & In house virtual multi-modal database of 429 pairs of images & Matching score for this multimodal approach is more than 93\%. \\
    
    9. \citep{Revathi2020} & Mel frequency Cepstrum (MFCC),  and Linear frequency cepstral coefficients (LFCC), perceptual feature into speech signal  & Minimum distance classifier and maximum log likelihood classifier & Speech database & 6.75\% EER and 96\% recognition accuracy. \\
    
    10. \citep{Toygar2019} & Local binary patterns (LBP), Local phase quantization (LPQ), and Binarized statistical image (BSIF) & k-NN classifier & ND-TWINS-2009-2010 dataset & For LBP and BSIF based recognition, rates are 100\% and 99.45\%, and equal error rates are 0.54\% and 1.63\%, respectively. \\
    
    11.  \citep{Phillips2011} & ------- & Multiple Biometric Evaluation (MBE) 2010 that is a commercial FRS. & Twins Days festival in Twinsburg, Ohio 2009 and 2010 & The EER of this commercial face recognition system is 0.12 to 0.21 (same day with ambient light and neutral expression). \\
    
    12. \citep{ref_27_Mahalingam2013} & Hierarchical three patch local binary patterns (H-3P-LBP), LBP and histogram of oriented gradients (HOG) & Similarity score (Euclidean distance between feature vector from the gallery and the probe) and score level fusion.  & Notre Dame / WVU dataset, CASIA Twins Face Dataset & Score level fusion outperformed in the face recognition system. \\
    
    13. \citep{pourZonoozi2019} & Multi-biometric normalized raw features and false rejection rate features that are obtained from genuine and imposter distributions. & Fusion of SVM, Gaussian Mixture Model, and discriminant linear combination (DLC).  & Iran University of Science and Technology (IUST) Twin database. & Sum of raw and FRR features provides better performance than raw feature. \\
    
    14. \citep{Vengatesan2019} & Gray Level Co-occurrence Matrix (GLCM) filters & SVM classifier & Dataset from Twins Day Festival.  & Precision is 79.82\%  \\
    
    15. \citep{Pruitt2011} & Local Region PCA & PittPatt, VeriLook, Cognitec & Twins Days festival in Twinsburg, Ohio 2009 and 2010 & In terms of EER metrics the Cognitec performance is better than others. 
    \\ \hline
\end{tabularx}

\begin{tabularx}{\linewidth}{ c  X  X  X  X }

    \caption{Facial Recognition of Identical Twins}\label{tab:2}\\
    \hline
    Author’s Name & Feature Extraction & Algorithm & Datasets & Findings \\
    \hline

    1. \citep{Athmajan2015} & PCA & Improved Euclidean Distance, Cosine similarity, Cross Correlation and Dynamic Time Warping (DTW). Fusion of those method. & Extended Yale B database of 20 subjects. Each subject consists of 20 images. & Fusion of single classifiers performs better than single methods. \\
    
    2.  \citep{Klare2011} & Multi-scale local binary patterns (MLBP), scale invariant feature transform (SIFT), and facial marks feature & Linear discriminant analysis (LDA).&Twins Days festival in Twinsburg, Ohio 2009 & By including facial marks features, true acceptance rate increase significantly (13\%). \\
    
    3.\citep{Paone2014} &---------- & Two biometric evaluations and four commercial algorithms. & Twinsburg, Ohio in 2009 and 2010 & Illumination, expression, gender, and age affect the performance. EER range of 4.1\% to 17.4\%. \\

    \hline
\end{tabularx}

\begin{tabularx}{\linewidth}{ c  X  X  X  X }

    \caption{Facial Recognition of Look-a-Like.}\label{tab:3}\\
    \hline
    Author’s Name & Feature Extraction & Algorithm & Datasets & Findings \\
    \hline
    
     1. \citep{Rathgeb2021} & Deep convolutional neural network for feature extraction & Support Vector Machine (SVM) with a Radial Basis Function (RBF) kernel HDA & Doppelgänger and Look-Alike Face databases & Detection equal error rate is 2.7\%. \\
    
    2. \citep{Lamba2011} & Dynamic feed-forward neural network architecture based 2D-log polar Gabor transform & SVM & AR, CMU–PIE, Notre Dame, Equinox Databases & Accuracy of 86.7\% that is a 3.6\% improvement over existing algorithms. \\
    3.

    [\citep{quarkcs} &

    --------- &

    Eigenfaces, Fischer’s Linear Discriminant Analysis (LDA), and Active Appearance Models (AAM) &

    IMM+IMDB dataset, Celebrity dataset &

    AAM+LDA based recognition is superior to AAM and AAM + Pmap. \\
    
    4.

   \citep{Sadovnik_2018_CVPR_Workshops} &

    -------- &

    Triplet loss based VGG-Face CNN  &

    Color-Feret, LFW &

    Look-alike network (easy and hard triplet) performs 3.16\% better than VGG face in term of accuracy. \\
    
    5.

    \citep{Khodabakhsh} &

    OpenFace to extract the landmark positions &

    CNN &

    VoxCeleb2 &

    This algorithm achieves 7.93\% EER \\
    \hline
\end{tabularx}

\begin{tabularx}{\linewidth}{ c  X  X  X }

    \caption{Facial Similarity.}\label{tab:4}\\
    \hline
    Author’s Name & Algorithm & Datasets & Findings \\
    \hline
    
    1.

    \citep{ref_33_Ramanathan_2004} &

    Half-Faces based Eigenfaces framework &

    AR Face database &

    The effect of facial similarity by varying illumination, disguise, and aging. Half-face approach is an effective tool for different illumination conditions of faces. \\
    
    2.

    \citep{Sadovnik_2018_CVPR_Workshops} &

    VGG-Face Network and triplet loss &

    Names100 and LFW datasets &

    The authors found that CNN based facial recognition system can be used for facial similarity measurement. \\
    
    3.

    \citep{Rottcher2020} &

    ArcFace and a commercial off-the-shelf software &

    FRGC-v2 &

    The facial similarity network can be used for morph detection and identity detection.  \\
    
    4.

     \citep{McCauley2021} &

    FaceNet matcher (Inception-ResNet v1 architecture based) &

    Twin Dataset and CelebA &

    This network is effective tools for similarity measurement for the twin and look-alike faces. \\
        \hline
\end{tabularx}

\subsection{Goals and Objectives}

The primary goal of the research effort presented here is to quantify the facial similarity of identical twins in general. This similarity measurement represents the worst-case baseline of facial similarity in any non-mated comparison in facial recognition. In addition to this baseline measurement, a method of quantitatively determining facial similarity must be developed in order to determine the worst-case baseline measurement. Finally, match experimentation will be carried out to demonstrate the effect of highly similar faces on both commercial off the shelf and academic machine learning based facial recognition tools and identify potential look-alike identities. 

The tasks required to successfully complete the goals laid out in this work are as follows. First, face matching experimentation must be carried out to demonstrate the effect of identical twin faces on two facial recognition tools. This task will evaluate both identical twin datasets and large-scale non-twin datasets to build upon previous work’s evaluation of the effect of identical twin and look-alike faces and identify potential look-alike pairs in the large-scale non-twin datasets. Second, a method of quantitatively determining the facial similarity of any two faces must be developed. This measure should be based on the facial similarity of identical twins, as twins represent the worst-case scenario of facial similarity in facial recognition. This measure must directly compare two faces and return a quantitative measure of how similar the faces are (i.e., a similarity score), not a comparison score between the faces. This measure will be based on a deep convolutional neural network designed to directly compare two face images. Third, the similarity measure must be evaluated using both twin and non-twin datasets. This evaluation will determine the worst-case baseline measure of facial similarity, identify highly similar faces within the non-twin datasets, and serve as a point of comparison between the similarity score and comparison score returned by facial recognition tools. Additionally, this evaluation will provide an estimate of look-alike identities in a large-scale non-twin face dataset.

\section{Materials and Methods}
\subsection{Dataset descriptions}

The face image datasets used in this work come from multiple sources. The most important of these datasets is the twin dataset. This dataset compiles face data from WVU twins days collections from 2010 – 2019. In each of these collections face images were captured at the yearly Twins Days festival using high resolution DSLR imagery. The face images were captured using the SAP 50/51 standard five pose face capture method \citep{bryson2011american}, with even illumination in front of a neutral gray backdrop. This dataset contains 2,269 total identities, with 1,438 of these being identical twin pairs. The rest of the identities in the dataset come from fraternal twin pairs, family members of the twin pairs, and unrelated single individuals. The second dataset used, the non-twin dataset, comes from WVU biometric data collections from 2008-2019 in which ground truth face images were captured. Again, these images were captured using the SAP 50/51 five pose face capture method, with even face illumination and neutral gray backdrop. This dataset contains 5,295 identities, mostly students attending WVU at the time of collection. The final dataset, the large scale non-twin dataset, comes from combining the non-twin dataset with the publicly available CelebA dataset \citep{Liu_2015_ICCV}. The CelebA dataset contains images of celebrities scraped from the internet and contains 10,160 identities in total. Images from the CelebA dataset are unconstrained face images taken in-the-wild. The number of samples in these three datasets are summarized in Table  \ref{tab:5}. 

For the first two datasets, the twin dataset and non-twin dataset, the images were captured using Canon DSLR cameras. Three different cameras were used throughout the collection of the datasets: the Canon 5D Mark II, the Canon 5D Mark III, and the Canon 5DSR. All the images were captured using even illumination, either using 5500K tungsten three-point lighting, or 5500K LED light panels. Additionally, all the images were captured with the subject positioned in front of a neutral gray backdrop. 

The demographics of the twin and non-twin datasets are shown below in Figures \ref{fig:fig2} \ref{fig:fig3} \ref{fig:fig4} \ref{fig:fig5} \ref{fig:fig6} \ref{fig:}, demographics are not published for the CelebA dataset and are not included here. As shown in the tables, the demographics for both the twin dataset and non-twin dataset are skewed towards Caucasian/white participants. This is a factor of the populations from which biometric data collection participants were selected. 

Pre-processing for the twin and non-twin datasets involved manual cropping of the face images. These images were cropped to SAP 50 resolution and position standards in three steps. First, the manual selection of facial landmarks (center of eye, tip of nose, and outside edge of the ear) was carried out. Next, image rotation was performed to center and align the facial landmarks to be horizontal in the final cropped image. Finally, the aligned images were cropped around the face to the correct resolution standard. The CelebA dataset was cropped and aligned in a similar method, but performed using a neural network approach known as the Multi-task Cascaded Neural Network \citep{Zhang2016}. This face cropping approach uses neural network architectures to detect the face locations within the images, rotate and align the face, and crop the face from the image based on the detected face position. The CelebA images were cropped to a size of 182x182 pixels, with a tight margin around the central region of the face. Flow diagrams for all pre-processing steps are shown in Figures \ref{fig:fig8}\ref{fig:fig9}\ref{fig:fig10}.

\begin{tabularx}{\linewidth}{ c  X  X }

    \caption{Datasets Description.}\label{tab:5}\\
    \hline
    &Dataset Name &

    Number of Identities \\
    \hline

    1 &

    Twin Dataset (WVU Twin 2010-2019)&

    2269 \\
    
    2 &

    Non-twin Dataset (WVU biometric) &

    5295 \\
    
    3 &

    CelebA &

    10160 \\
\end{tabularx}

\begin{figure}

    \includegraphics[width=\columnwidth]{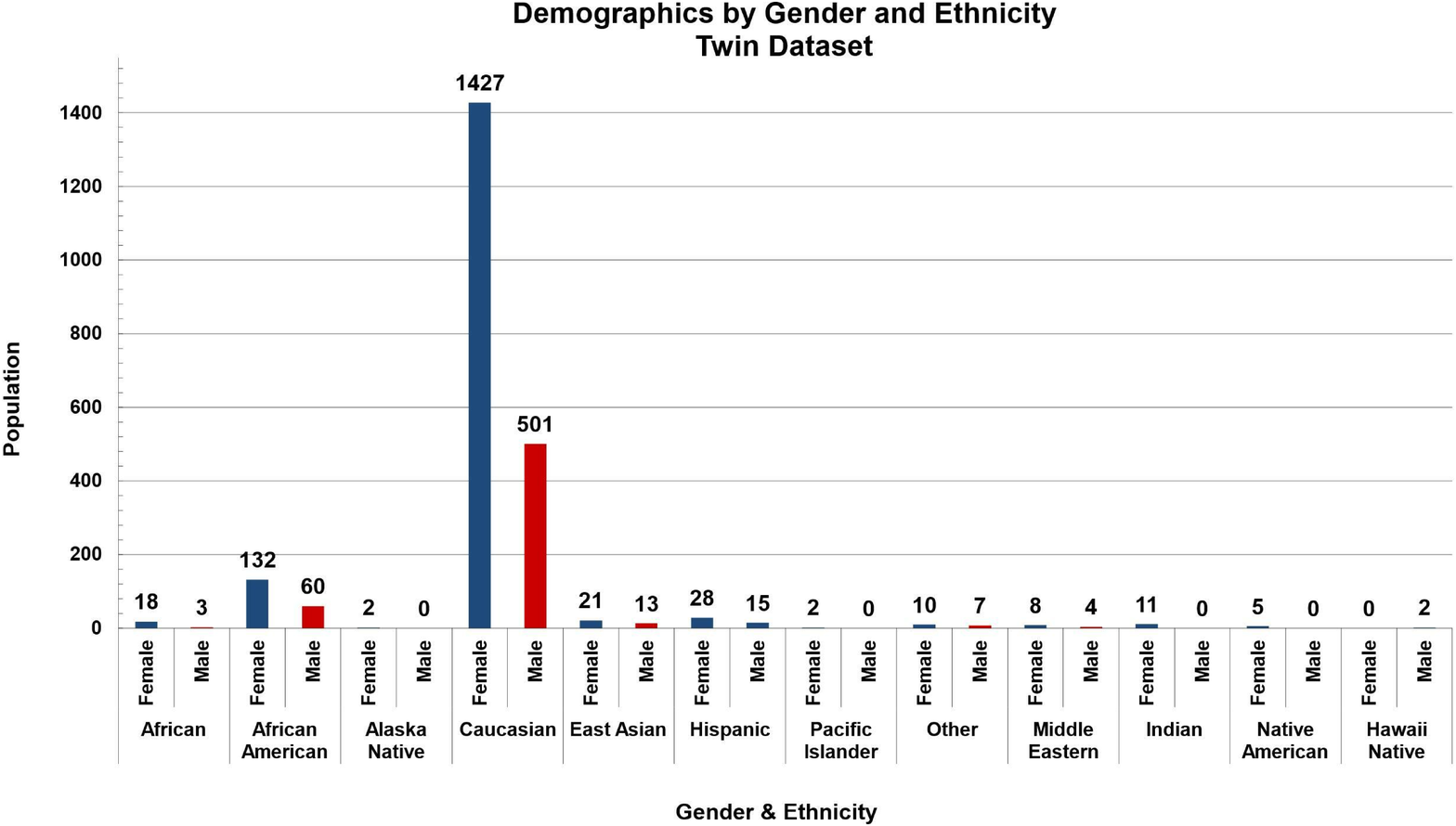}
    \caption{Twin dataset gender and ethnicity demographics chart. }
    	\label{fig:fig2}
    \includegraphics[width=\columnwidth]{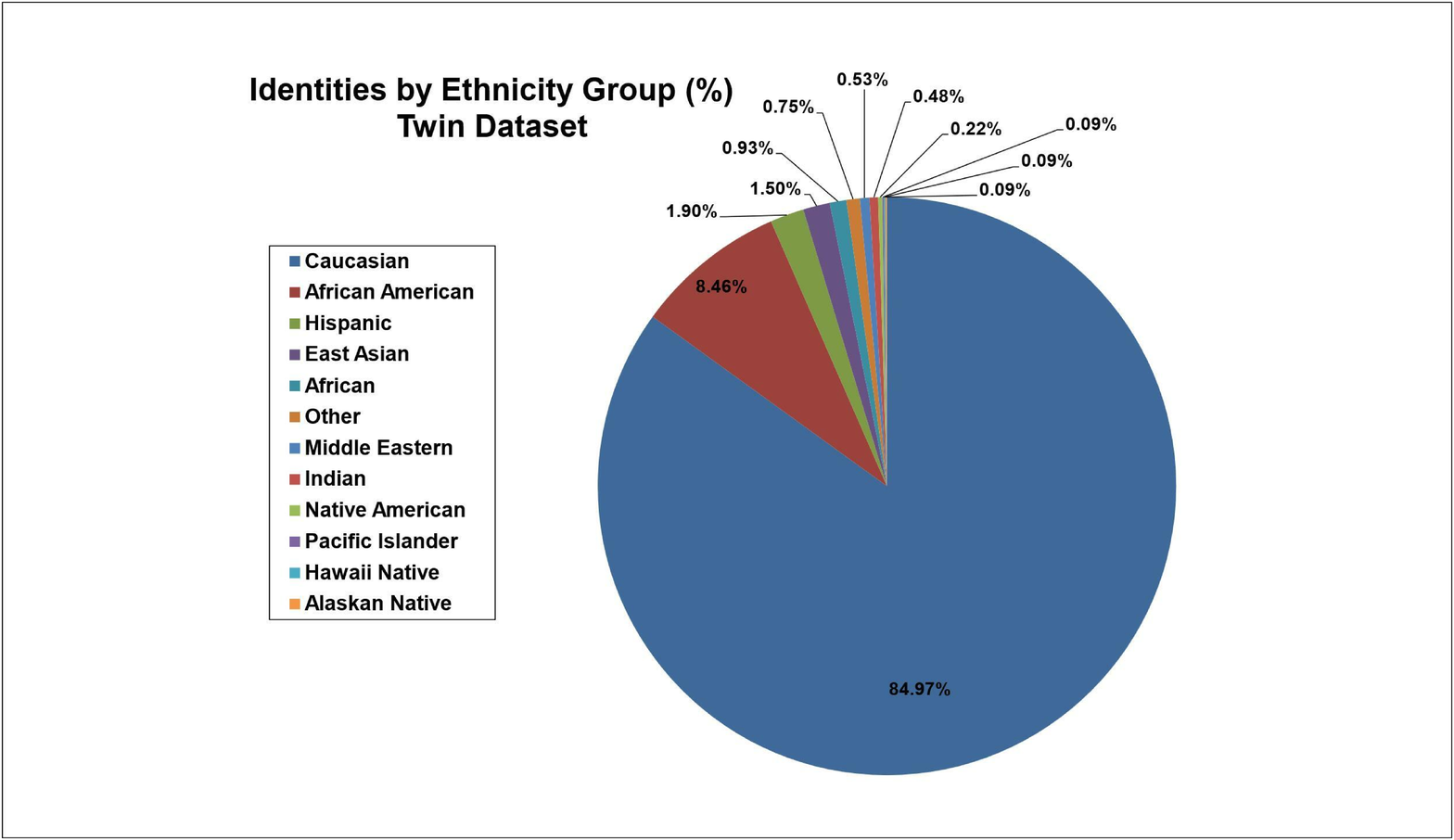}
    \caption{Twin dataset ethnicity demographics chart. }
    \label{fig:fig3}
\end{figure}
\begin{figure}
    \centering
    \includegraphics[width=\columnwidth]{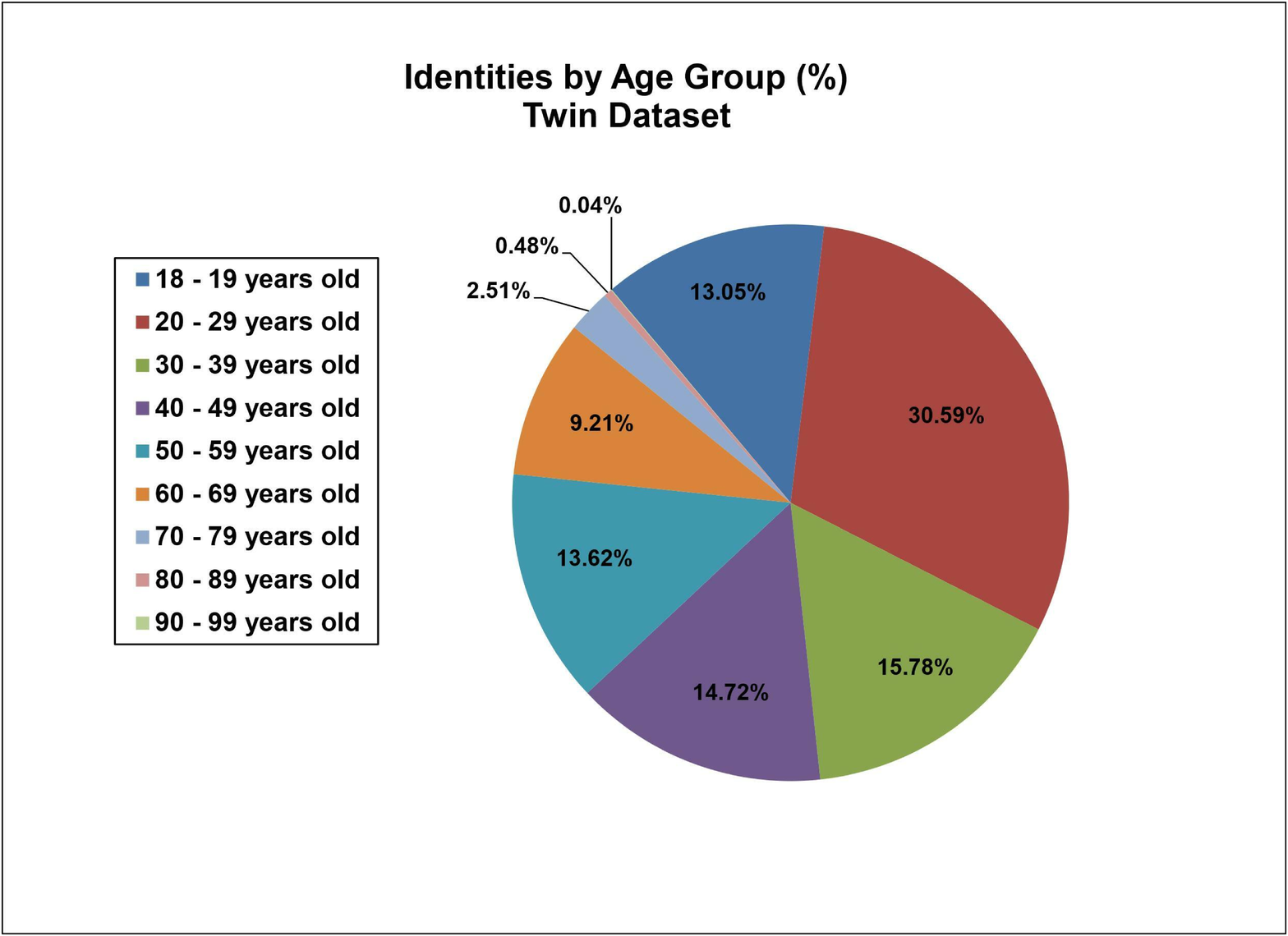}
    \caption{Twin dataset age group demographics chart.}
    \label{fig:fig4}
    \includegraphics[width=\columnwidth]{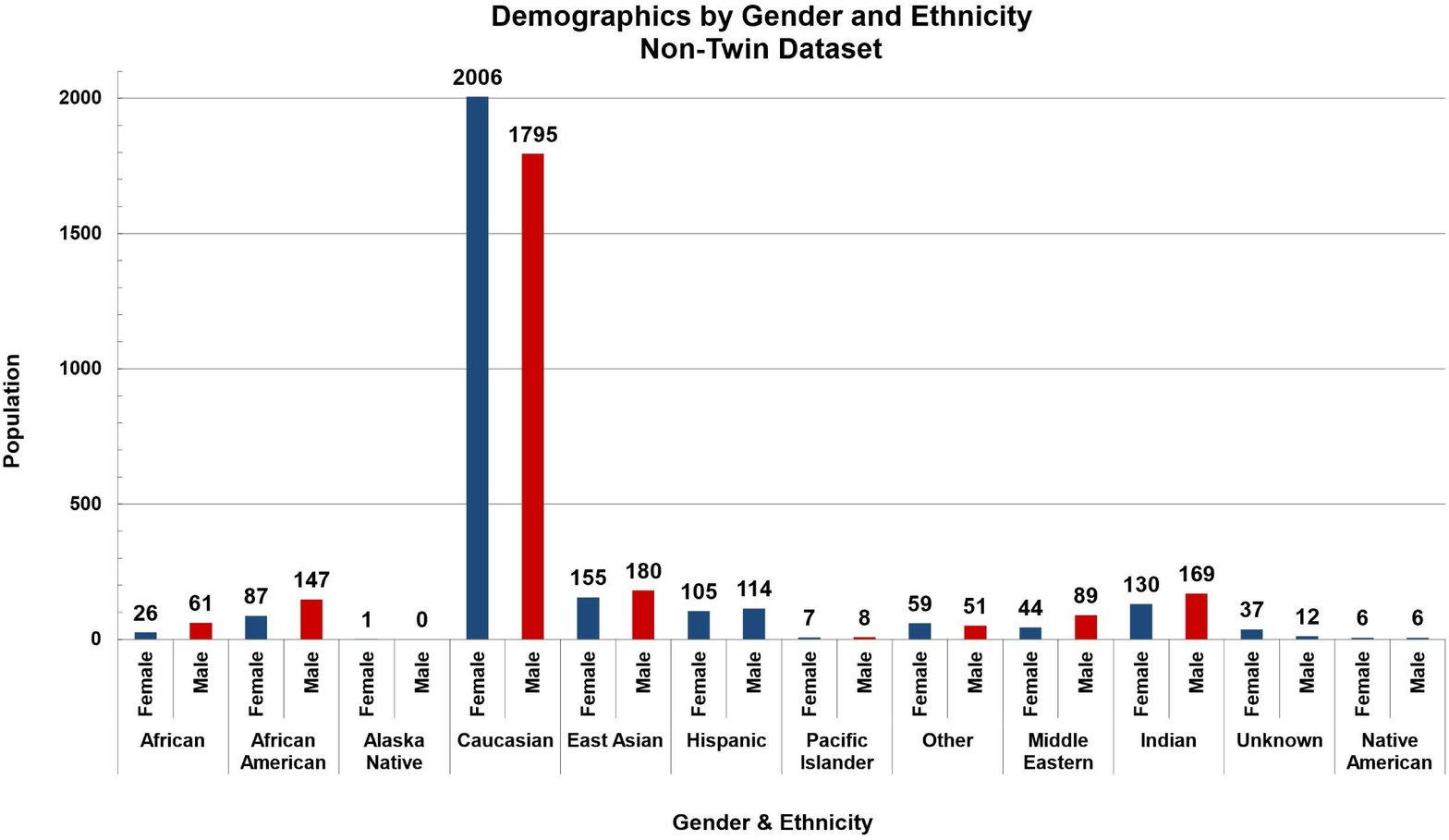}
    \caption{Non-twin dataset gender and ethnicity demographics chart. }
    \label{fig:fig5}
\end{figure}
\begin{figure}
    \centering
    \includegraphics[width=\columnwidth]{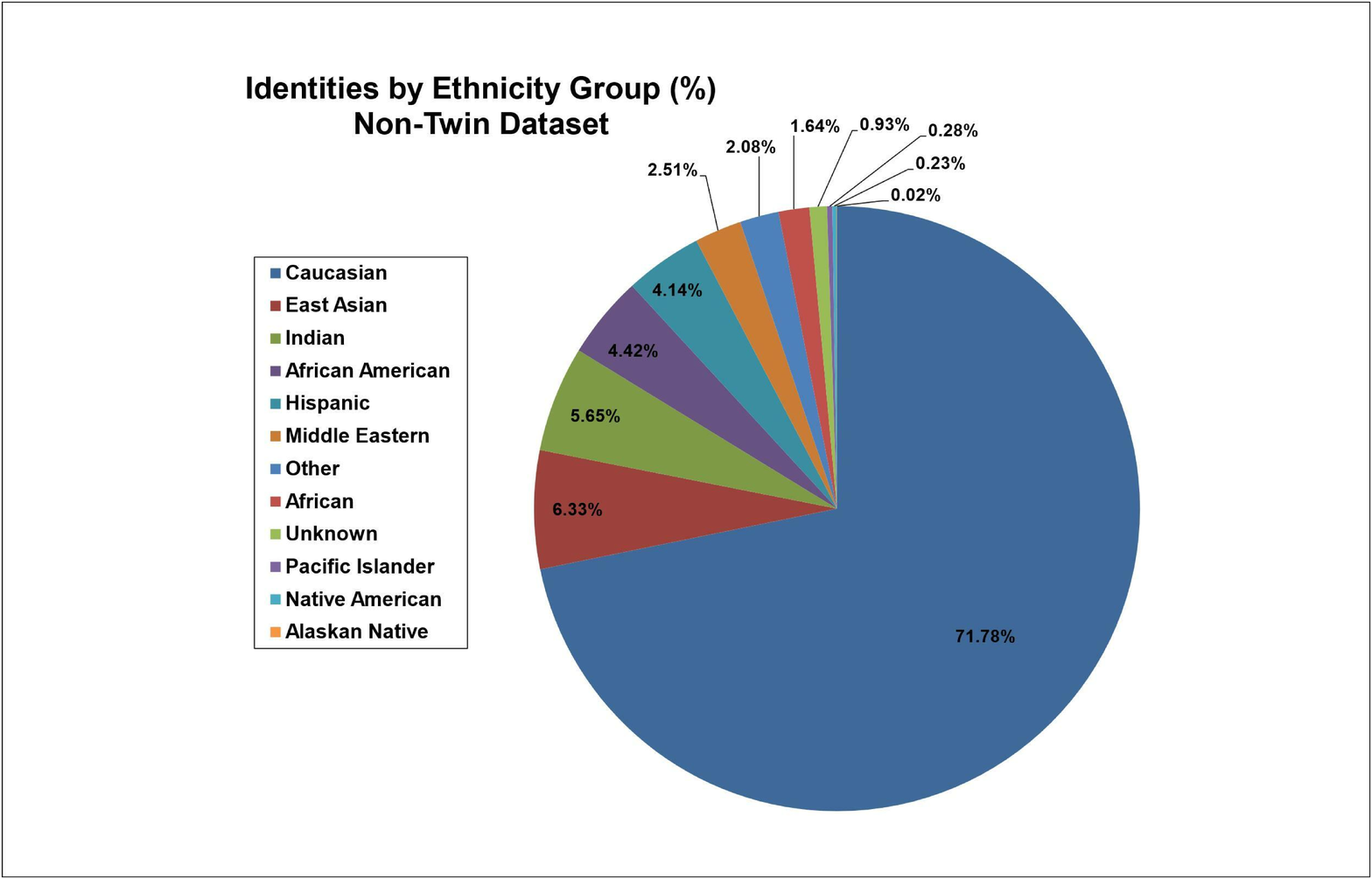}
    \caption{Non-twin dataset ethnicity demographics chart. }
    \label{fig:fig6}
    \includegraphics[width=\columnwidth]{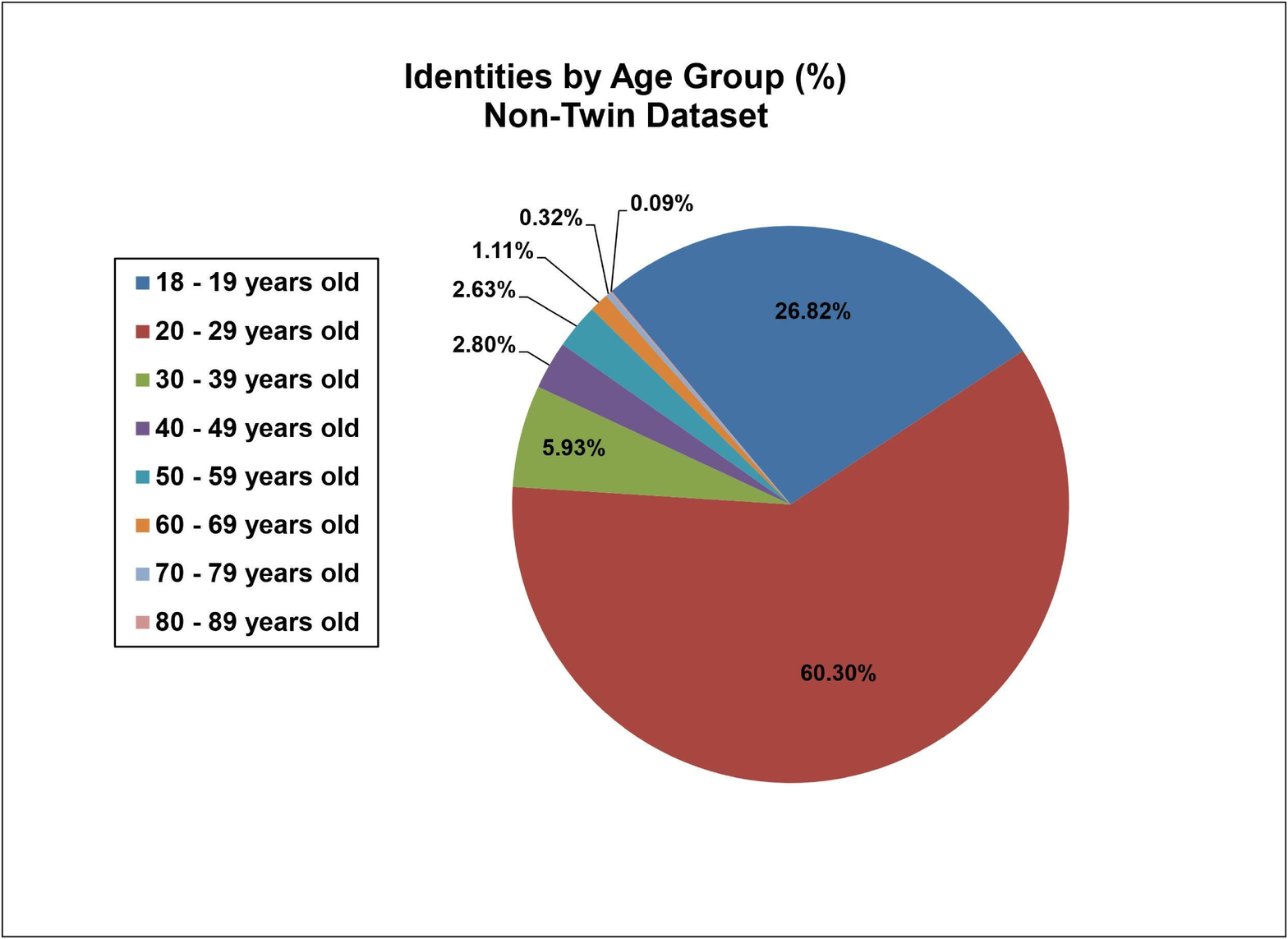}
    \caption{ Non-twin dataset ethnicity demographics chart.}
    \label{fig:}
\end{figure}

\begin{figure*}
\begin{multicols}{2}
\centering
    \includegraphics[width=\linewidth]{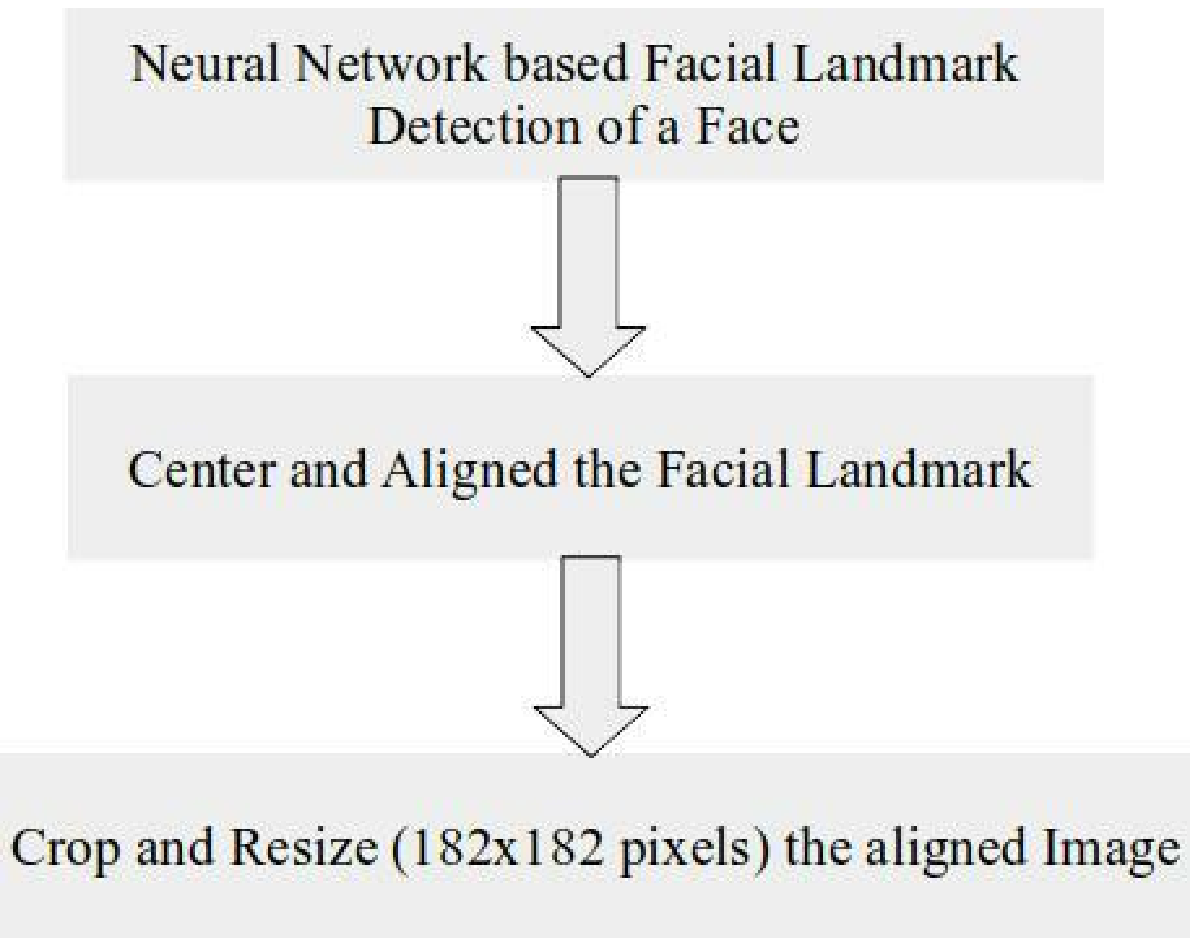}
    \caption{MTCNN-based Image Pre-Preprocessing.} 
    \label{fig:fig8}
    \includegraphics[width=0.6\linewidth]{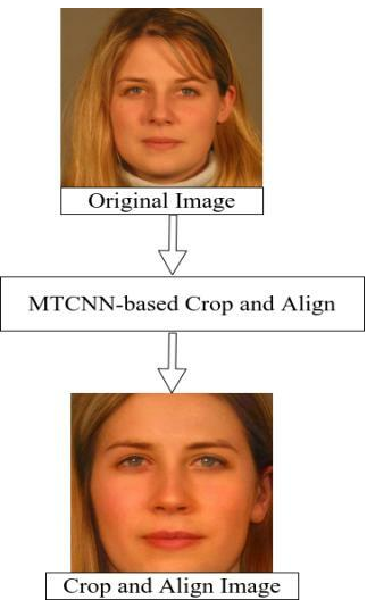}
    \caption{\centering{Sample Image Cropped and Aligned by MTCNN.}}
    \label{fig:fig9}

    \end{multicols}
    \centering
    \includegraphics[width=0.5\linewidth]{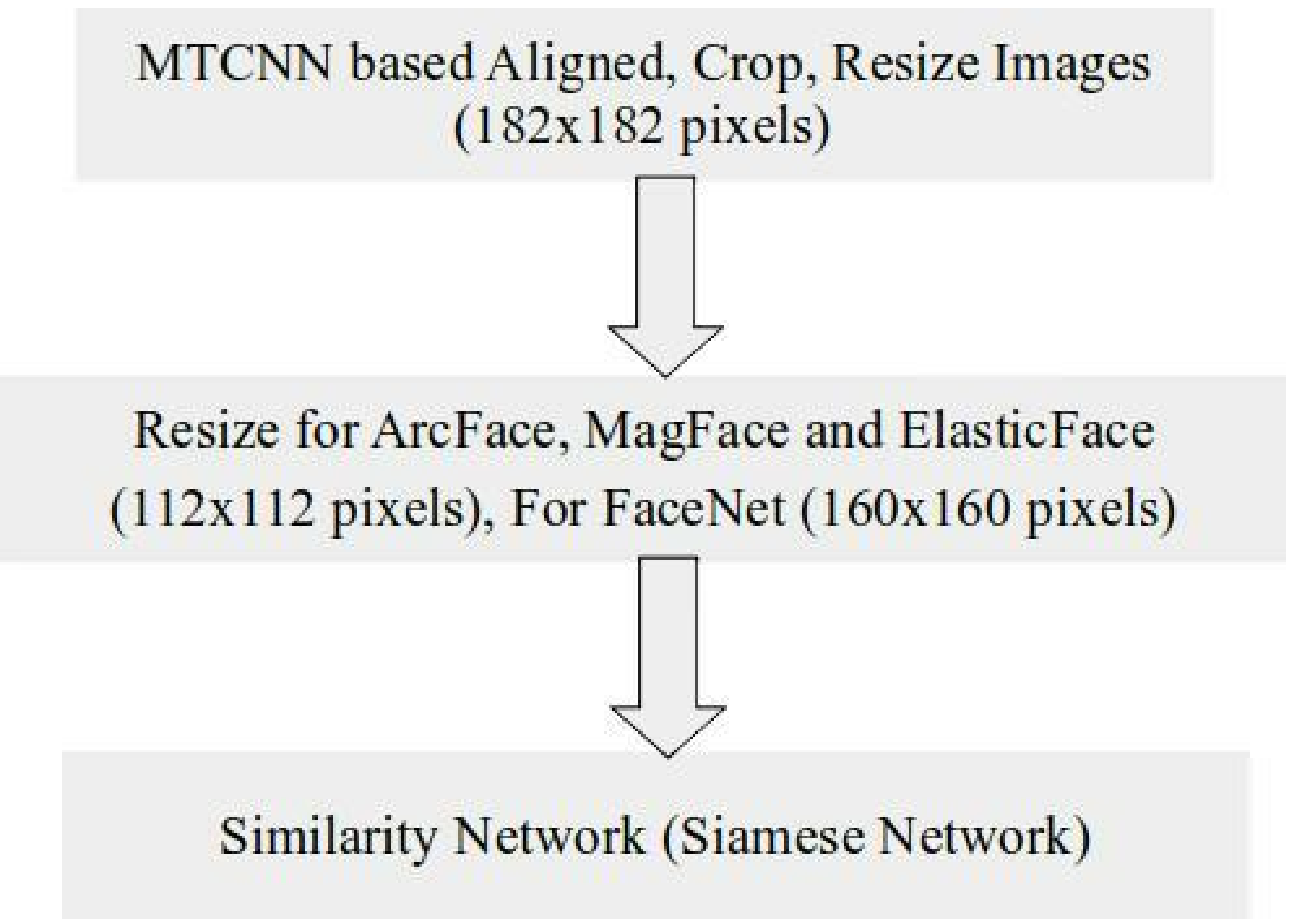}
    \caption{Image Resizing in the Similarity Network.}
    \label{fig:fig10}

\end{figure*}

\label{sec:headings}

\subsection{Matching Experimentation} 

After compiling and preparing the face datasets, the initial task of the work presented here was to analyze the performance of two facial recognition systems on these datasets. A summary of the face recognition tools used in this work is provided in Table \ref{tab:6}.  

\begin{tabularx}{\linewidth}{ c |  X | X |c }
    
    \caption{Facial recognition tools used in this effort.}\label{tab:6}\\
    \hline
    
    & Name of Facial Recognition Tool &  Description of FR tools & Network Parameters \\
    \hline
    
    01 &

    COTS (Commercial off the shelf) matcher \citep{Chen_2018-bt} &

    Neurotechnology V4 VeriLook matcher; a multi-step feature-based matcher. &

    N/A \\
\hline
    02 &

    FaceNet matcher \citep{Schroff_2015_CVPR} &

    Based on the Inception-ResNet v1 architecture. &

    23,482,624 \\
    \hline
    03 &

    ArcFace loss \citep{deng2019arcface} based matcher  &

    The backbone network is ResNet-100. &

    65,156,160 \\
    \hline
    04 &

    ElasticFace loss \citep{Boutros_2022_CVPR} based matcher  &

    The backbone network is ResNet-100. &

    65,156,160 \\
    \hline
    05 &

    MagFace loss \citep{meng2021magface} based matcher  &

    The backbone network is Mobile FaceNet \citep{chen2018mobilefacenets}. &

    1,200,512 \\
        \hline
\end{tabularx}

The commercial off the shelf matcher used was the Neurotechnology V4 Verilook matcher \citep{neurotechnology_2021}. This matcher is a multi-step feature-based matcher which first creates feature templates of a given face image, then compares those feature templates to return a comparison score for the given faces. The exact features used in the template creation and the comparison method used by the matcher are proprietary technology, hence the “black-box” name given to this type of facial recognition tool. 

The first experiment performed was a baseline analysis of the matchers when presented with only the identical twin pairs from the twin dataset to determine the effect of highly similar faces on the non-mated distribution of a FR experiment. In addition, a mated comparison was made for each identity to show the relationship between the identical twin non-mated distribution and mated distribution. The mean comparison score of the identical twins in this baseline experiment is used as the threshold for each of the remaining comparison experiments. This score represents an experimental threshold for individuals with high facial similarity (as determined by their comparison score, not similarity score), and is used later in this work to extract potential look-alikes in our dataset for further analysis by the proposed similarity network. For the remaining comparison experiments, the approach presented in Howard et al.\citep{Howard2019} was used, wherein all-to-all matching was performed on each of the face datasets retaining only the non-mated, or impostor, distributions. The remaining comparison experiments correspond to each of the face datasets used in this work and are as follows: twin dataset, non-twin dataset, and the combined non-twin and CelebA dataset (referred to as the large scale non-twin dataset henceforth). 

These match experiments are designed to show the effect of highly similar face pairs on two typical facial recognition tools, as well as identify potential look-alike pairs from the datasets used in this study. By first performing both mated and non-mated matching using only the identical twin pairs, a baseline of the worst case of facial recognition performance can be ascertained. As stated above, the average result of the non-mated comparison score distribution from this initial baseline experiment represents an experimental threshold for later matching experiments. This threshold represents the average non-mated comparison score of the absolute worst case presented to a face recognition tool, and it is expected that any non-mated comparison score from a non-twin pair falling above this threshold may be a look-alike pair. By performing all-to-all non-mated matching on all the datasets and extracting the pairs with comparison scores falling above the experimental threshold, these potential look-alike pairs can be extracted and examined further with the proposed similarity network described in the next section. 

\subsection{Similarity Network }

\begin{figure}
    \includegraphics[width=\columnwidth]{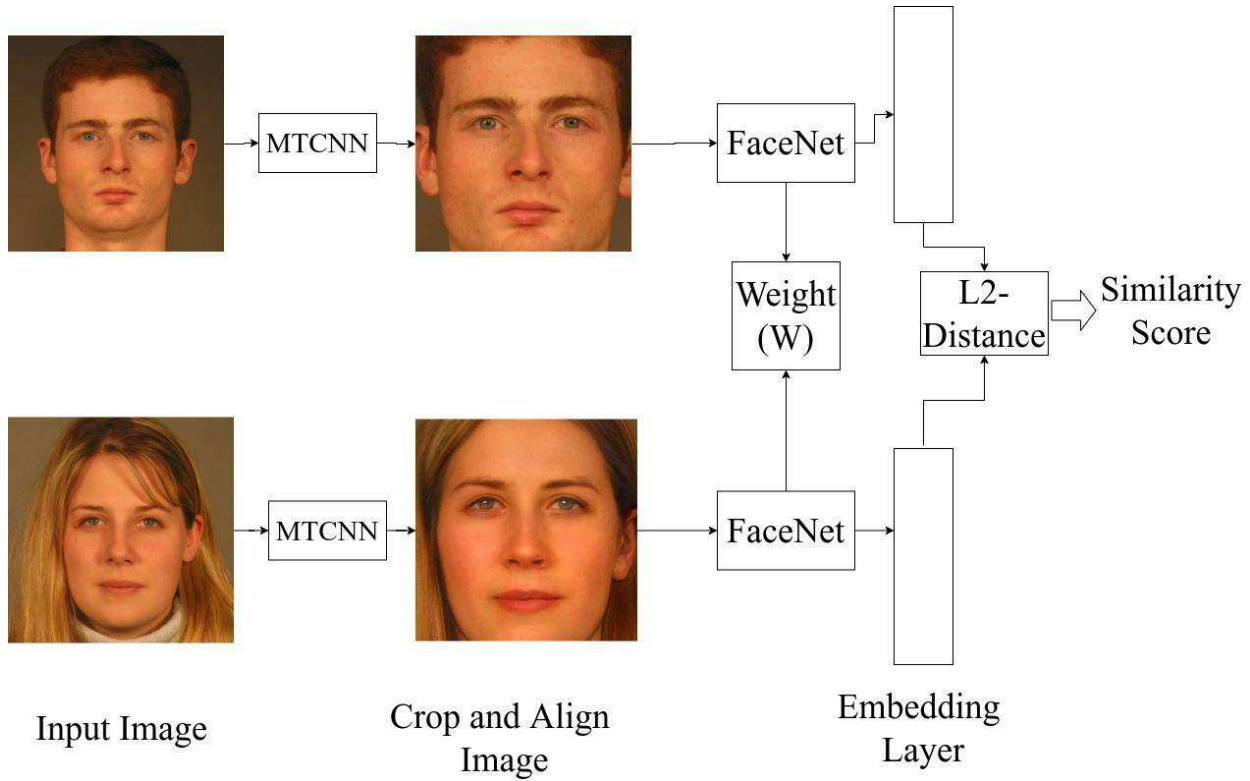}
    \caption{Similarity network diagram.}
    \label{fig:fig11}
\end{figure}

To determine a quantitative measure of similarity between identical twins, a deep convolutional neural network was implemented. This network was designed with a Siamese architecture to directly compare two faces. Each half of this network was comprised of a FaceNet or ArcFace or ElasticFace or MagFace architecture, with the weights of the network shared between the two halves (Figure \ref{fig:fig11}, for FaceNet). 

The FaceNet architecture was used as the foundation of this network for its high accuracy on typically difficult, similar face images. In addition to the high accuracy of the network, this network was also chosen for its ability to generate highly representative face embeddings. In the original publication \citep{Schroff_2015_CVPR}, it is stated that, “The network is trained such that the squared L2 distances in the embedding space directly correspond to face similarity: faces of the same person have small distances and faces of distinct people have large distances.” This is advantageous, as this work seeks to quantify the similarity of identical twin pairs. 

The proposed network was optimized using the contrastive loss function shown in the equation below \citep{Hadsell}:

\[L = (1-y)*D(x_{1},x_{2})^{2} + y*(0,m - D(x_{1},x_{2}))^{2} \]

where $(x_{1}$ and $x_{2})$are the two input face images, $D(*)$ is the L2 difference between the embeddings generated by the network, 
y is the ground-truth label for the image pair, and 
m is the contrastive margin. Contrastive loss seeks to minimize the L2 distance between similar samples who reside close to one another in the feature space while maximizing the L2 distance between the dissimilar samples. 

The output of this network consists of the L2 distance between two samples in the feature space. As this result gives similar samples a low score and dissimilar samples a high score, for clarity, the scores are inverted such that similar samples have a high score and dissimilar samples a low score. This was achieved by subtracting each resultant similarity score from the maximum similarity score in a given set of scores. 

The training phase of the network for the task at hand consisted of fine tuning the weights with data from the twin dataset. Starting with network pre-trained on the VGGFace2 dataset, the network was fine-tuned on a subset of the twin database. This fine tuning consisted of training over four epochs at a learning rate of $10^{-5}$ with a margin equal to 0.5. This fine tuning was performed on a tailored verification task where a pair of identical twin images represents the positive case, and a pair of unrelated look-alikes represents the negative case. Training was stopped when the AUC of the network began to degrade on the verification task. This training encouraged the network to group together those samples with the facial similarity of identical twins in the embedded feature space, and inversely pushed apart those samples not as similar as identical twins. 

The main training and testing dataset for this network was comprised of a subset of the twin dataset. This dataset contained images of identical twin pairs and imposter look-alikes to the twin pairs sorted into an equal number of genuine and impostor pairs, where a genuine pair consists of (Twin A vs. Twin B), and an impostor pair consists of (Twin A vs. look-alike). The look-alikes for each identity were found by selecting the identities with the highest FaceNet match score to each twin identity. This training schema is chosen because the network should learn to determine facial similarity from the most similar face pairs available (i.e., identical twins). It is expected that an individual’s identical twin will be more similar to that individual than any potential look-alike, as such, the network is trained to identify the face pairs with the highest facial similarity. The dataset contains 645 identical twin identities, with a total of 3,203 images, split 80/20 for training and testing, respectively. The twin identities used in the training of the similarity network were chosen based upon their non-mated comparison scores from the FaceNet matcher against their identical twin. Approximately 50\% of the twin identities with the highest non-mated Twin A vs Twin B non-mated comparison scores make up the 645 training identities. 

For both the training and testing of the network the input face images were sorted into dataset folders containing either A) images of twin A and twin B from an identical twin pair or B) images of twin A and images of twin A’s look-alike(s) as described above. During training and testing the network was fed samples randomly from either folder A or B for each identity with a 50/50 chance of the input samples coming from either category A or B. This balances the input sample distribution so that on average the network learns from an equal number of positive and negative cases from the compiled dataset.

\begin{figure}

\begin{multicols}{2}
    \includegraphics[width=\columnwidth]{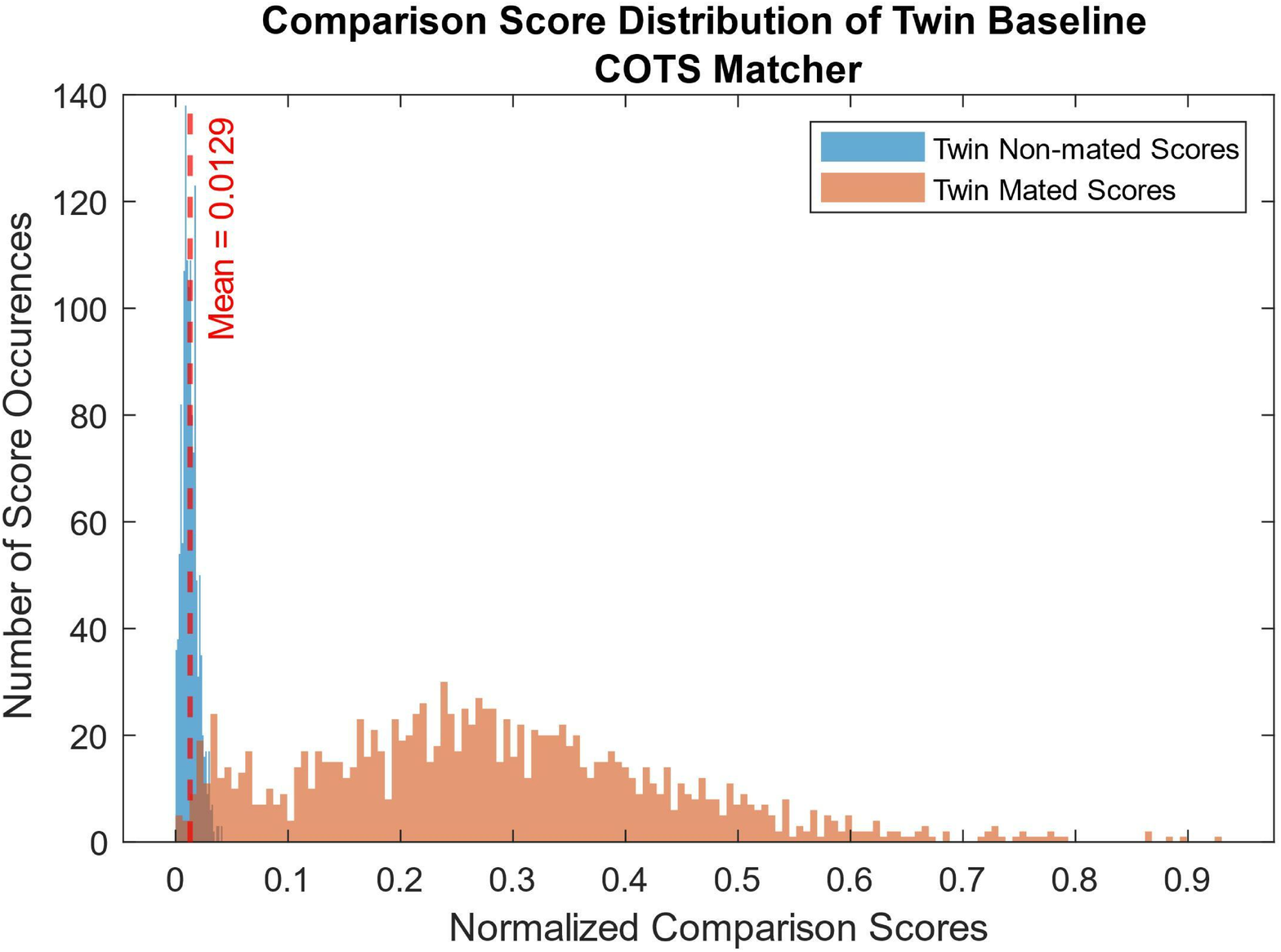}
    \caption{\centering{Twin baseline match experiment results, Neurotechnology matcher.}}
    \label{fig:fig12}

    \centering
    \includegraphics[width=\columnwidth]{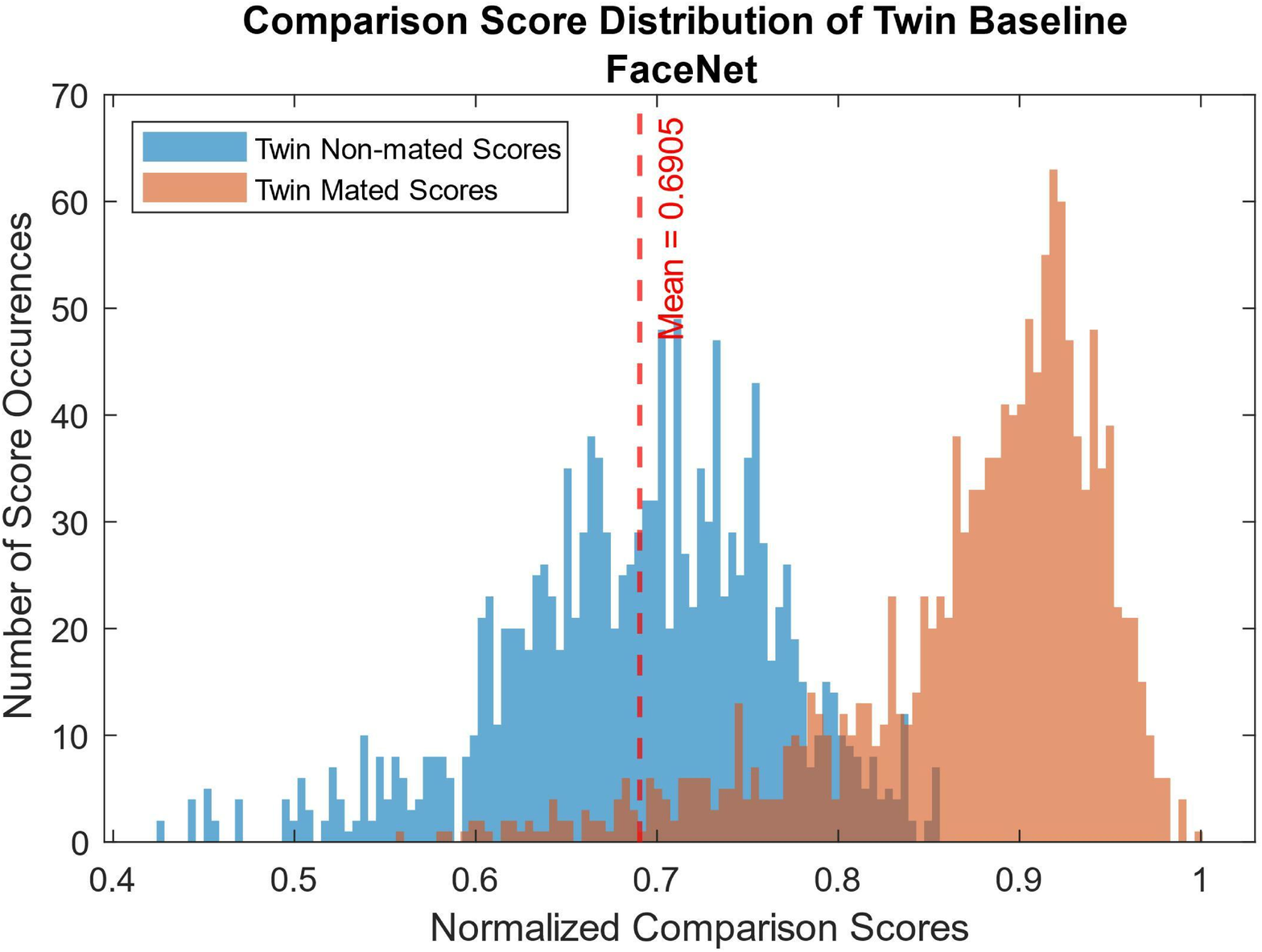}
    \caption{\centering{ Twin baseline match experiment results, FaceNet matcher.}}
    \label{fig:fig13}
    \includegraphics[width=\columnwidth]{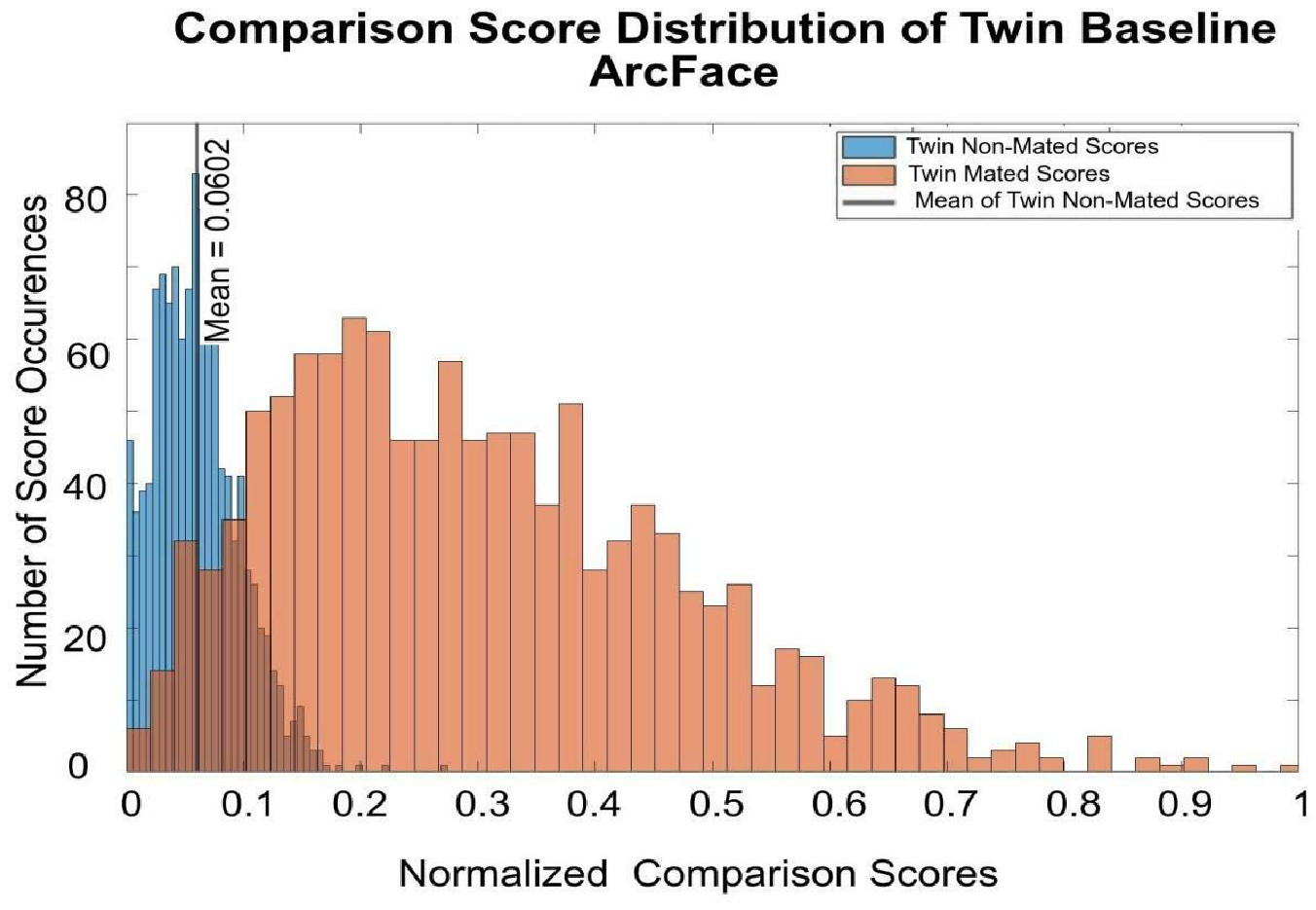}
    \caption{\centering{ Twin baseline match experiment results, ArcFace (ResNet-100 backbone) matcher.}}
    \label{fig:fig14}
        \includegraphics[width=\columnwidth]{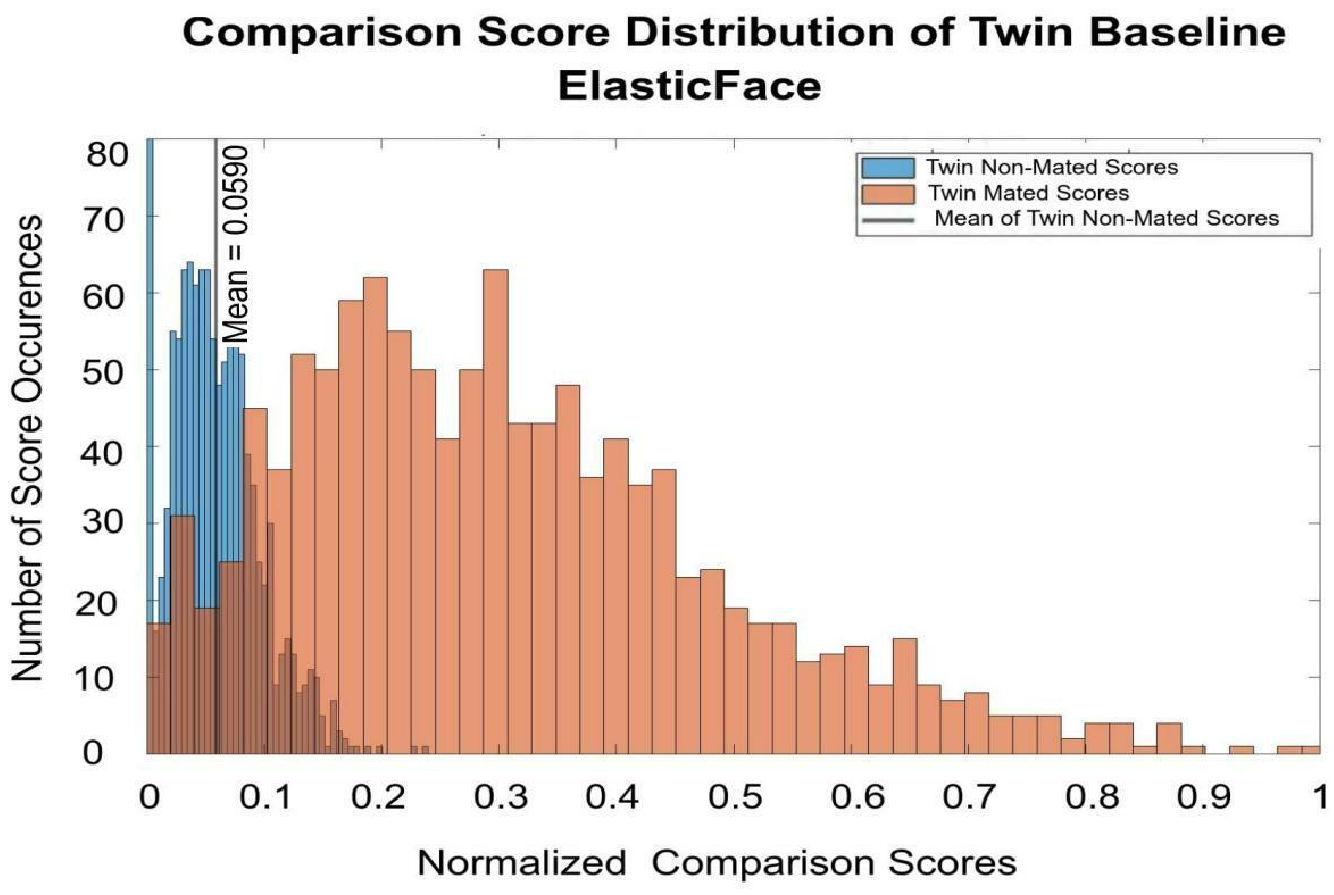}
    \caption{\centering{Twin baseline match experiment results, ElasticFace (ResNet-100 backbone)}}
    \label{fig:fig15}
    \includegraphics[width=\columnwidth]{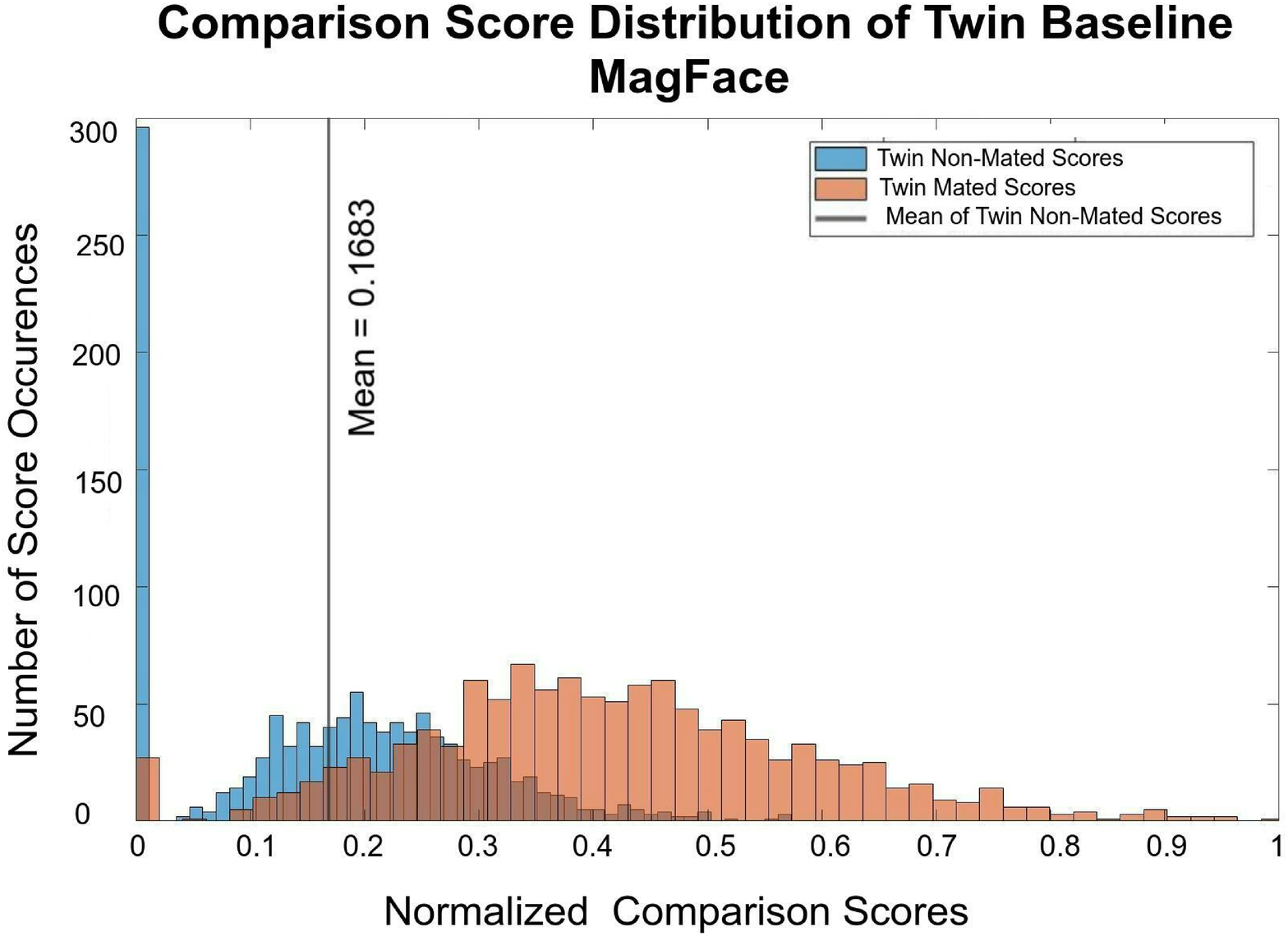}
    \caption{\centering{Twin baseline match experiment results, MagFace (Mobile FaceNet backbone) matcher.}}
    
    \label{fig:fig16}
    \includegraphics[width=\columnwidth]{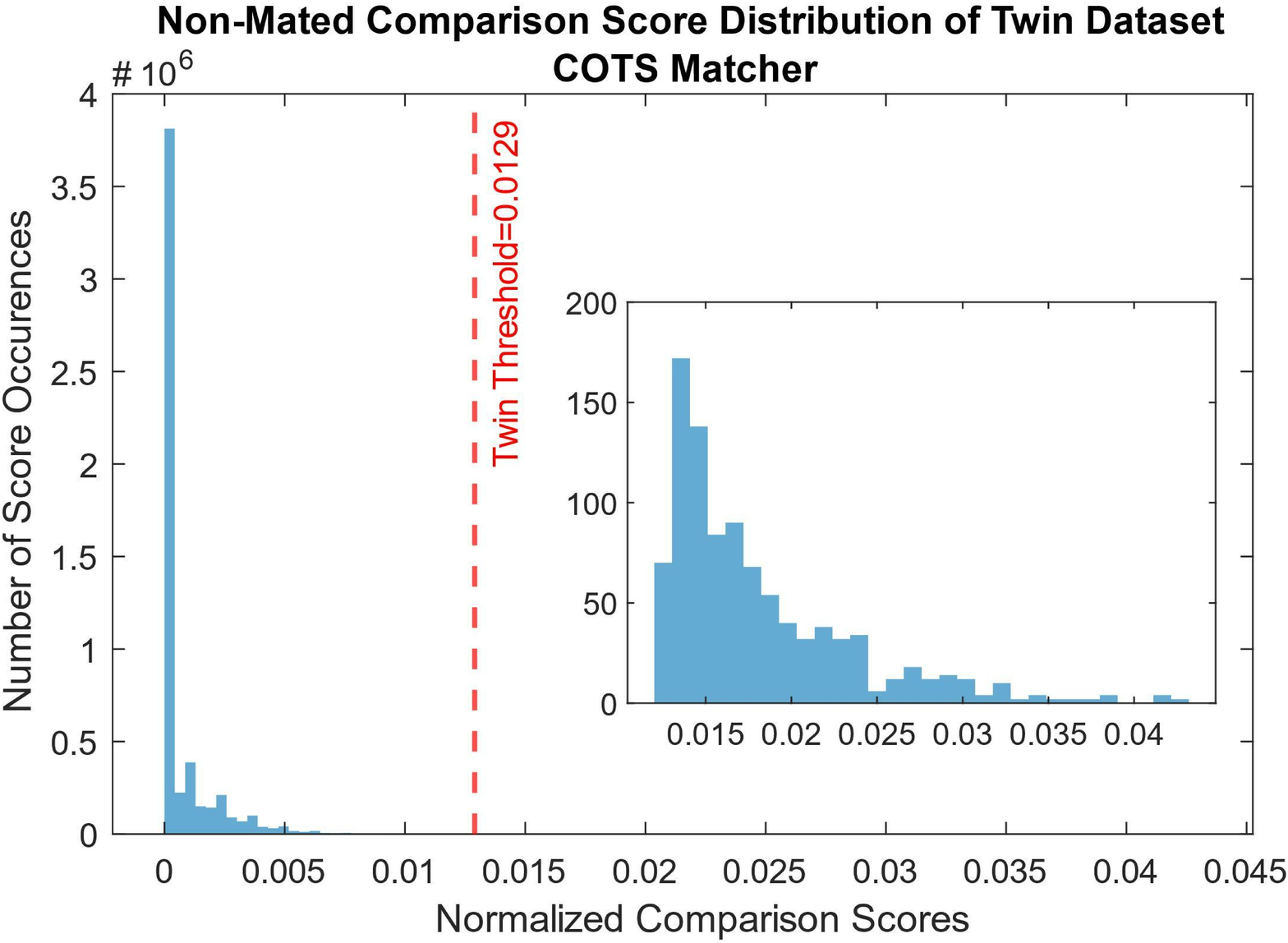}
    \caption{\centering{Twin dataset all-to-all non-mated match experiment results, Neurotechnology COTS matcher.}}
    \label{fig:fig17}
    \end{multicols}
\end{figure}

\begin{figure}
    \begin{multicols}{2}

    \includegraphics[width=\columnwidth]{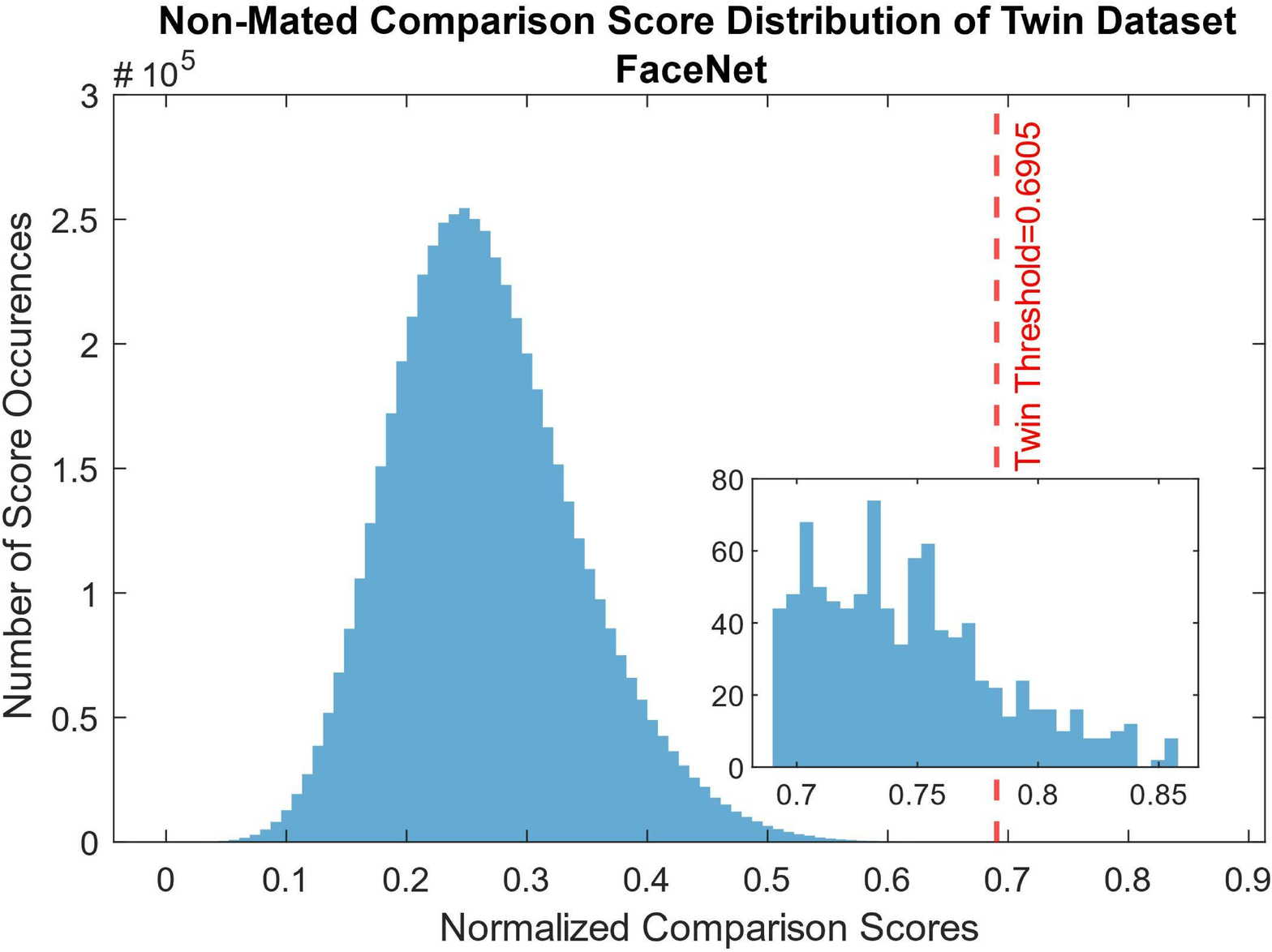}
    \caption{Twin dataset all-to-all non-mated match experiment results, FaceNet matcher. }
    \label{fig:fig18}

    \centering
    \includegraphics[width=\columnwidth]{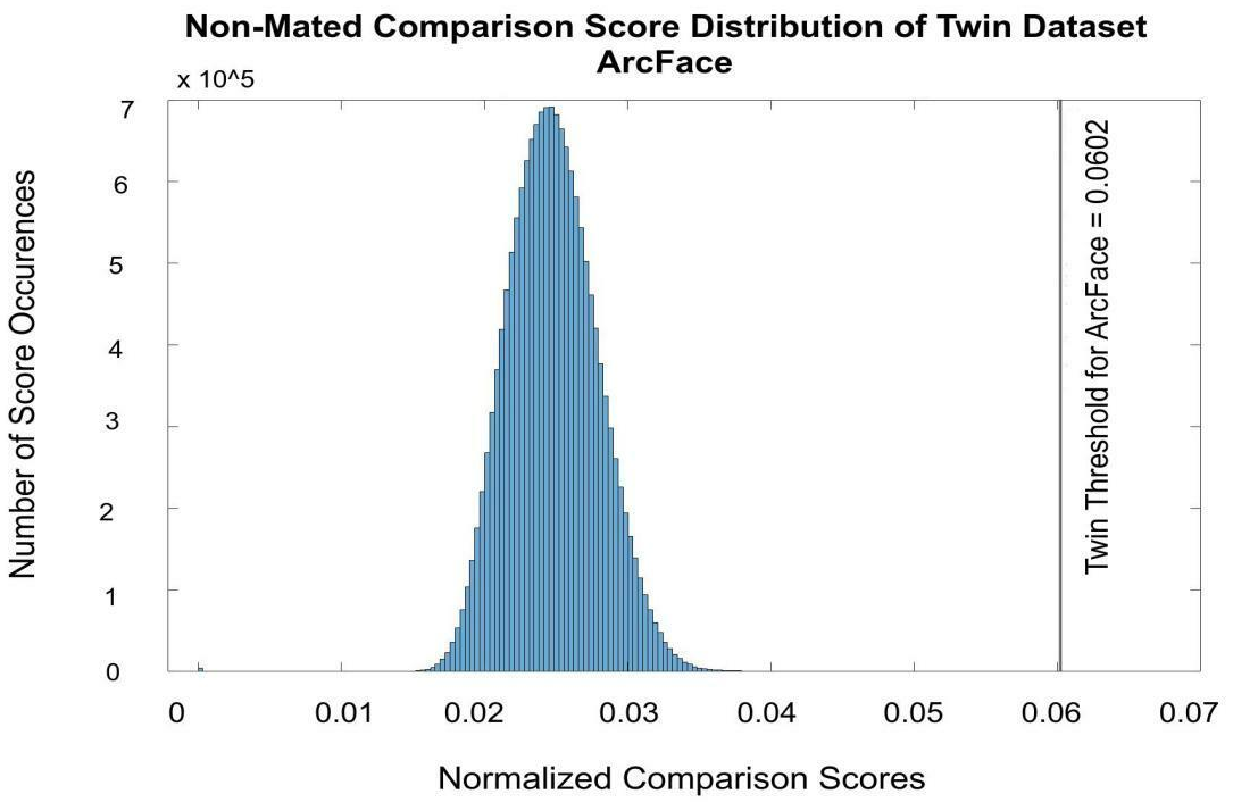}
    \caption{Twin dataset all-to-all non-mated match experiment results, ArcFace (ResNet-100) matcher. }
    \label{fig:fig19}
    \includegraphics[width=\columnwidth]{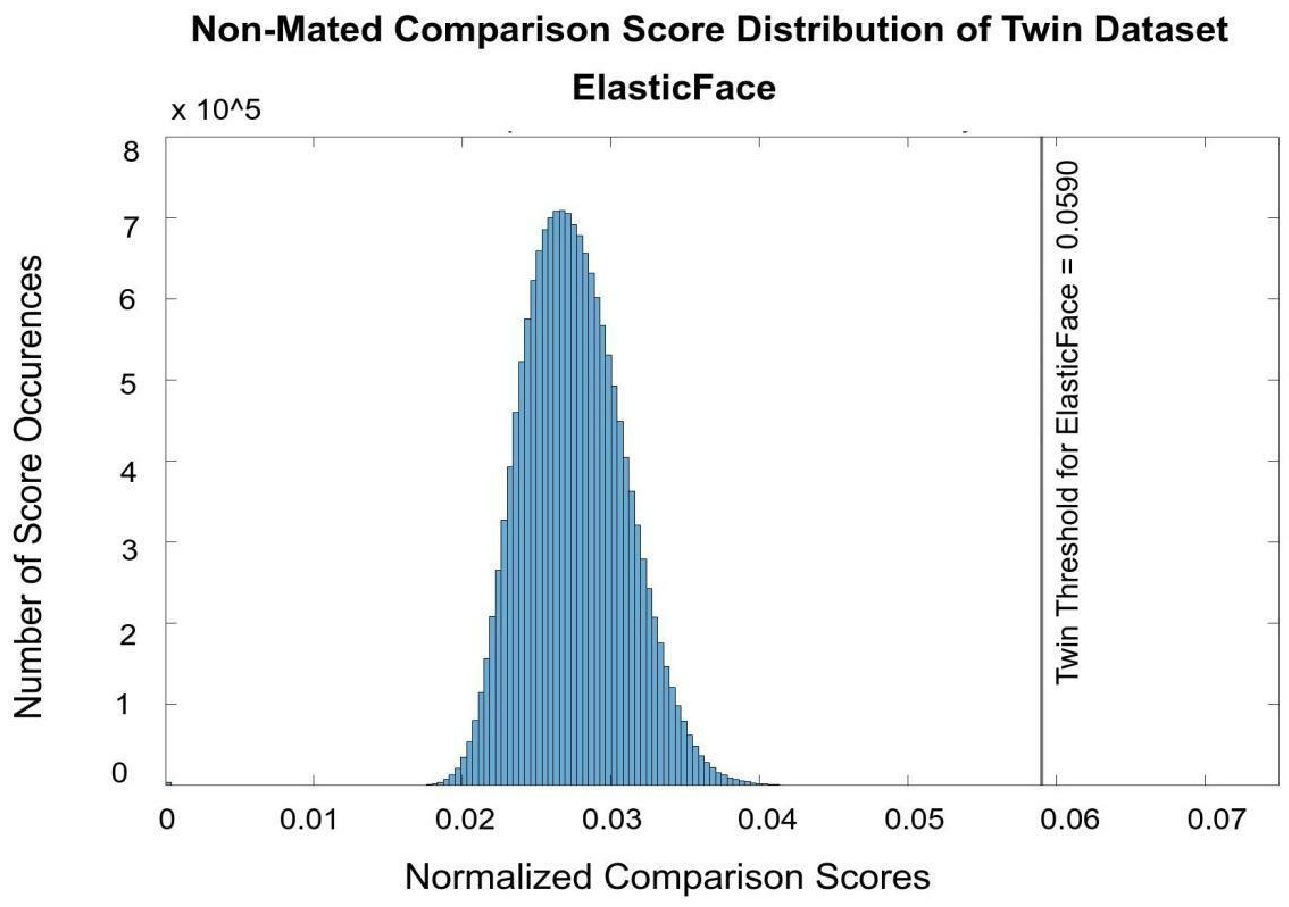}
    \caption{Twin dataset all-to-all non-mated match experiment results, ElasticFace (ResNet-100) matcher. }
    \label{fig:fig20}

    \centering
    \includegraphics[width=\columnwidth]{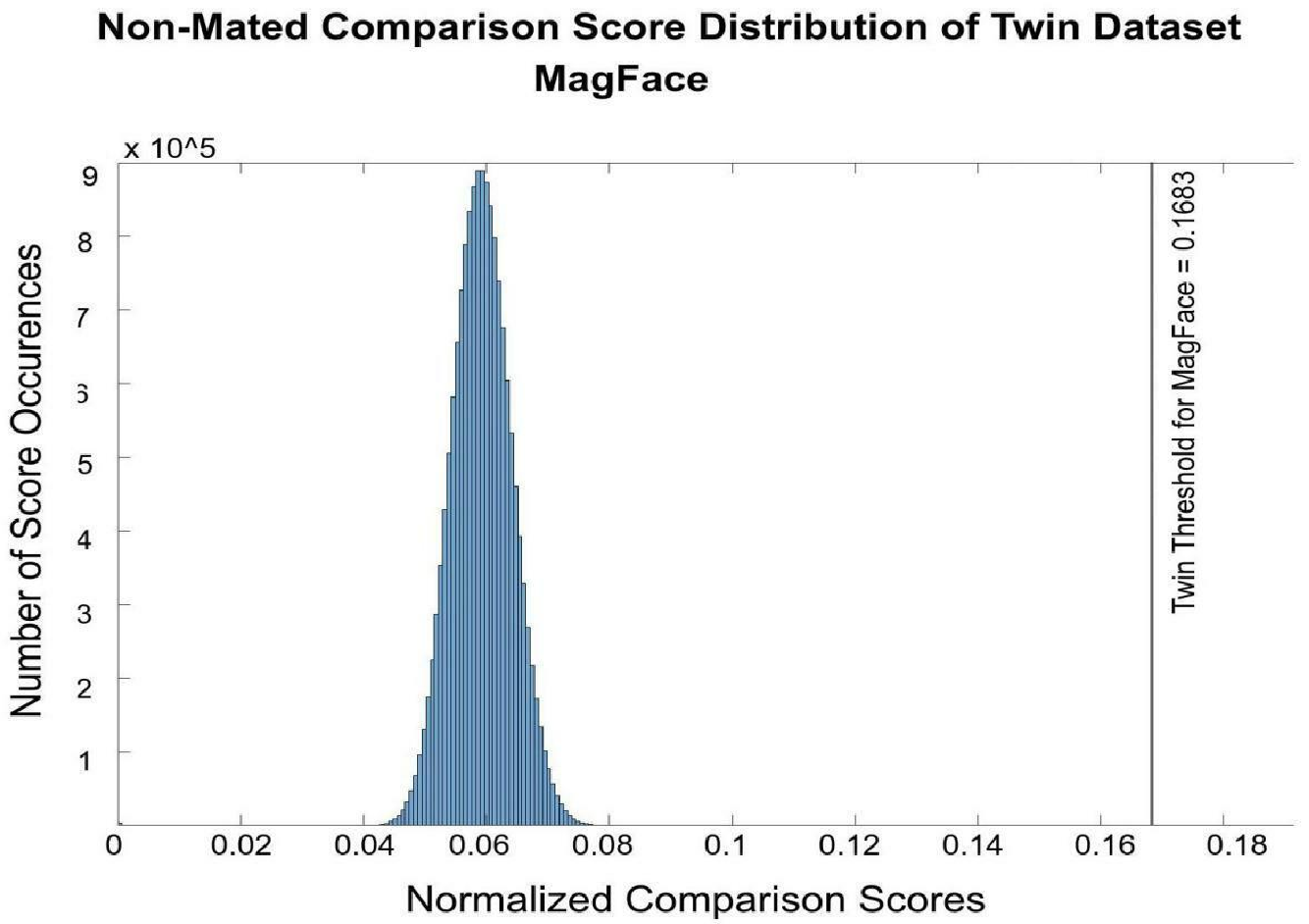}
    \caption{Twin dataset all-to-all non-mated match experiment results, MagFace (Mobile FaceNet) matcher.}
    \label{fig:fig21}
    \includegraphics[width=\columnwidth]{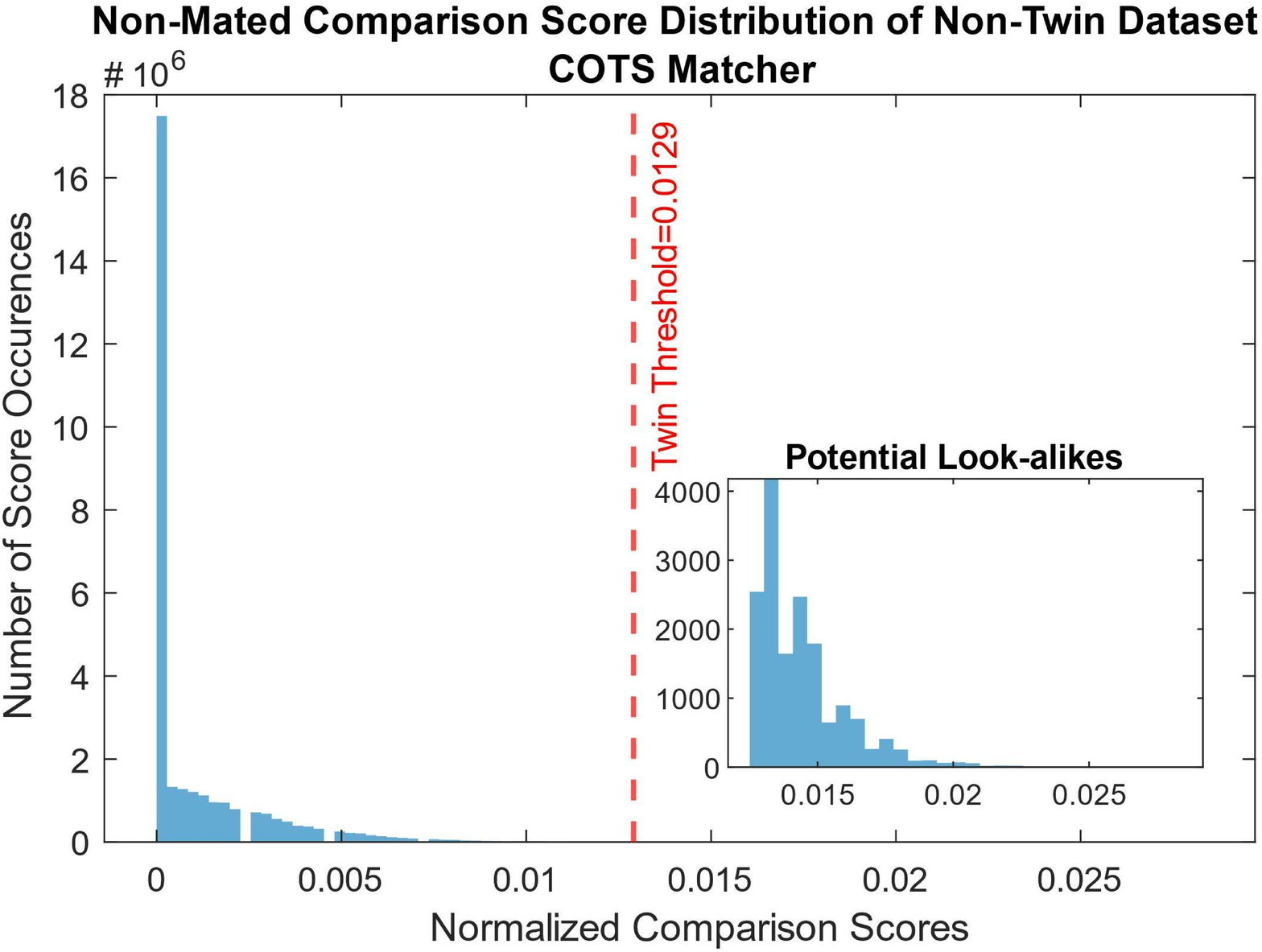}
    \caption{Non-twin dataset all-to-all non-mated match experiment results, Neurotechnology COTS matcher.}
    \label{fig:fig22}

    \includegraphics[width=\columnwidth]{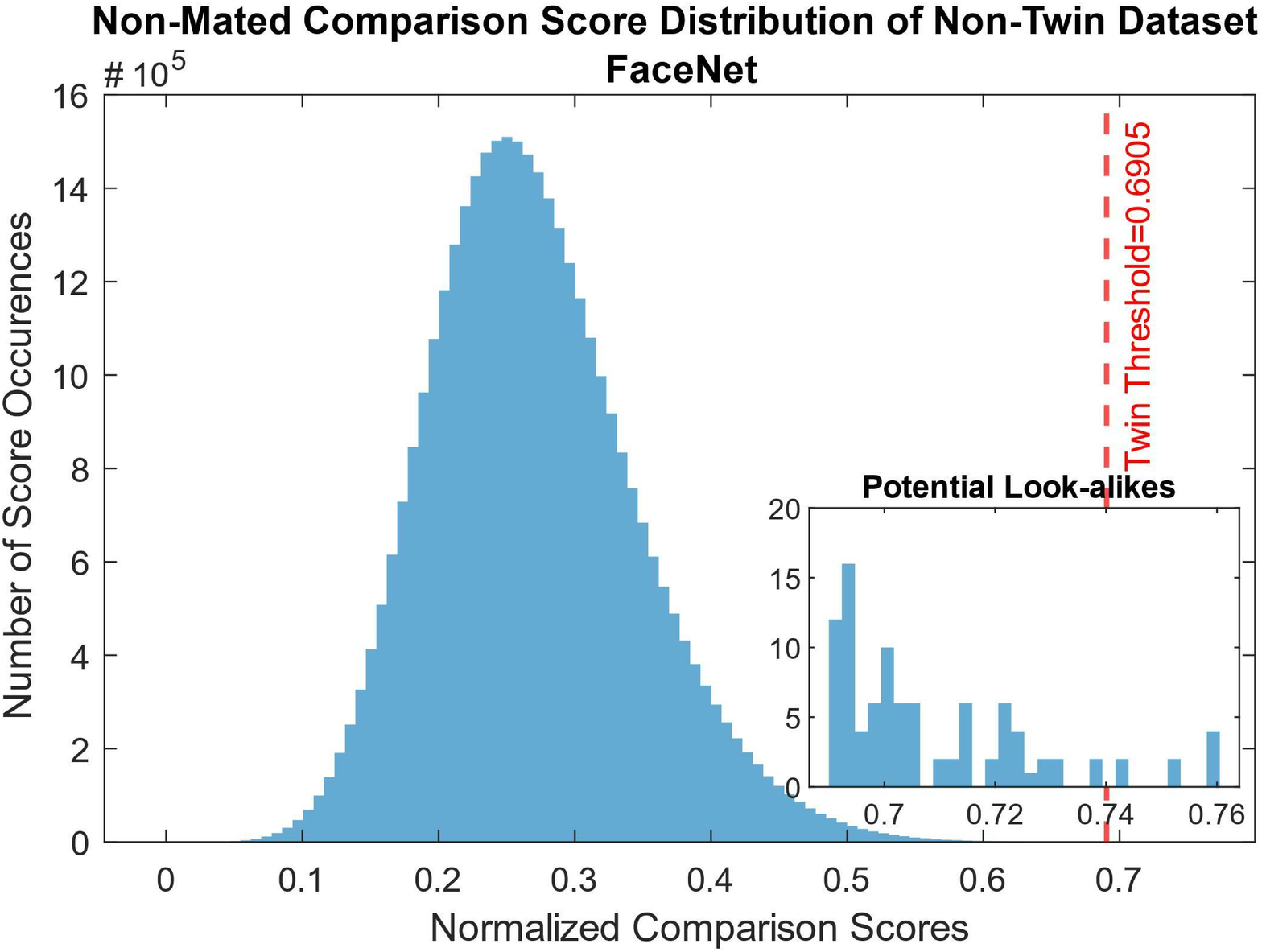}
    \caption{Non-twin dataset all-to-all non-mated match experiment results, FaceNet matcher.}
    \label{fig:fig23}

    \end{multicols}
\end{figure}

\begin{figure}
    \begin{multicols}{2}
    \includegraphics[width=\columnwidth]{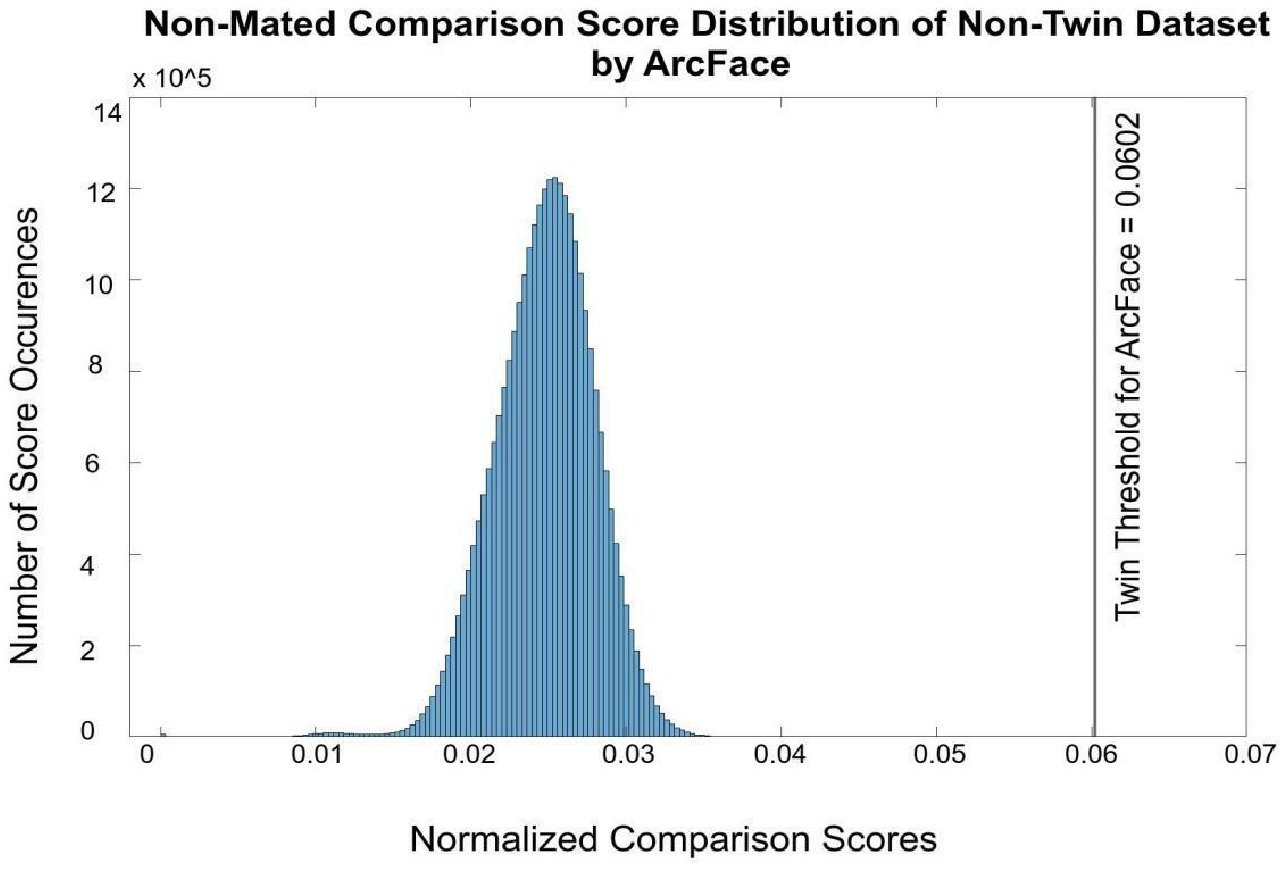}
    \caption{\centering{Non-twin dataset all-to-all non-mated match experiment results, ArcFace (ResNet-100 backbone) matcher. }}
    \label{fig:fig24}
    \includegraphics[width=\columnwidth]{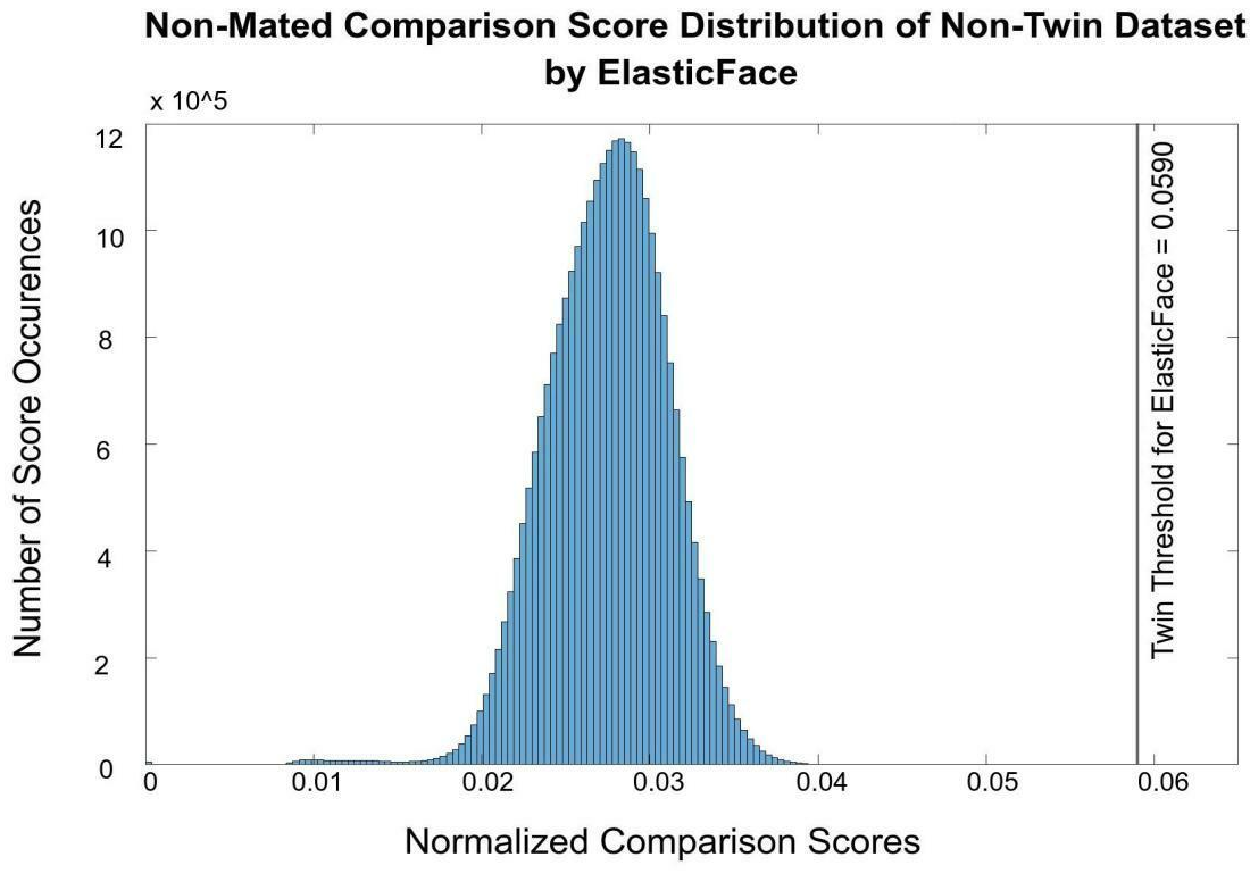}
    \caption{\centering{Non-twin dataset all-to-all non-mated match experiment results, ElasticFace (ResNet-100 backbone) matcher.}}
    \label{fig:fig25}
    \includegraphics[width=\columnwidth]{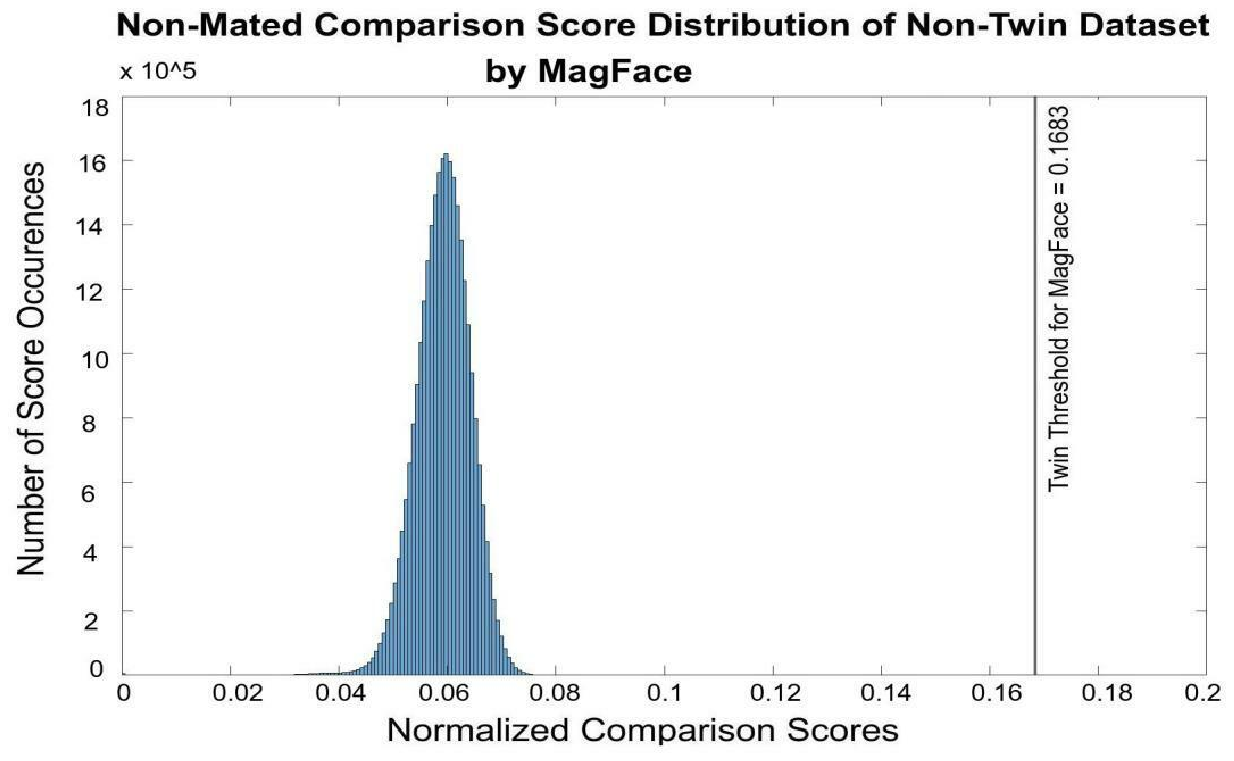}
    \caption{\centering{Non-twin dataset all-to-all non-mated match experiment results, MagFace (Mobile-FaceNet backbone) matcher.}}
    \label{fig:fig26}

    \centering
    \includegraphics[width=\columnwidth]{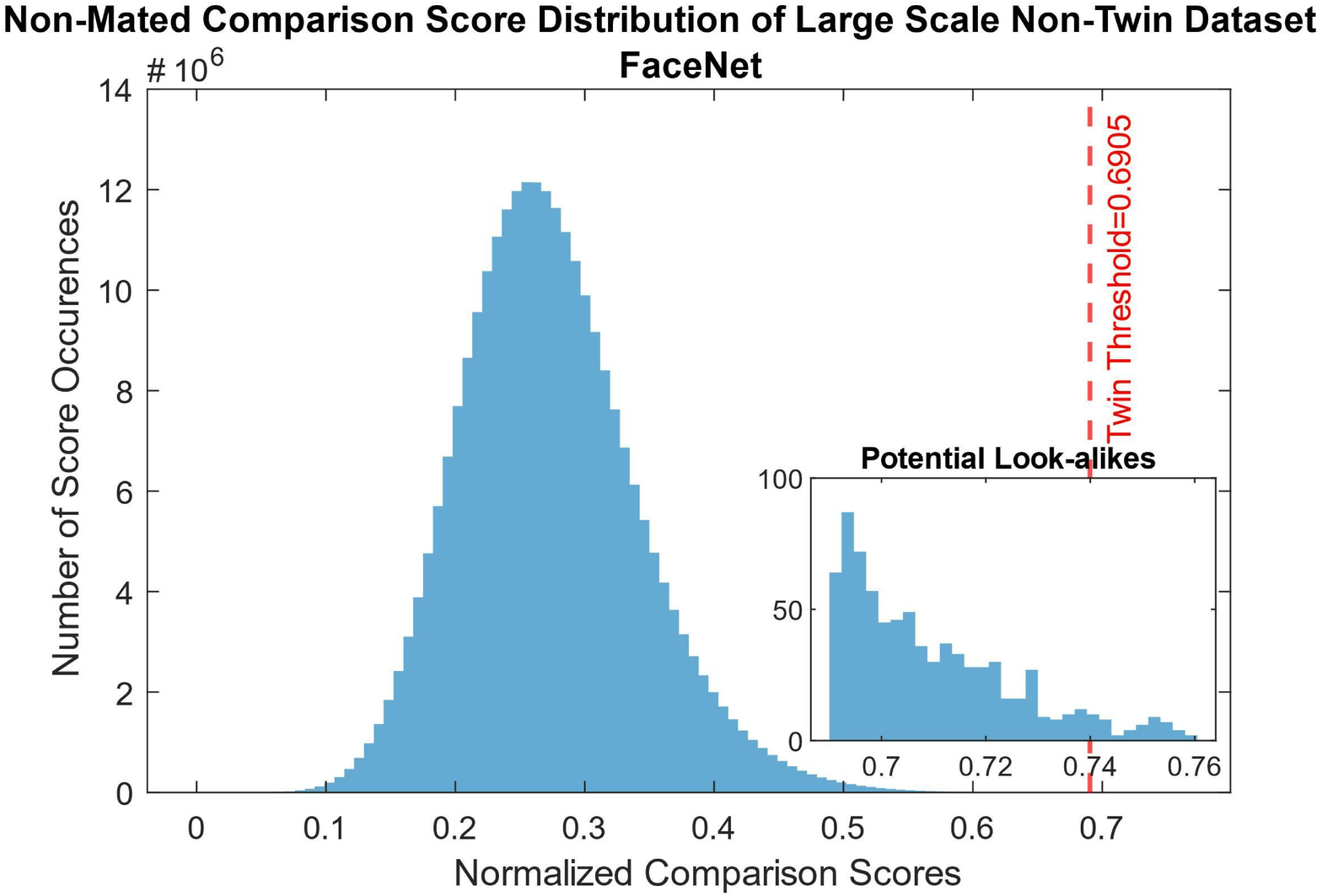}
    \caption{\centering{Large Scale non-twin dataset all-to-all non-mated match experiment results, FaceNet matcher.}}
    \label{fig:fig27}
    \end{multicols}
\end{figure}
\begin{figure}
    \begin{multicols}{2}
    \includegraphics[width=\columnwidth]{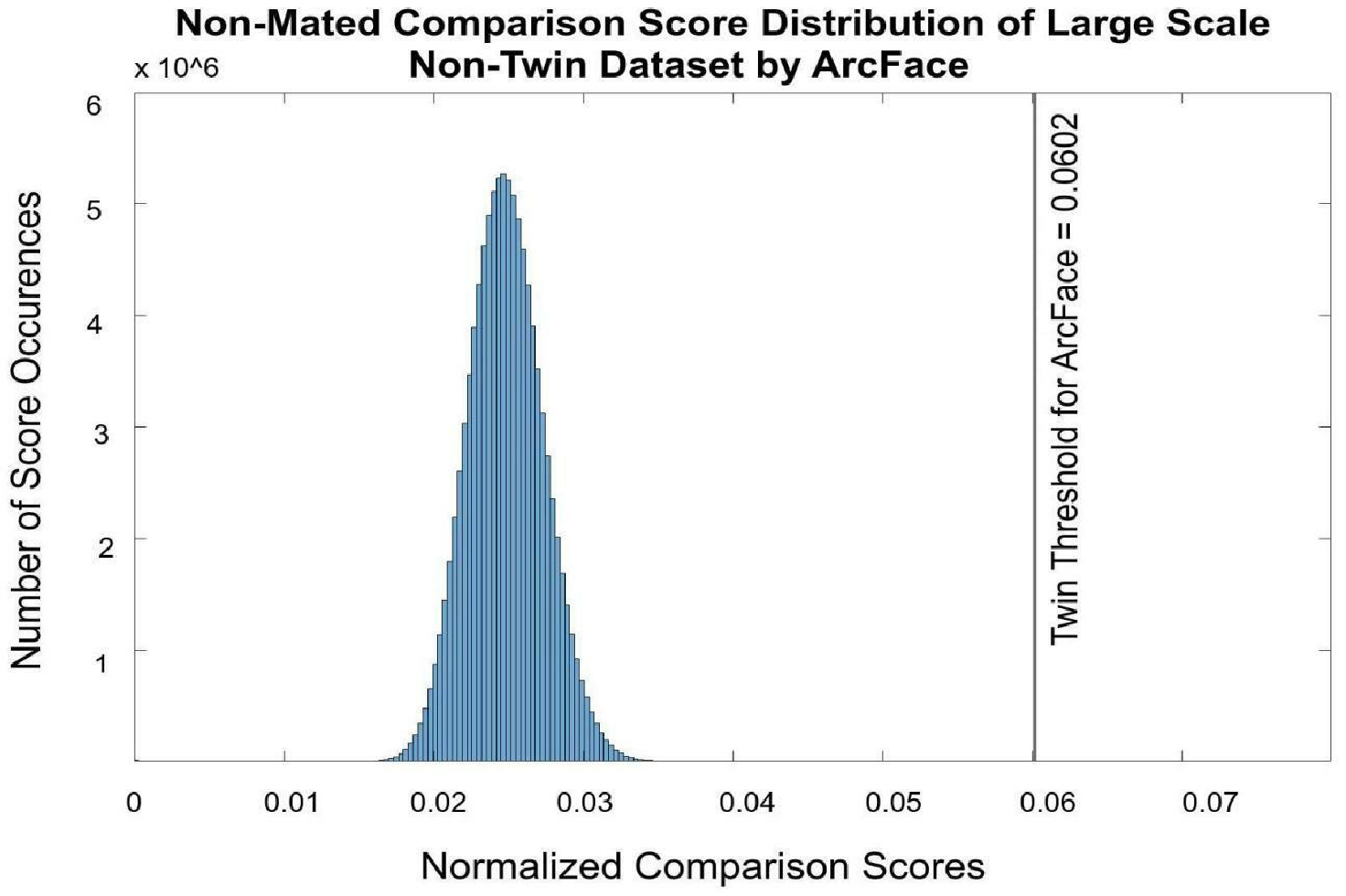}
    \caption{\centering{Large Scale non-twin dataset all-to-all non-mated match experiment results, ArcFace (ResNet-100 backbone) matcher.}}
    \label{fig:fig28}

    \includegraphics[width=\columnwidth]{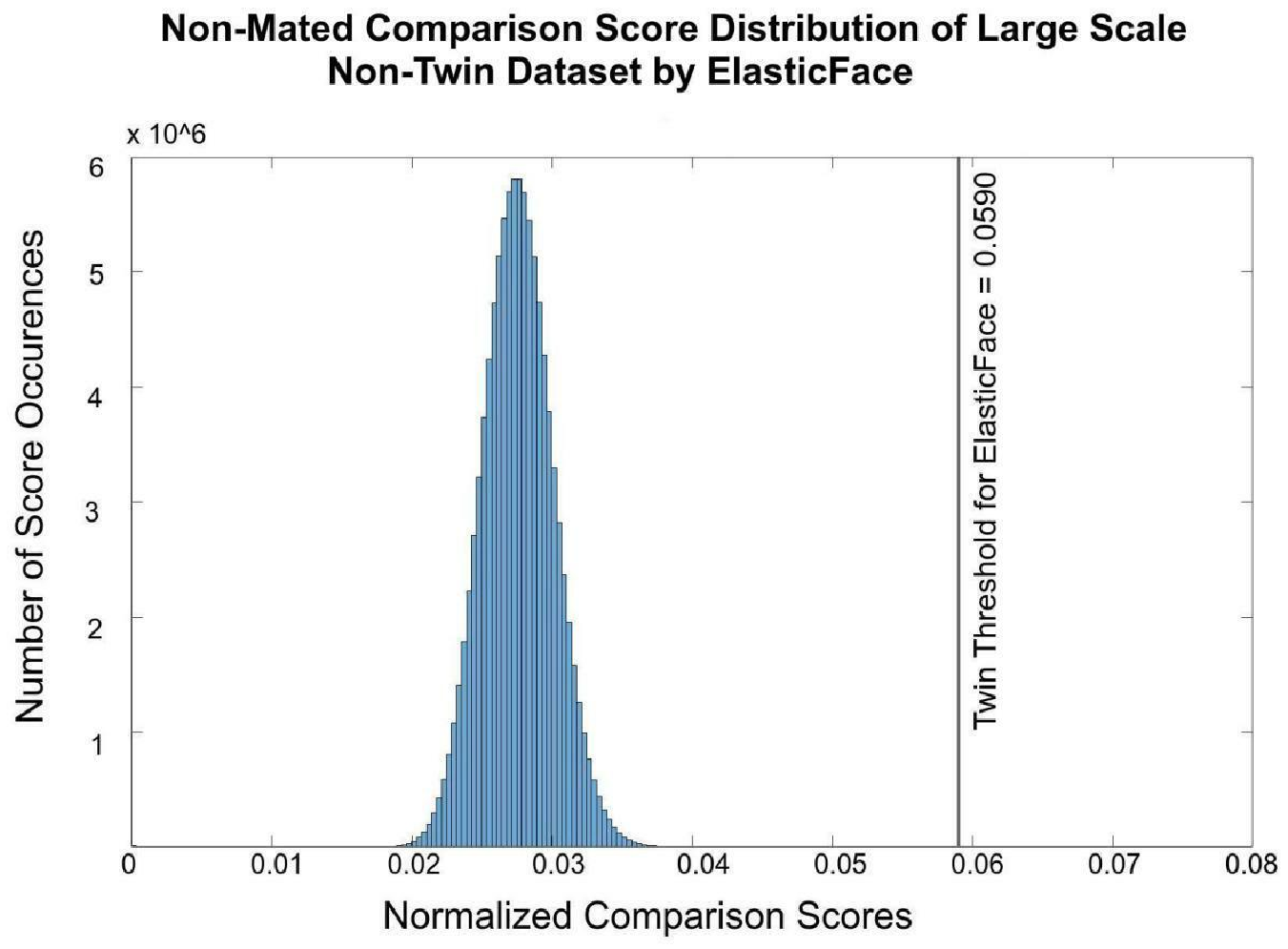}
    \caption{\centering{Large Scale non-twin dataset all-to-all non-mated match experiment results, ElasticFace (ResNet-100 backbone) matcher.}}
    \label{fig:fig29}
    \end{multicols}
\end{figure}
\begin{figure}
    \begin{multicols}{2}
    \includegraphics[width=\columnwidth]{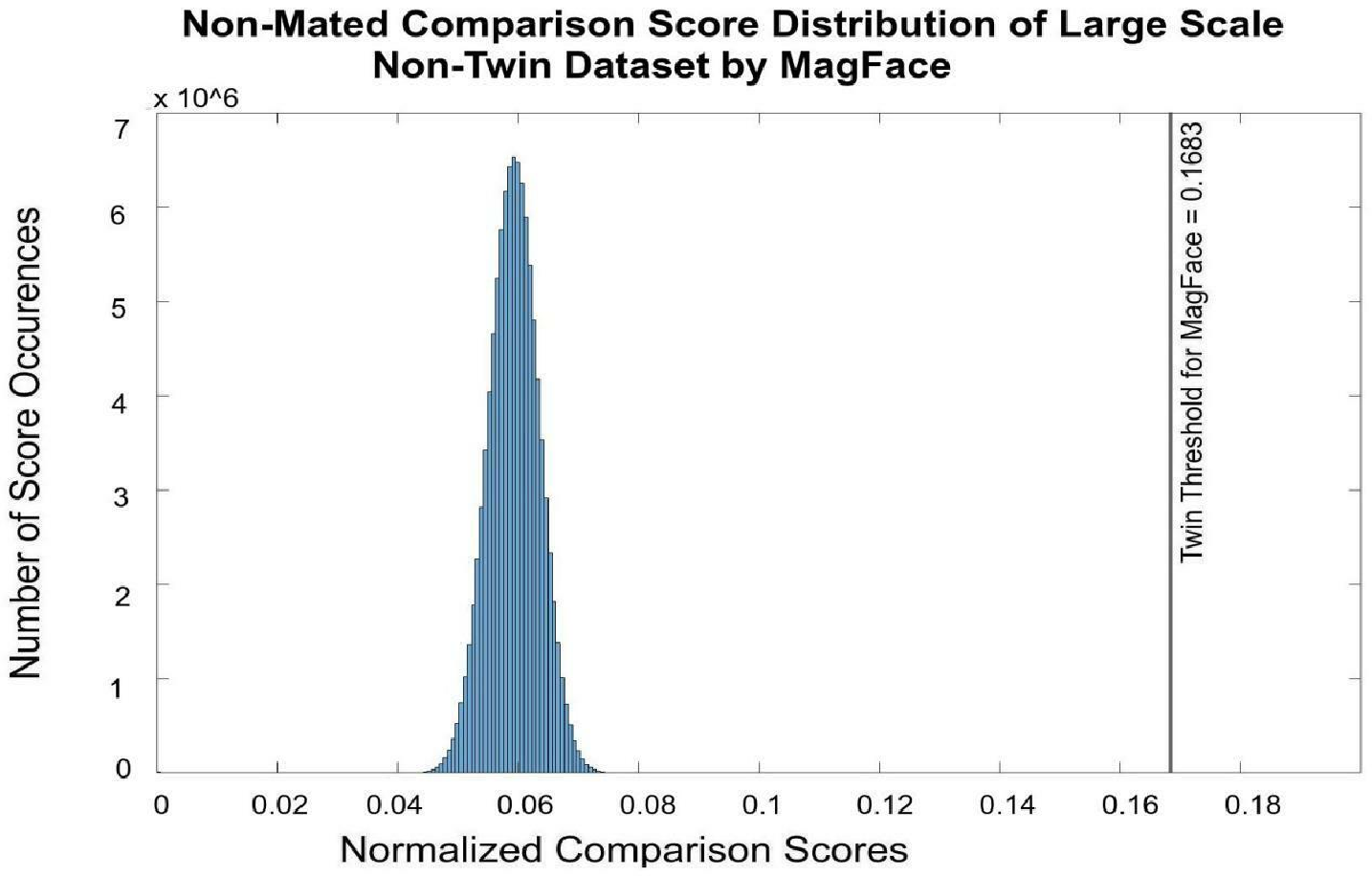}
    \caption{\centering{Large Scale non-twin dataset all-to-all non-mated match experiment results, MagFace (Mobile FaceNet backbone) matcher.}}\label{fig:fig30}

    \centering
    \includegraphics[width=\columnwidth]{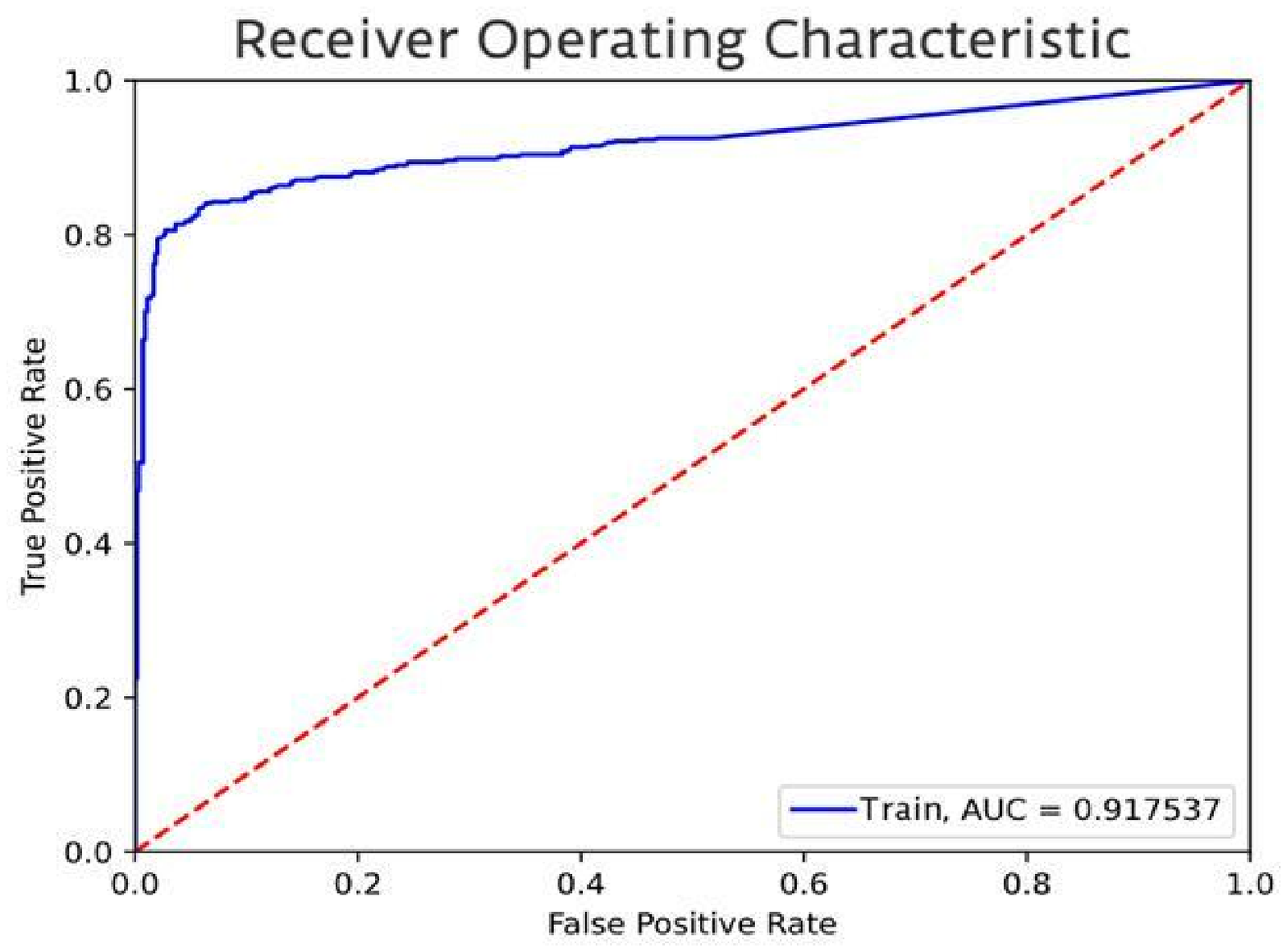}
    \caption{\centering{Train ROC curve of similarity network (FaceNet) on the tailored verification task.}}
    \label{fig:fig31}
    \includegraphics[width=\columnwidth]{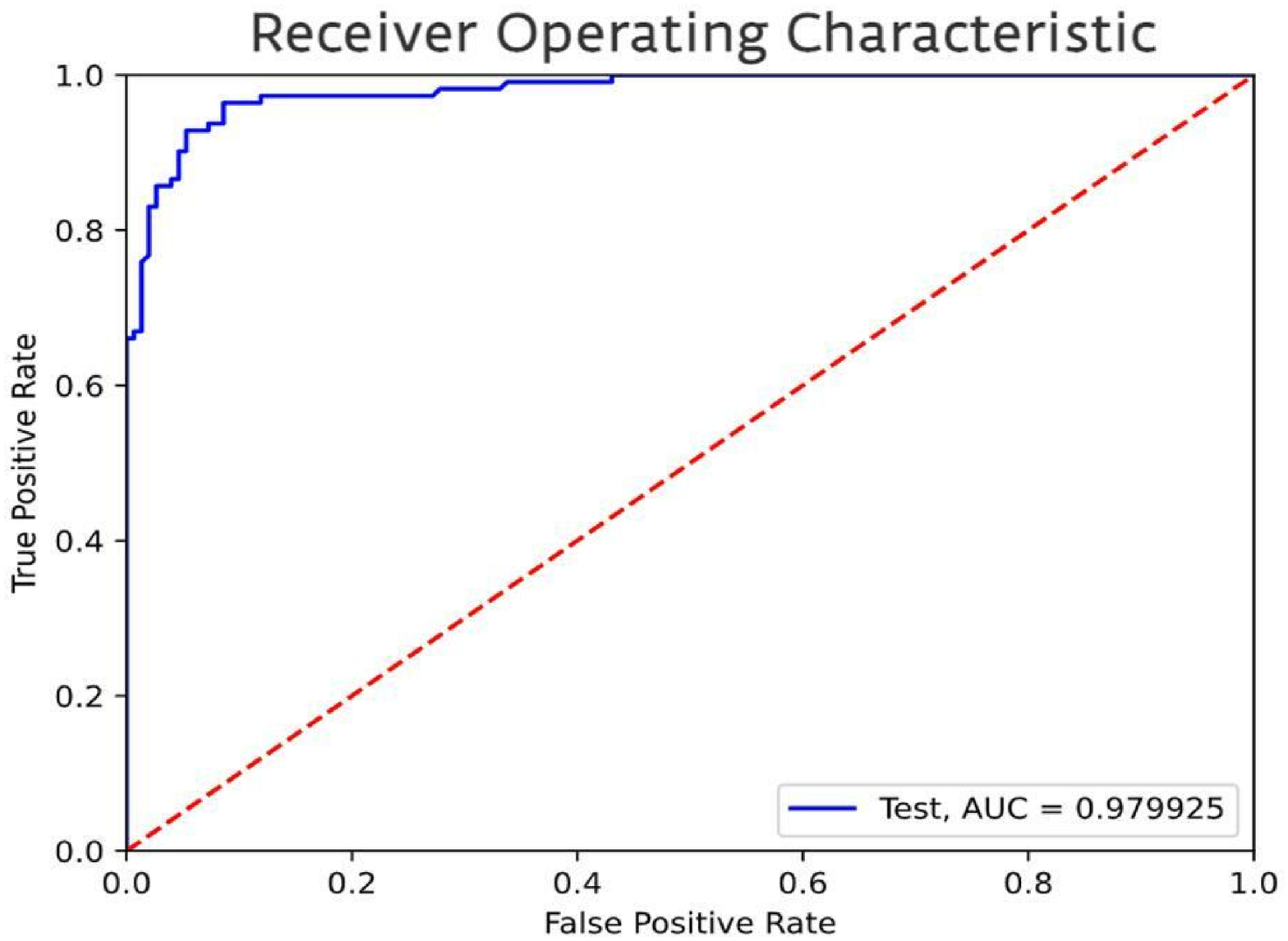}
    \caption{\centering{Test ROC curve of similarity network (FaceNet) on the tailored verification task.}}
    \label{fig:fig32}

    \centering
    \includegraphics[width=\columnwidth]{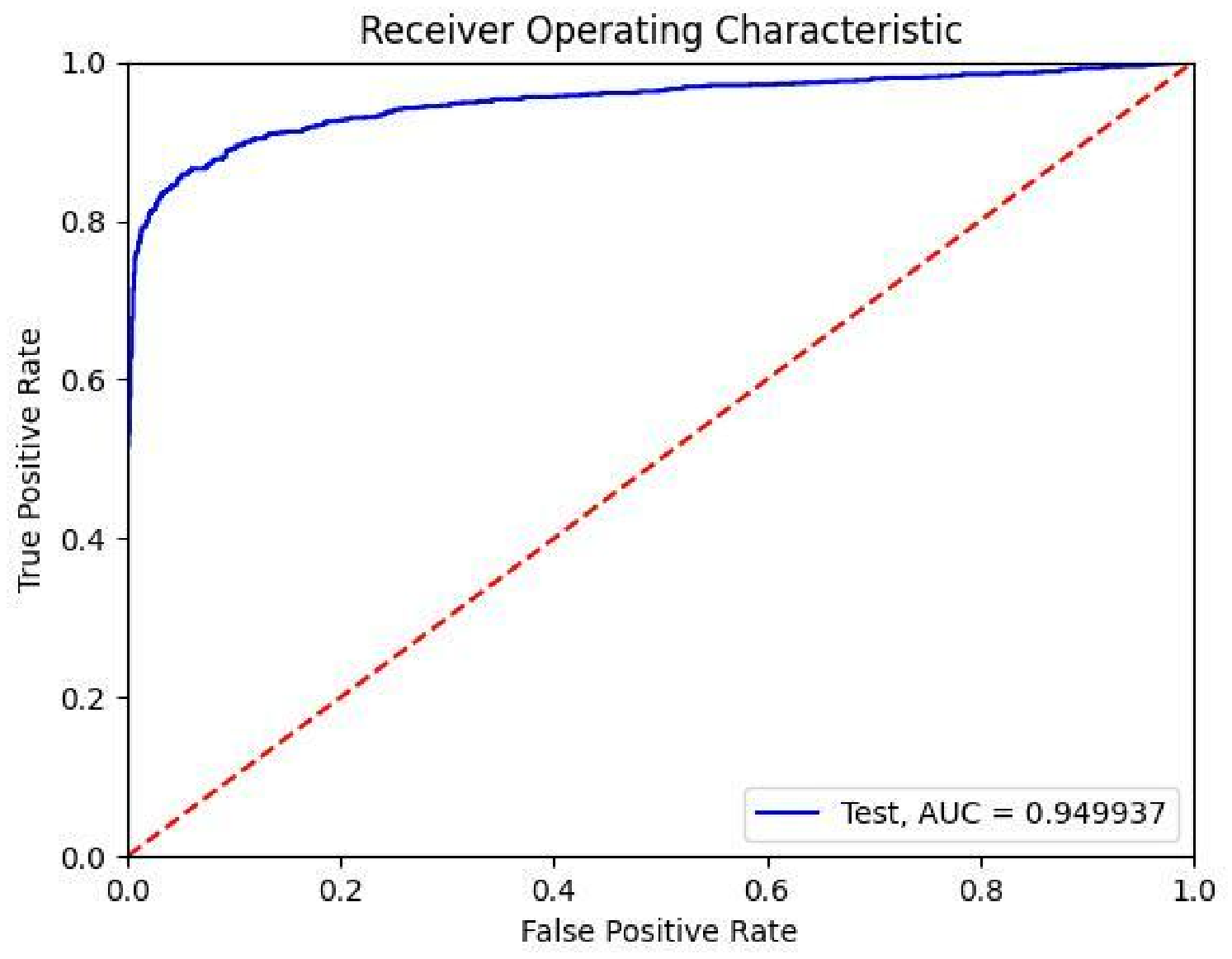}
    \caption{\centering{Test ROC curve of similarity network (ArcFace loss in ResNet-100) on the tailored verification task. }}
    \label{fig:fig33}
    \includegraphics[width=\columnwidth]{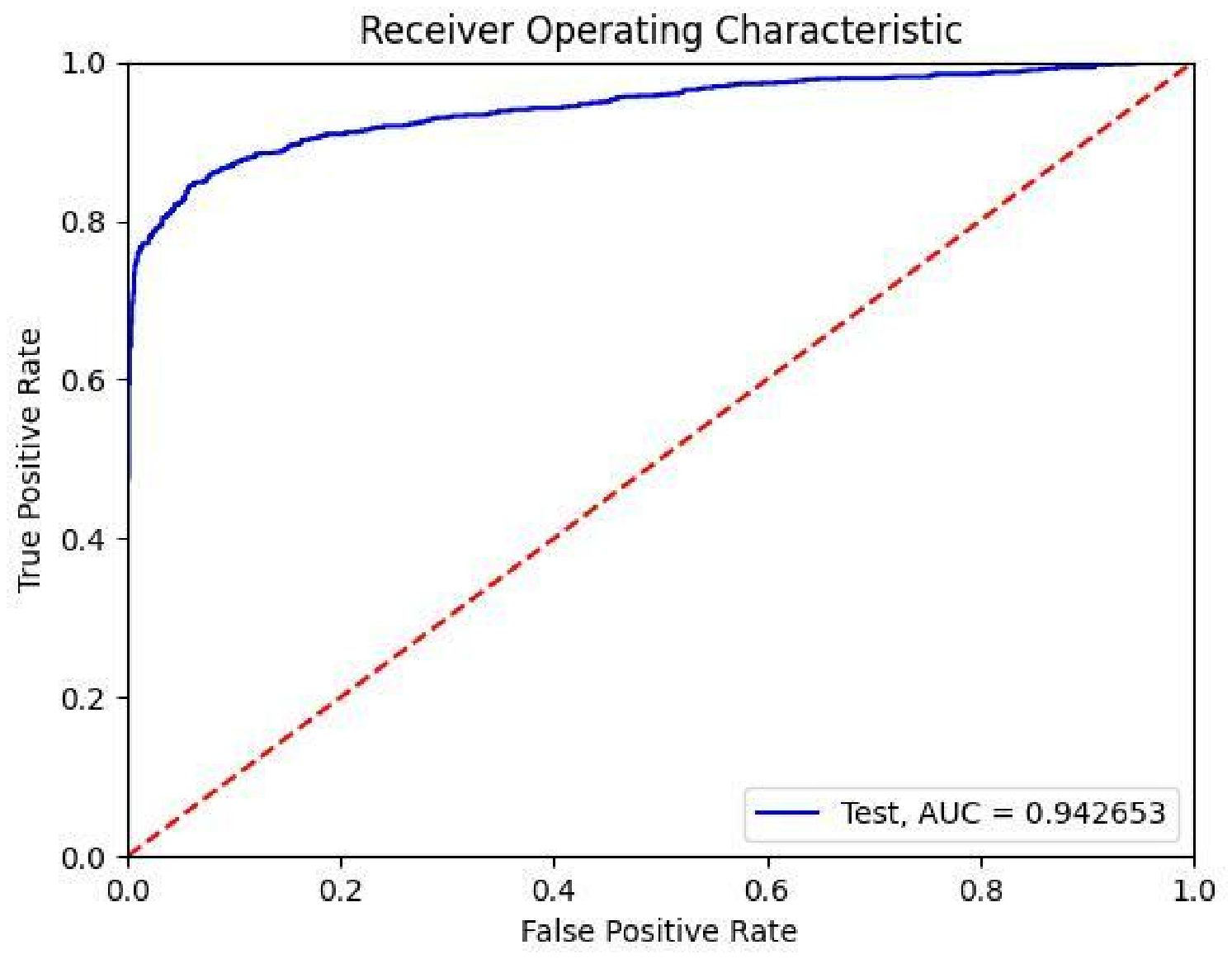}
    \caption{\centering{Test ROC curve of similarity network (ElasticFace loss in ResNet-100) on the tailored verification task.} }
    \label{fig:fig34}

    \includegraphics[width=\columnwidth]{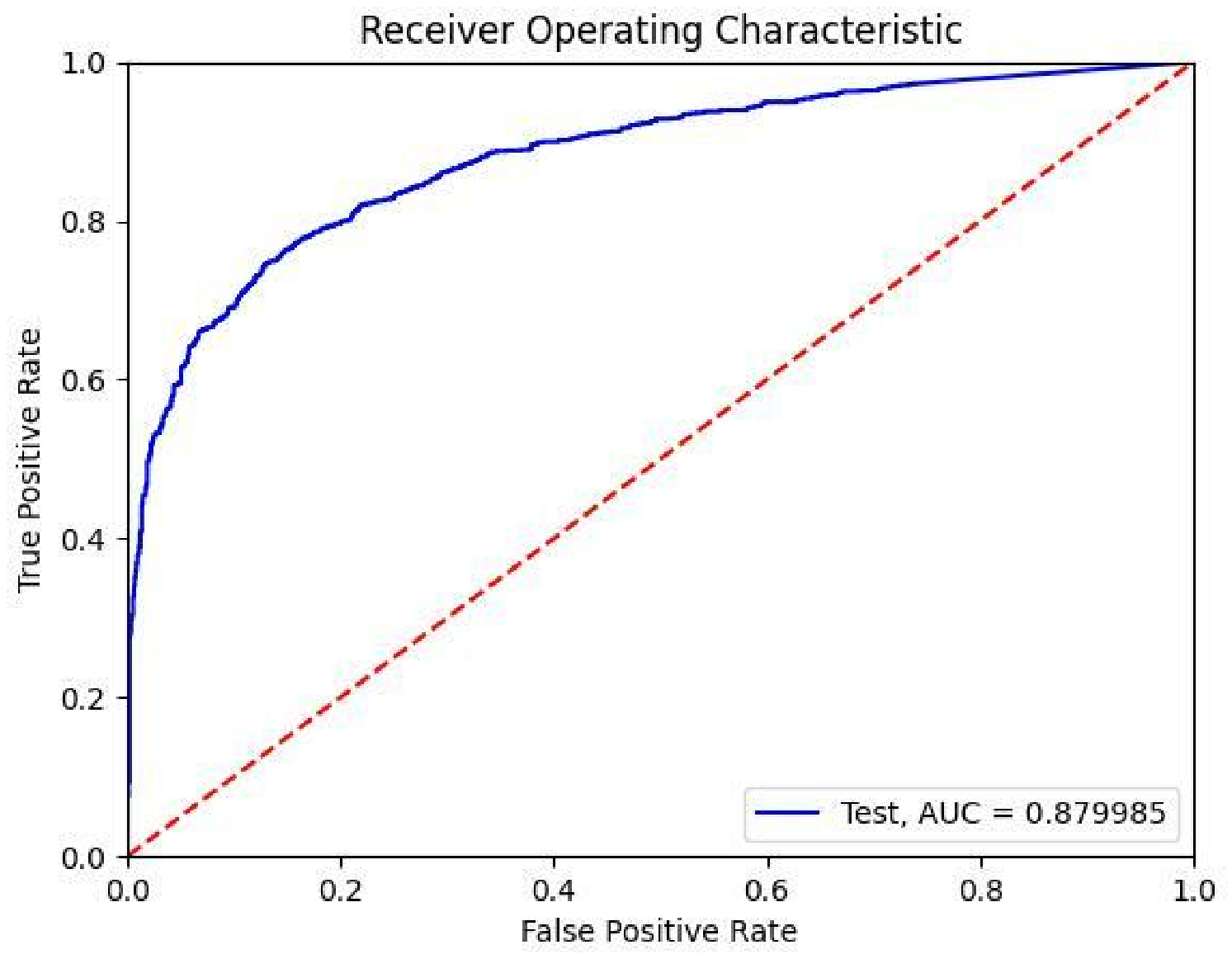}
    \caption{\centering{Test ROC curve of similarity network (MagFace loss in Mobile FaceNet) on the tailored verification task. }}
    \label{fig:fig35}
    \end{multicols}
    
\end{figure}
\begin{figure}
\centering
    \includegraphics[width=0.7\columnwidth]{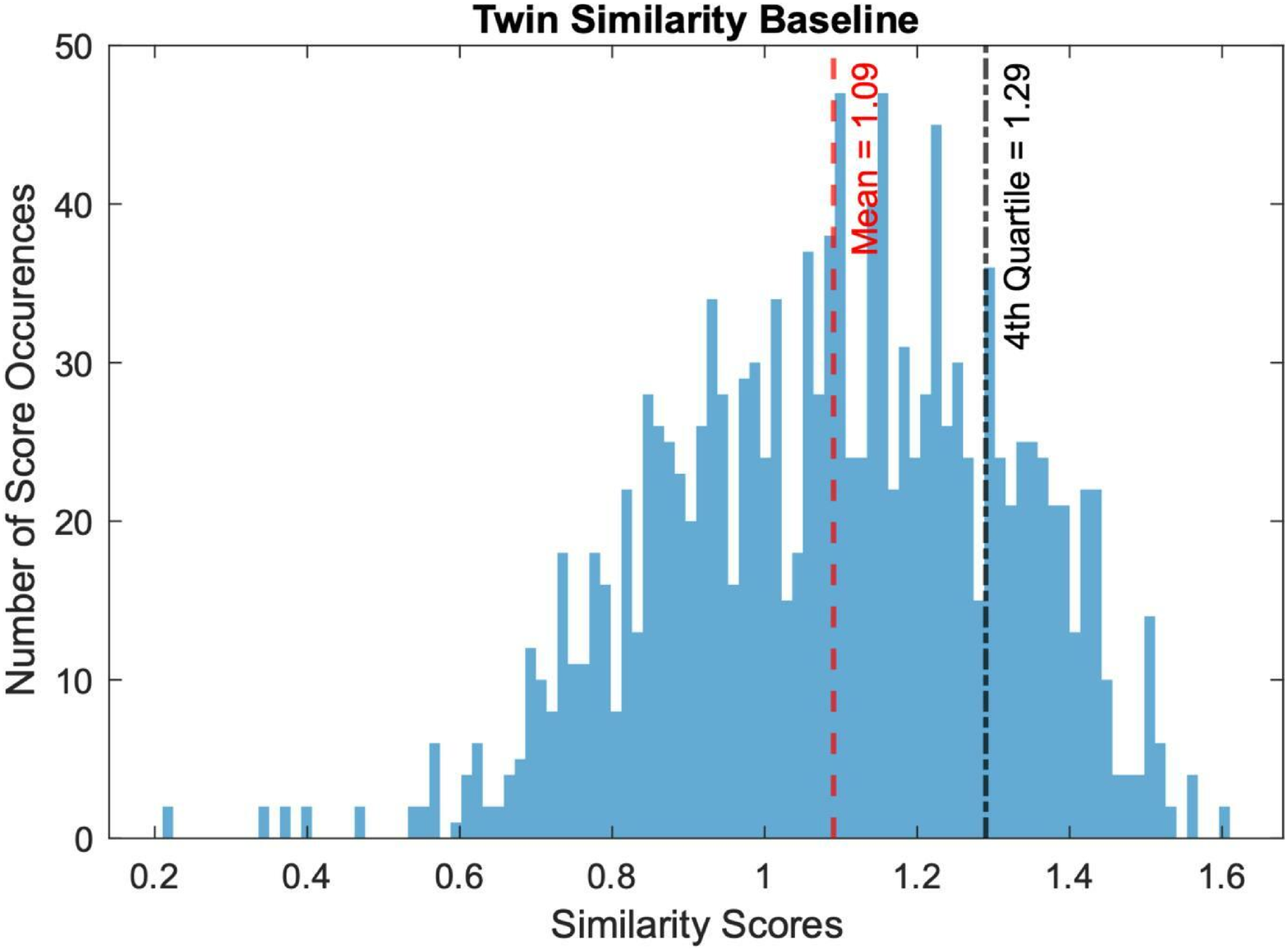}
    \caption{\centering{Twin similarity score baseline experiment. }}
    \label{fig:fig36}
    \includegraphics[width=\columnwidth]{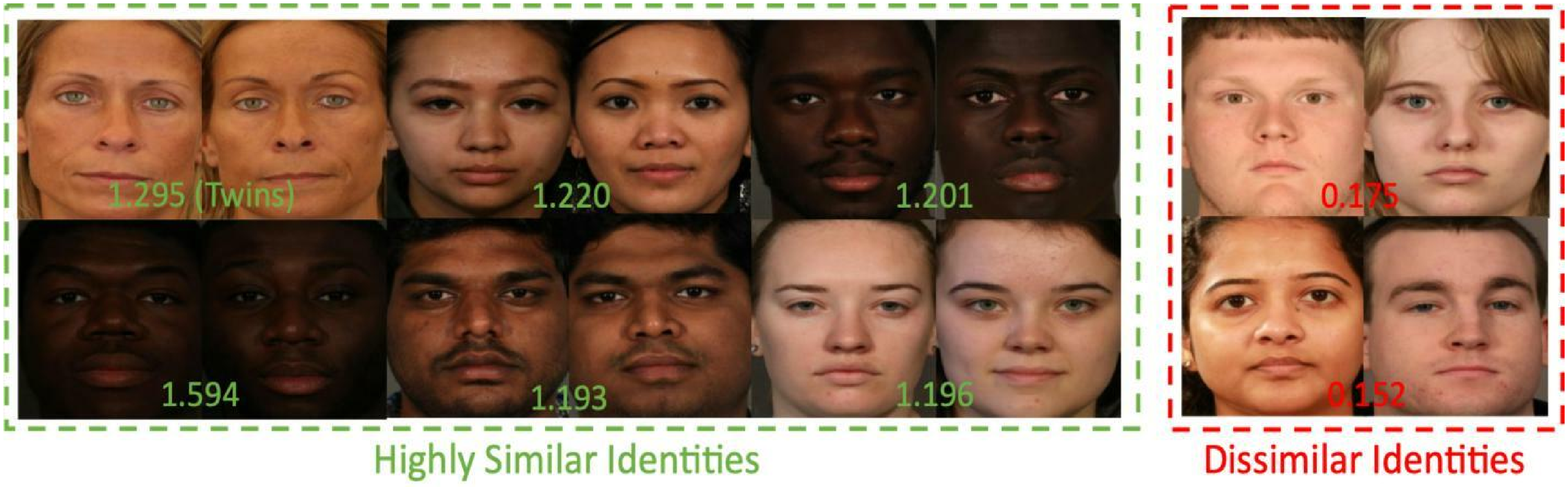}
    \caption{\centering{Examples of highly similar and dissimilar face pairs as determined by the proposed similarity network.}}
    \label{fig:fig37}
    \includegraphics[width=\columnwidth]{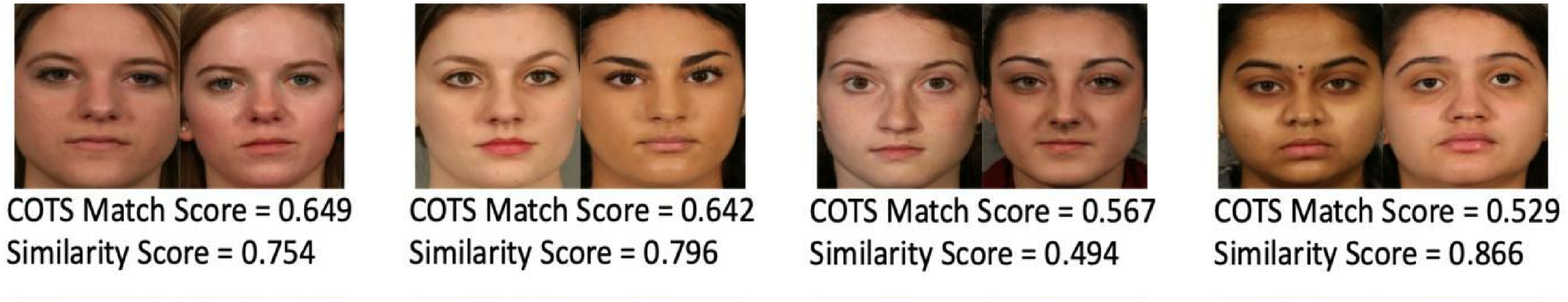}
    \caption{\centering{High COTS match score face pairs and the corresponding similarity scores for each pair.}}
    \label{fig:fig38}
\end{figure}
\begin{figure}
    \centering
    \includegraphics[width=\columnwidth]{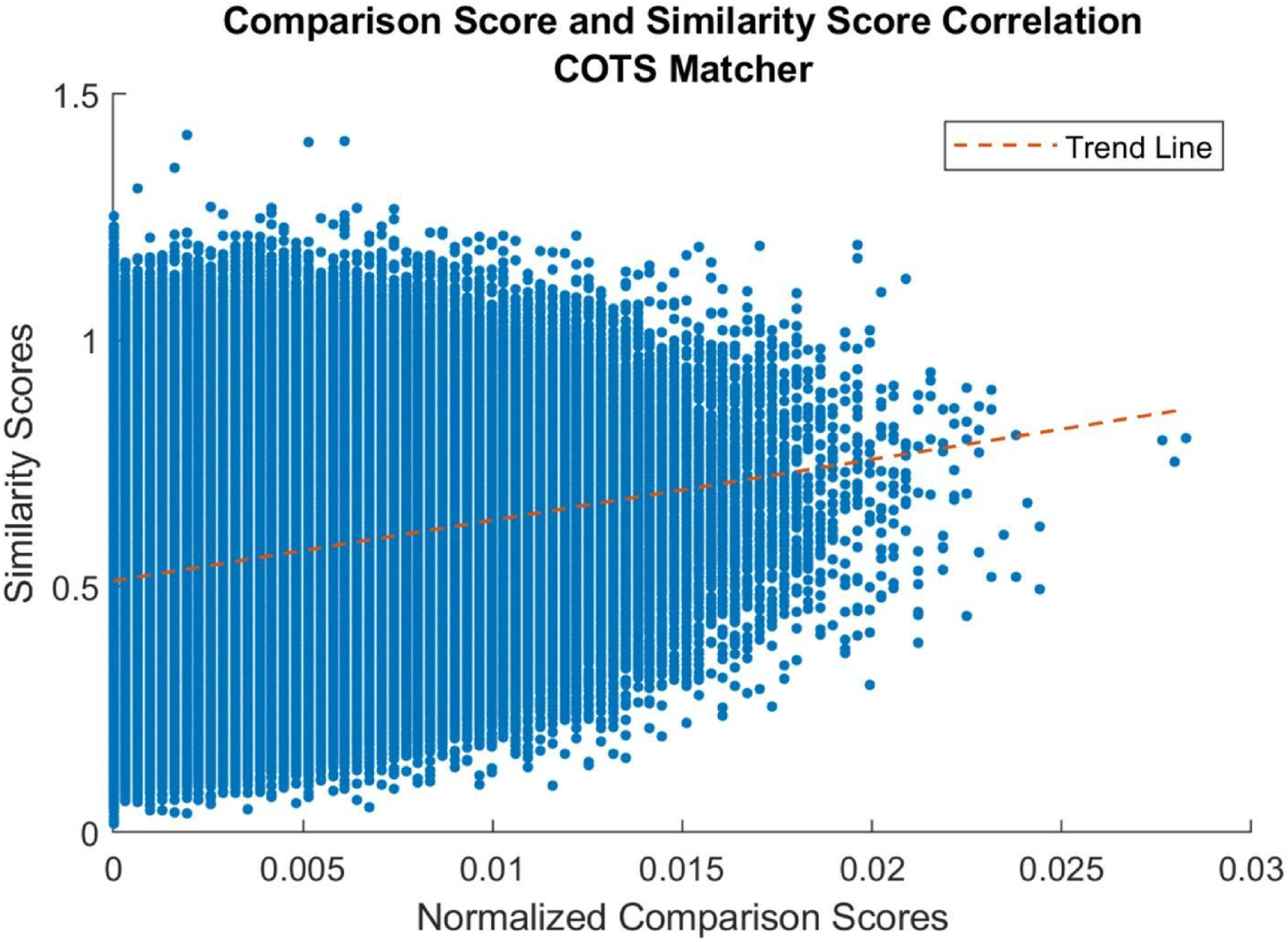}
    \caption{\centering{Correlation scatter plot of comparison scores and similarity scores, COTS matcher.}}
    \label{fig:fig39}
    \includegraphics[width=\columnwidth]{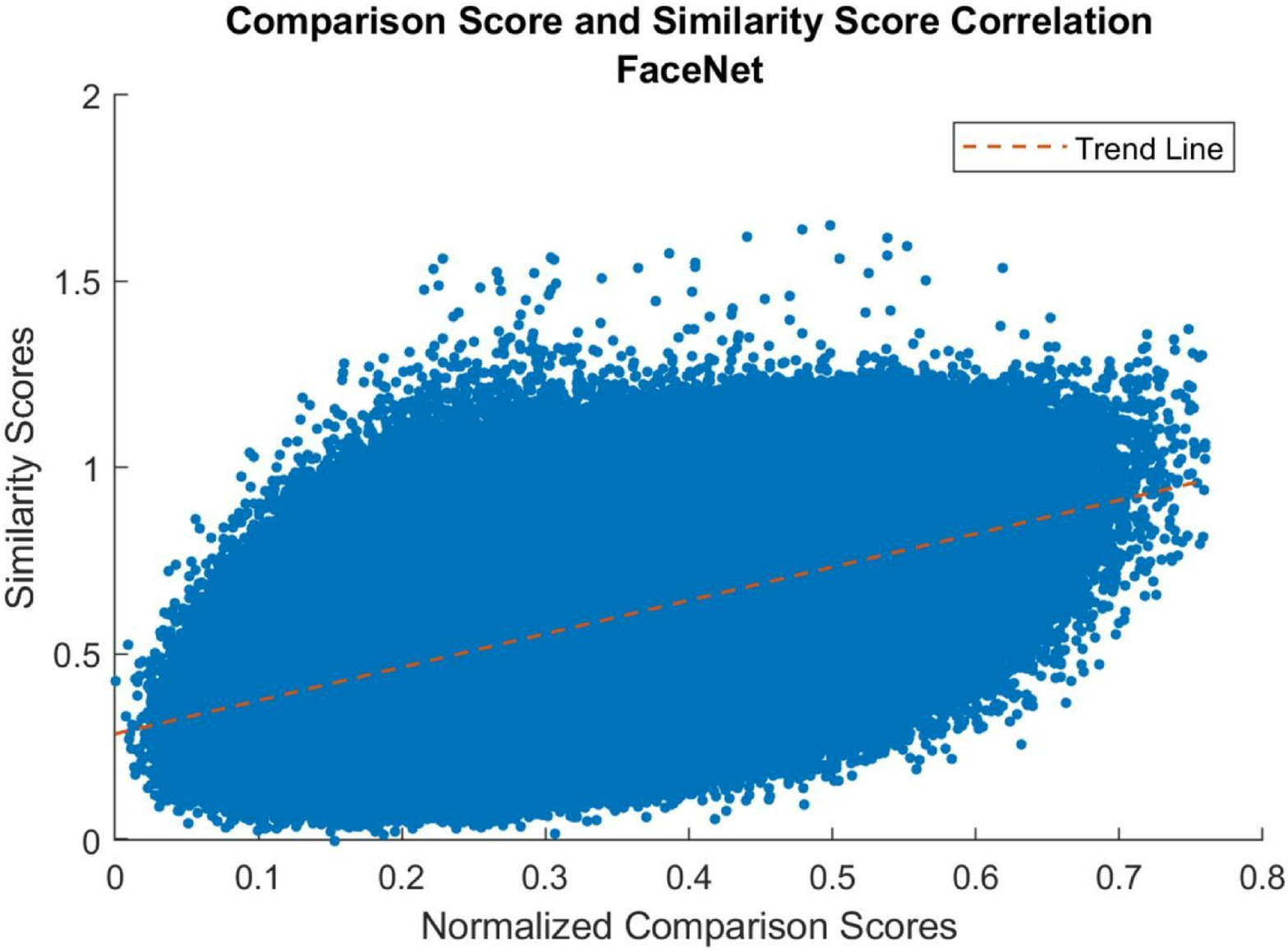}
    \caption{Correlation scatter plot of comparison scores and similarity scores, FaceNet matcher. }
    \label{fig:fig40}
\end{figure}

\begin{figure}

    \includegraphics[width=\columnwidth]{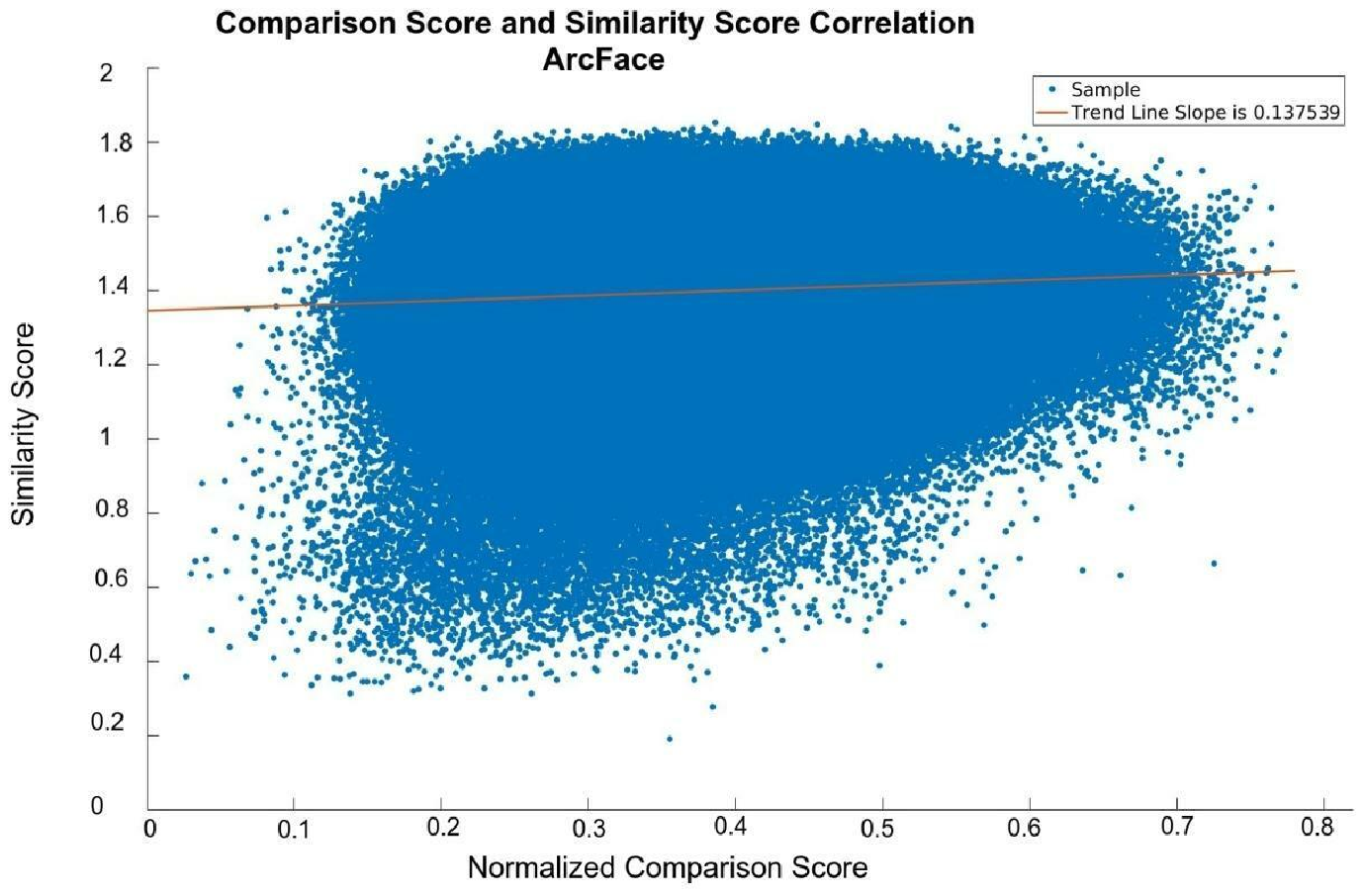}
    \caption{Correlation scatter plot of comparison scores and similarity scores, ArcFace (ResNet-100) matcher. }
    \label{fig:fig41}
    \includegraphics[width=\columnwidth]{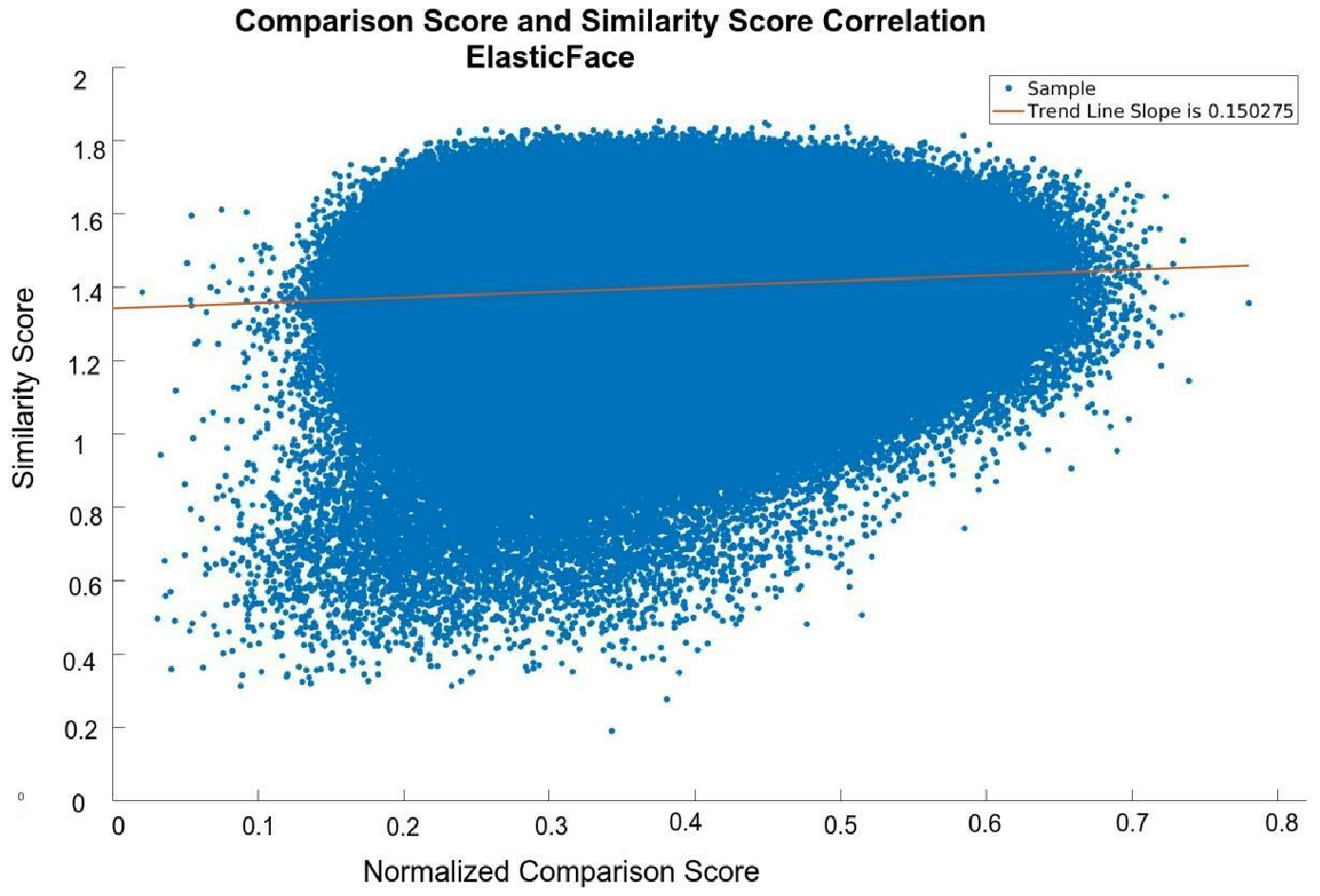}
    \caption{Correlation scatter plot of comparison scores and similarity scores, ElasticFace (ResNet-100) matcher.}
    \label{fig:fig42}
\end{figure}
\begin{figure}
    \centering
    \includegraphics[width=\columnwidth]{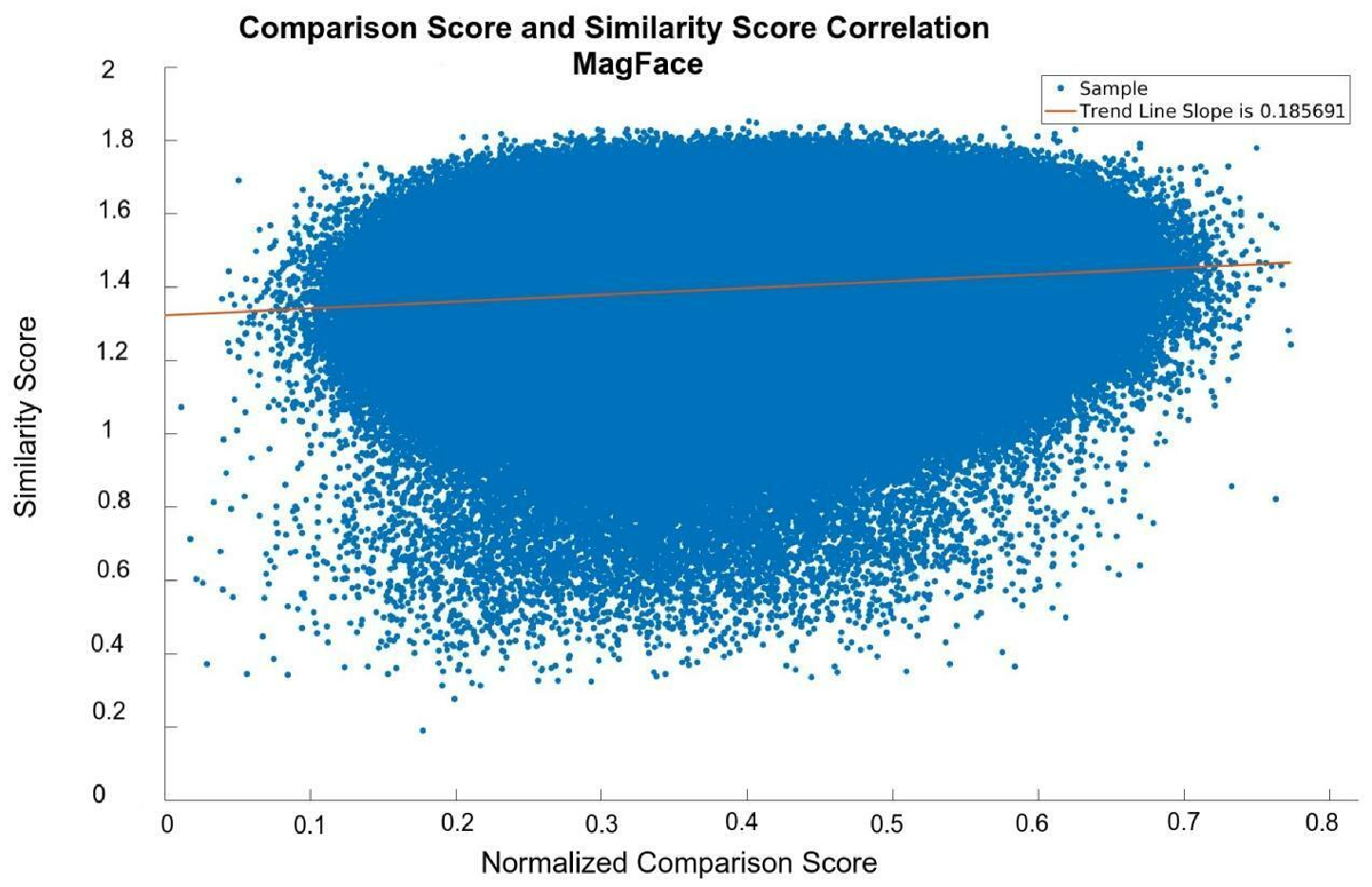}
    \caption{Correlation scatter plot of comparison scores and similarity scores, MagFace (Mobile FaceNet) matcher. }
    \label{fig:fig43}
    \includegraphics[width=\columnwidth]{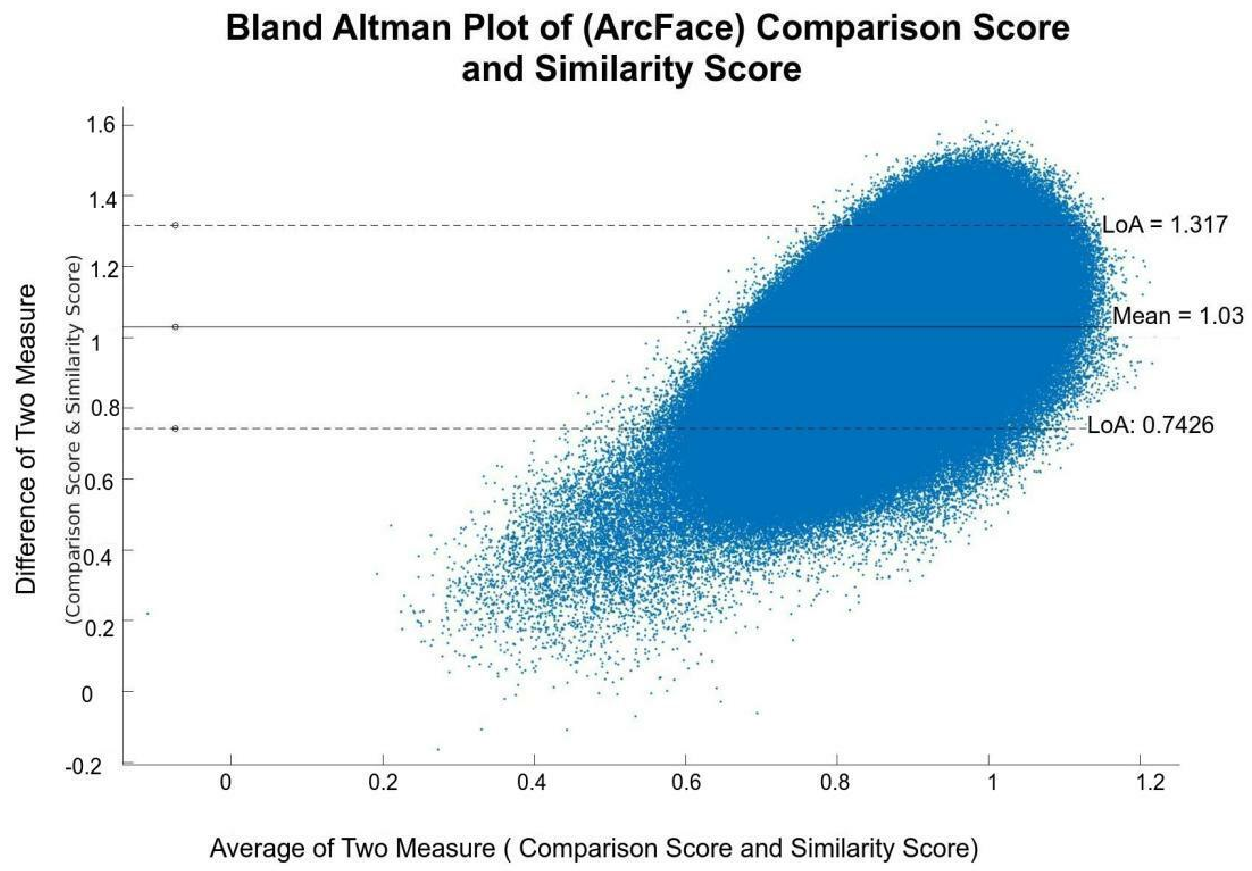}
    \caption{ Bland- Altman plot of comparison scores and similarity scores, ArcFace matcher. }
    \label{fig:fig44}
\end{figure}
\begin{figure}
    \centering
    \includegraphics[width=\columnwidth]{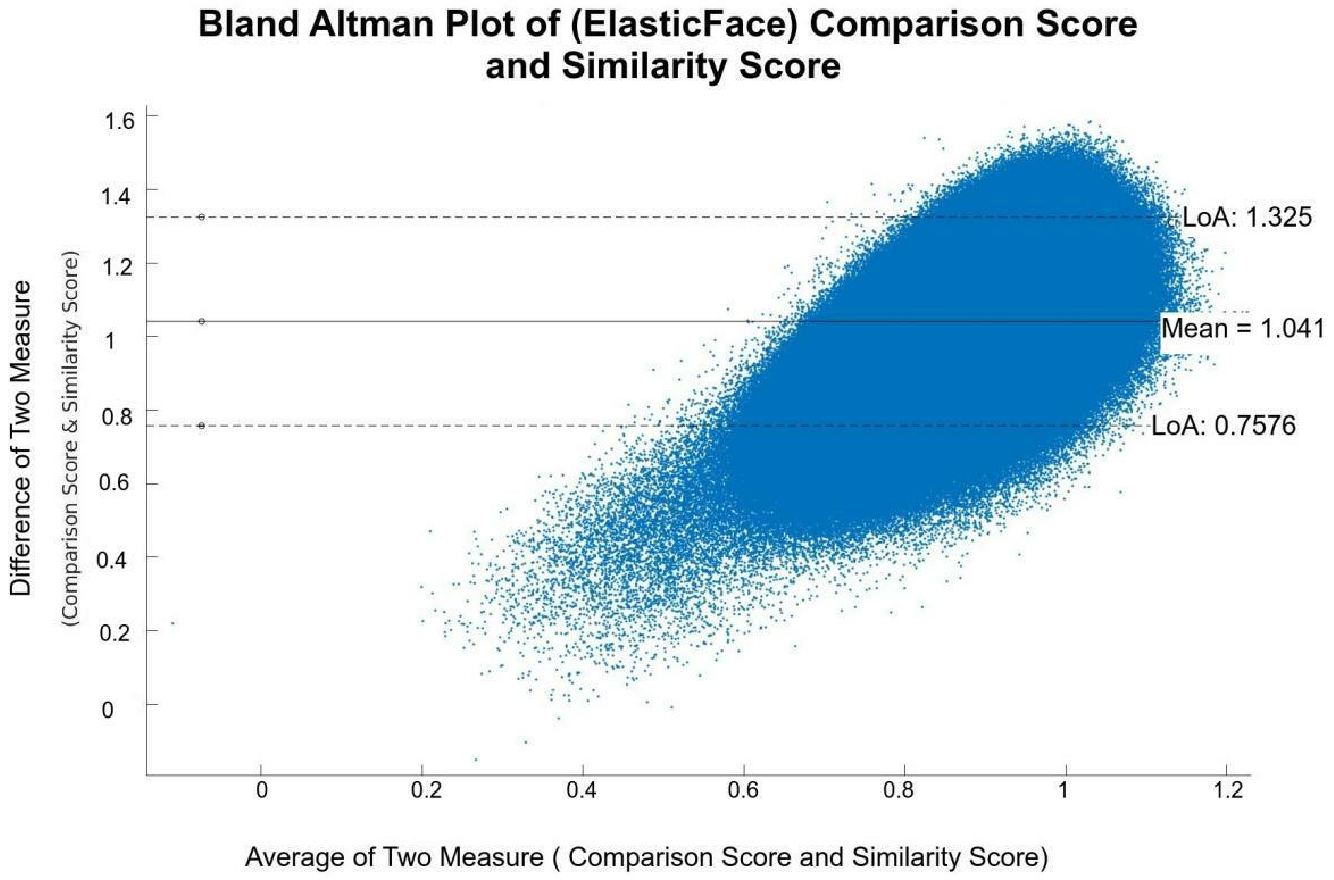}
    \caption{Bland- Altman plot of comparison scores and similarity scores, ElasticFace matcher.}
    \label{fig:fig45}
    \includegraphics[width=\columnwidth]{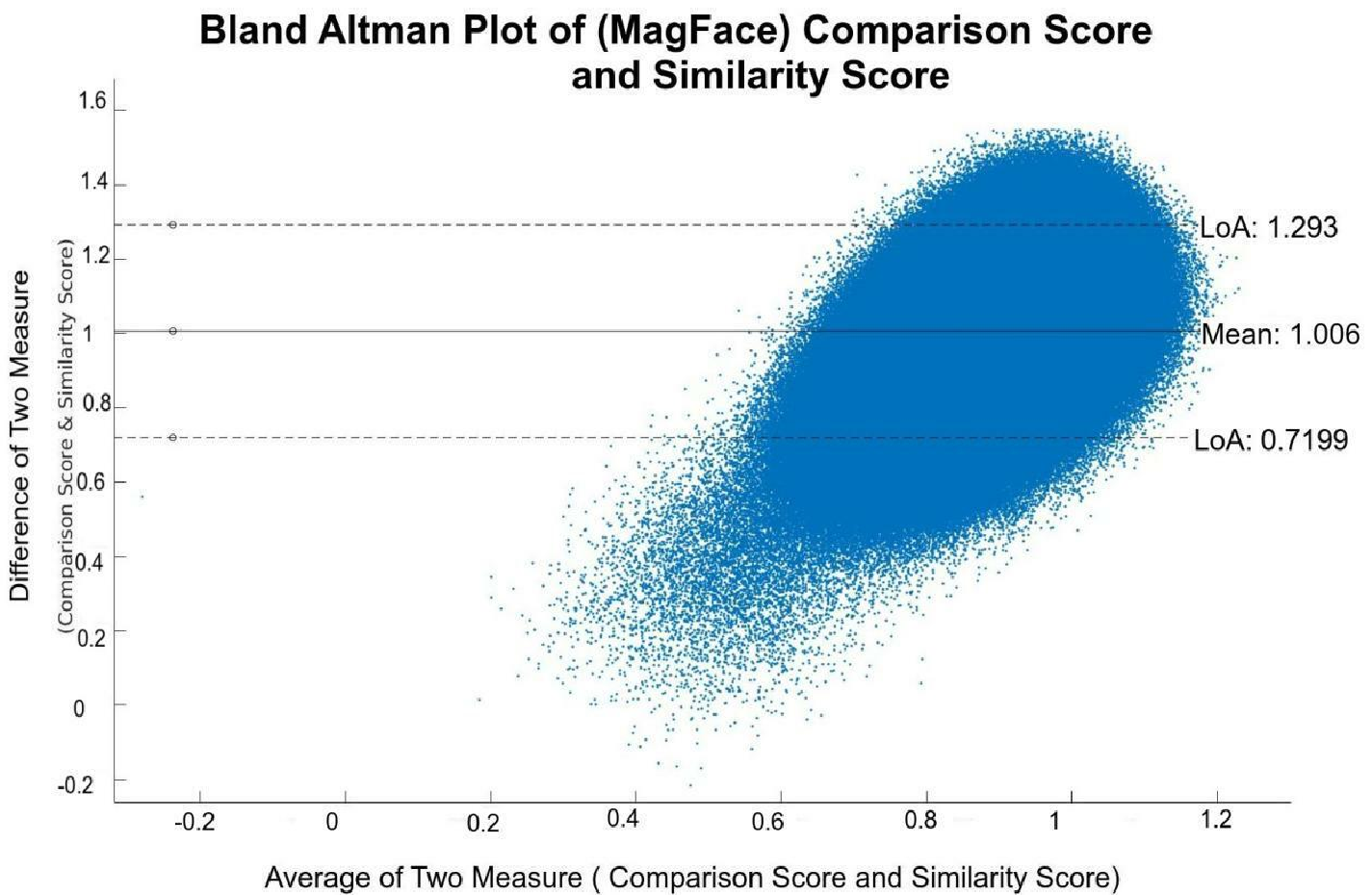}
    \caption{Bland- Altman plot of comparison scores and similarity scores, MagFace matcher. }
    \label{fig:fig46}
\end{figure}
\begin{figure}
    \centering
    \includegraphics[width=\columnwidth]{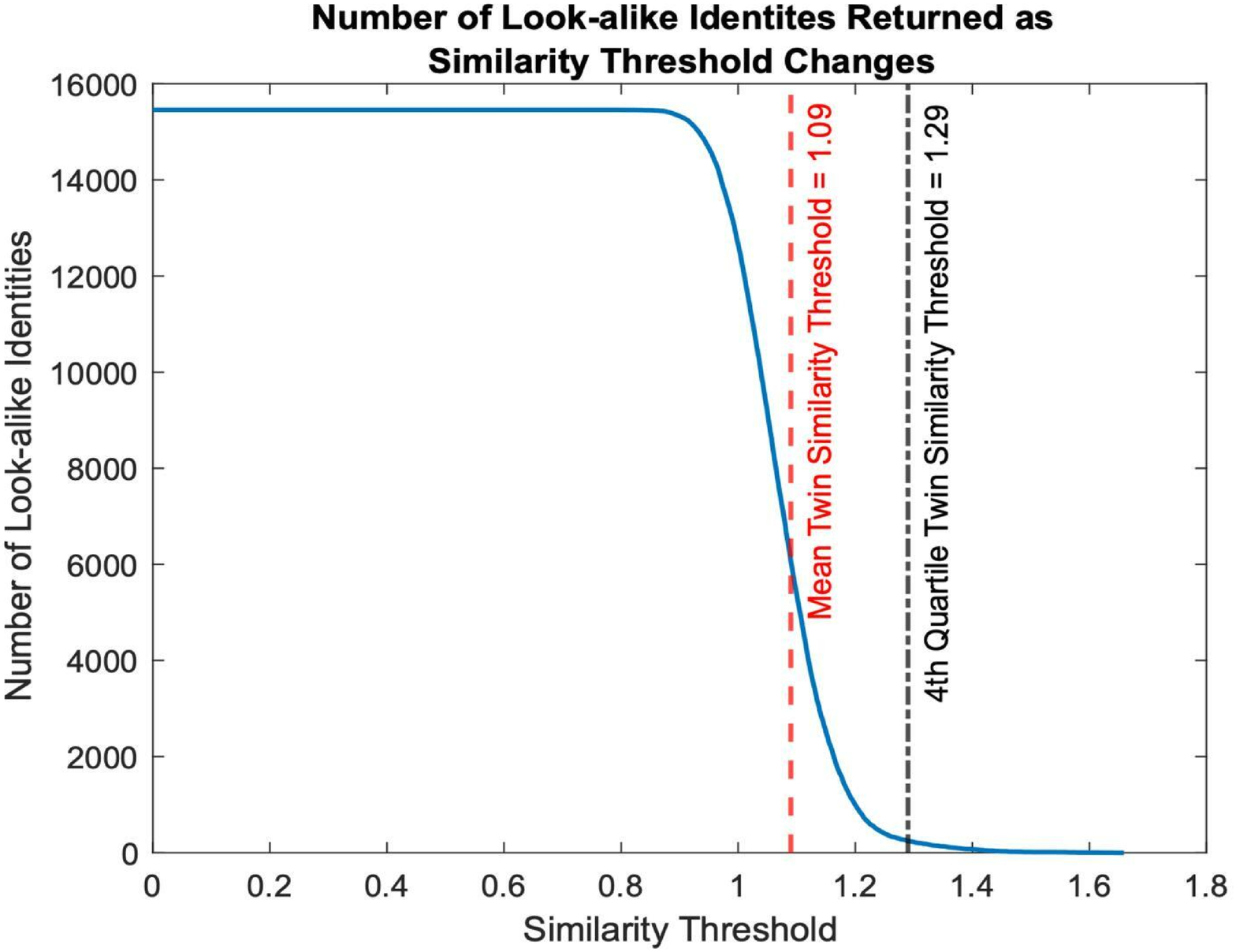}
    \caption{\centering{Number of look-alike identities returned from the large scale non-twin dataset based on similarity threshold used.}}
    \label{fig:fig47}
\end{figure}

\section{Results }

In this section, we determine different thresholds and verification performance for different similarity networks. In addition, the distribution of normalized comparison scores and the correlation between comparison scores and similarity scores are depicted here. We conclude this section with discussion on the results and limitations of the proposed framework.

\section{Match Experiment Results }

\subsection{Identical Twin Baseline Experiments }

Figures \ref{fig:fig12}\ref{fig:fig13}\ref{fig:fig14}\ref{fig:fig15}\ref{fig:fig16}illustrate the results of the identical twin baseline experiments, indicating that the average comparison score for identical twins trends higher than the comparison score for non-twin matches.  The mean comparison score for identical twins in this baseline represents the experimental twin threshold T, and, when compared to the mated score distribution, this threshold approximates the left tail of the mated distribution for both matchers tested. This result indicates that for both matchers the mean non-mated comparison score for identical twins is a good estimator of the hardest cases presented to these matchers.

\subsection{All-to-all Non-mated Matching Experiments }

After the baseline matching experiments were performed, all-to-all impostor or non-mated matching was performed using both matchers. These experiments were performed on each of the three datasets: the twin dataset, the non-twin dataset, and the large scale non-twin dataset. The comparison scores for each of these experiments were analyzed to extract the scores falling at and above the experimental twin threshold, T.

\subsection{Twin Dataset Match Experiment Results }

Figures \ref{fig:fig17}\ref{fig:fig18}\ref{fig:fig19}\ref{fig:fig20}\ref{fig:fig21} show the non-mated or impostor distributions from the twin dataset matching experiment. In this experiment “all-to-all” matching is performed using every identity in the dataset. The distributions for this experiment show that, as expected, the overwhelming majority of match scores fall below the twin threshold for both matchers. Additionally, most of the scores falling above the threshold are comparisons between two identical twins or two family members for both matchers tested.

\subsection{Non-twin Dataset Match Experiment Results }

For the non-twin dataset, as in the twin dataset experiment, all-to-all non-mated or impostor matching was performed. Similarly, the non-twin dataset experiment shows a large majority of the scores falling below the experimental threshold, as illustrated in Figures \ref{fig:fig22}\ref{fig:fig23}\ref{fig:fig24}\ref{fig:fig25}\ref{fig:fig26}. As the identities in the non-twin datasets are unrelated individuals, the scores above the threshold represent potential look-alike pairs.

\subsection{Large Scale Non-Twin Dataset Match Experiment Results }

Due to performance limitations of the COTS matcher, the final experiment was carried out using the FaceNet, ArcFace, ElasticFace and MagFace-based matcher, with results presented in Figure \ref{fig:fig27}\ref{fig:fig28}\ref{fig:fig29}\ref{fig:fig30}. Again, in the large-scale non-twin experiment, the large majority of match scores fall below the twin threshold, but the scores above the threshold represent potential look-alike pairs.

\subsection{Match Scores Above Experimental Threshold T}

For each of the previous match experiments, the matches with scores above the chosen threshold were extracted for further analysis. Tables \ref{tab:7}, \ref{tab:8} and \ref{tab:9}present the information about match scores falling above the twin threshold, T, for each matcher and each dataset.

\begin{tabularx}{\linewidth}{ c  c  c c c }
   
    \caption{\centering{Twin dataset non-mated match experiment results, comparison scores above experimental twin threshold T.}} \label{tab:7}\\
    
    \\\hline

    Relationship&

    \# Scores $$>= T$$ &

    Average Score &

    Score Range &

    Percent of Matches \\
    \\\hline	
    \\Twin Dataset COTS Matcher \\
    \hline
    
    Ident. Twin &

    601 &

    0.01865&

    [0.0129-0.0431] &

    0.0117\% \\
    
    Ident. Mirror Twin&

    112 &

    0.02&

    [0.0132-0.955] &

    0.0022\% \\
    
    Fraternal Twin&

    48&

    0.0199 &

    [0.0135-0.799]&

    0.00093\% \\
    
    Mother&

    1 &

    0.0158&

    [0.0158]&

    0.000019\% \\
    
    Child &

    1 &

    0.0158&

    [0.0158]&

    0.000019\% \\
    
    No Relation&

    199 &

    0.0137 &

    [0.0129-0.0189] &

    0.0038\% \\
    
    Total &

    962 &&&0.0187\%\\

     \\\hline
     Relationship &

    \#Scores >= T&

    Average Score &

    Score Range&

    Percent of Matches \\

    \\\hline
    \\Twin Dataset FaceNet Matcher \\
    \\\hline

    Ident. Twin &

    724 &

    0.746&

    [0.6905-0.856]&

    0.014\% \\
    
    Ident. Mirror Twin &

    144 &

    0.746 &

    [0.6905-0.84] &

    0.00279\% \\
    
    Fraternal Twin&

    50 &

    0.753&

    [0.6905-0.83]&

    0.00097\% \\
    
    No Relation &

    4 &

    0.706 &

    [0.694-0.7177] &

    0.000077\%\\ 
    
    Unknown &

    2 &

    0.694 &

    [0.694]&

    0.000038\% \\
    
    Total &

    924 &&&

    0.018\% \\
    \\\hline

\end{tabularx}

\begin{tabularx}{\linewidth}{ c  c  c c c }
    \caption{\centering{Non-twin dataset non-mated match experiment results, comparison scores above experimental twin threshold T. }}\label{tab:8}\\
    \hline

    Relationship&

    \# Scores >= T &

    Average Score&

    Score Range &

    Percent of Matches \\
    \hline
    Non-twin dataset – COTS Matcher \\
    \hline

    No Relation &

    16274&

    0.0144&

    [0.0129-0.0283]&

    0.0580\% \\
    
    Non-twin dataset – FaceNet Matcher \\
    \hline
    Relationship &

    \# Scores >= T &

    Average Score &

    Score Range &

    Percent of Matches\\ 
    \hline
    
    No Relation &

    97&

    0.704&

    [0.6905-0.76] &

    0.000346\% \\
    \hline

\end{tabularx}

\begin{tabularx}{\linewidth}{ c  c  c c c }
    \caption{\centering{Large scale non-twin dataset non-mated match experiment results, comparison scores above experimental twin threshold T.}}\label{tab:9}\\
    \hline

    Relationship&

    \# Scores $$>= T $$&

    Average Score&

    Score Range &

    Percent of Matches\\
    \hline
    
    Large Scale Non-twin dataset – FaceNet Matcher\\

    No Relation &

    792&

    0.71 &

    [0.6905-0.76] &

    0.000331\% \\
    \hline

\end{tabularx}

In each of the matching experiments, it is shown that an overwhelming majority of match scores fall below the experimental twin threshold. This leads to the observation that impostor look-alikes are a rare occurrence in the population used in this study. Due to the relatively small number of identities in the datasets used for this evaluation, it is not possible to accurately predict the frequency of look-alike occurrence in the general population based on these results. However, the similarity measure described in the next section provides a method of finding highly similar faces in any given dataset. 
\section{Similarity Network Results }

\subsection{Network Training and Testing }
After training and testing, the proposed network achieved a train AUC of 0.917, and test AUC of 0.979 in the classification of a pair of face images as a twin pair or look-alike pair (Figures \ref{fig:fig31} \ref{fig:fig32}). ArcFace, ElasticFace and MagFace based similarity network AUC, EER and FNMR are provided in Table \ref{tab:10}. In this table, we observe that FaceNet-based verification performs better than other three networks. Moreover, the ROC curve of ArcFace, ElasticFace and MagFace based networks are presented in Figures \ref{fig:fig33} \ref{fig:fig34} \ref{fig:fig35}. While the end goal of this network is not verification, the accuracy of the network on the tailored verification task shows that the network can accurately identify highly similar face pairs. 
\begin{tabularx}{\linewidth}{  |X | c |c |c |}
    \caption{\centering{Performance comparison of different similarity networks.}}\label{tab:10}\\
    \hline

    Algorithm&

    FNMR@FMR (1e-3) &

    EER &

    AUC \\    \hline

    FaceNet (Train on Twin Dataset) &

    0.5385&

    0.0798&

    0.9799 \\    \hline

    ArcFace (ResNet-100 backbone) &

    0.7415&

    0.1032 &

    0.9499 \\    \hline

    ElasticFace (ResNet-100 backbone) &

    0.7622 &

    0.1199 &

    0.9427 \\    \hline

    MagFace (Mobile FaceNet backbone)&

    0.9553&

    0.1914 &

    0.8799 \\    \hline

\end{tabularx}

\subsection{Identical Twin Similarity Baseline }

This similarity network was then applied to both the twin dataset and large-scale non-twin dataset to observe the general similarity of twin and non-twin individuals. Initially, the similarity score of only the identical twin pairs was calculated (Figure \ref{fig:fig36}). This distribution of similarity scores for identical twin pairs is the foundation of the worst-case baseline measure of similarity. As identical twins exist on a spectrum of similarity, two measurements of the baseline similarity between identical twin pairs are reported. The mean similarity score between identical twin pairs, 1.09, captures the similarity of both highly similar and dissimilar twins, while the fourth quartile of the similarity score distribution, $ \geq 1.29$, represents only the most similar twin pairs. In this experiment, the fourth quartile score of the distribution may more accurately represent the worst case of similarity presented to FR systems.

\subsection{Look-alike Pair Extraction }

After determining the worst-case baseline for facial similarity, this measure was used to set the threshold for the similarity scores of the large-scale non-twin dataset. Since the network was fine-tuned using only ideal face images, the similarity score returned for “in-the-wild” face images may not be as robust as the similarity score returned for controlled images. Several examples of identical twin pairs and non-mated pairs with similarity scores exceeding the baseline measurements are shown in Figure \ref{fig:fig37}.

\subsection{Comparison Score Versus Similarity Score Analysis }

An additional analysis was performed to correlate the comparison score results of the COTS matcher to the similarity score obtained from our similarity network. Using the non-mated pairs whose comparison scores exceeded the experimental twin threshold in the matching experiments detailed above, a comparison was made to the similarity score calculated for the same pairs. Examples of individuals with high COTS comparison scores and the corresponding similarity score are highlighted below (see Figure \ref{fig:fig38}).

As Figure \ref{fig:fig38} indicates, the COTS comparison score for each of these face pairs was high; however, none of the pairs’ similarity scores are above the twin similarity threshold. This indicates that the comparison score returned by the COTS matcher may not be directly correlated with facial similarity, and instead, may rely on other features of the image in its comparison score determination process. 

A further analysis of the relationship between the comparison score returned by a facial recognition tool and the similarity score returned by the proposed network was carried out. The first experiment in this analysis shows a scatter plot on which each point represents the comparison score returned by the COTS matcher and the similarity score returned by the proposed network for each face comparison in the non-twin dataset (Figure \ref{fig:fig39}). This same experiment was carried out for the FaceNet, ArcFace, ElasticFace, and MagFace-based comparison scores and similarity scores returned for the large scale non-twin dataset (Figure  \ref{fig:fig40}\ref{fig:fig41}\ref{fig:fig42}\ref{fig:fig43}). The trend line shown on each plot was determined via a linear polynomial curve fitting algorithm and shows the overall trend of the datapoints over the range of comparison scores. Moreover, ArcFace, ElasticFace, and MagFace-based comparison scores and similarity scores generated for the large-scale non-twin dataset are plotted in Bland-Altman diagrams (Figure \ref{fig:fig44}\ref{fig:fig45}\ref{fig:fig46}). These figures depict the correlation between comparison scores and similarity scores, since similarity scores is derived from the FaceNet (Inception-ResNet v1)-based network, and comparison scores come from ResNet-100 (ArcFace, ElasticFace) or Mobile FaceNet (MagFace) networks.

As shown in each of the above experiments, the COTS, FaceNet, ArcFace, ElasticFace, and MagFace matcher comparison scores show a positive trend with the similarity score returned by the proposed network for each of the tested face pairs. However, it is shown that for both tested matchers there is a wide range of similarity scores returned over the entire range of comparison scores for both matchers. These results indicate that while perceived facial similarity does play a role in the determination of the comparison score between face pairs, the facial similarity of the pair may not be the chief factor in the determination of comparison score. This is the expected result, as facial recognition is a distinct challenge from that of determining facial similarity, and the comparison score returned by a facial recognition tool is generally tuned to maximize recognition performance. This is especially true for the COTS matcher used in this work, as this matcher utilizes face template comparisons which may focus on features that do not correspond directly to perceived facial similarity. This is also true for the FaceNet matcher, because while the authors claim that the network’s face embeddings are directly correlated with facial similarity, the network is optimized for maximal facial recognition performance. This may lead the network to produce embeddings which are designed for facial recognition comparisons, and not the determination of facial similarity. 
\subsection{Frequency of Look-alike Identity Occurrence Estimation }

Finally, an investigation into the number of potential look-alike pairs returned by the network while varying the similarity threshold was performed to further understand the occurrence of look-alikes in a given population of unrelated individuals (Figure \ref{fig:fig47}).

Given the mean twin similarity threshold of 1.09, 6,153 of the total 15,455 identities in the large-scale non-twin dataset have at least one similarity comparison at or above the threshold. This means 39.8\% of the identities have one or more potential look-alike at this level of similarity. At and above the fourth quartile threshold, only 228 identities have one or more potential look-alike, or 1.475\% of identities in the dataset. While there are not enough identities present in the datasets used in this work to estimate the frequency of look-alike occurrence in general using the proposed network, given a large enough dataset such an analysis could be carried out. Furthermore, the proposed network can be applied to any dataset to gauge the number of look-alike identities within the dataset by applying the worst-case similarity threshold to an all-to-all non-mated similarity comparison for the dataset. 

\section{Discussion} 

The results shown in the previous section lead to several interesting conclusions. First, the results of the identical twin match experimentation further highlight the difficulty of identical twin pairs when presented to facial recognition tools. This conclusion is drawn from the fact that the non-mated distribution of identical twin matches lies close to the mated distribution and is in fact a good estimator of the left tail of the mated distribution for both tested matchers. Second, the occurrence of non-mated look-alikes in the tested datasets is quite rare. This is drawn from the low occurrence of scores falling above the twin comparison score threshold T, and the small percentage of identities having at least one look-alike as determined by the worst-case similarity baseline measurement shown in Figure \ref{fig:fig47}. Finally, the results shown in Figures \ref{fig:fig39} and \ref{fig:fig40}lead to the conclusion that the determination of a comparison score from a facial recognition tool may not be directly correlated with perceived facial similarity. It is shown that the comparison score returned from each of the matchers does show a positive trend with the similarity score returned by the proposed network, but that the similarity of the faces in question may not be the chief determining factor of the comparison score. It is hypothesized that the comparison score returned by a facial recognition tool is instead optimized for peak recognition performance, and that the determination of a quantitative measure of facial similarity is a distinct task from that of facial recognition. 

The main limitation of this work is the lack of diversity in our Twin and non-Twin datasets.  The twin dataset mostly consists of Caucasians subjects (84.97\%), while the non-Twin dataset consists of Caucasians subjects (71.78\%) and age group 20-29 (60.30\%). This lack of diversity limits the generalization capability of the proposed framework. To overcome this shortcoming, we aim to consider a more diverse non-Twin dataset and collect samples from non-Caucasian twins in the future works.   

\section{Conclusions and Future Work}
\subsection{Conclusion}
This work explores the problem of facial similarity and the facial recognition of identical twins and look-alike or doppelganger pairs. These face pairs represent the hardest cases presented to facial recognition systems and examining the facial similarity of these pairs allows for further exploration into the relationship between facial similarity and the comparison score returned by a facial recognition system. The primary goal of this work is to quantify the facial similarity of identical twins in general. This similarity measurement represents the worst-case baseline of facial similarity in any non-mated comparison in facial recognition. This measure was developed via a deep convolutional neural network, trained specifically to identify highly similar face pairs. This network returns a quantitative measure of facial similarity (i.e., a similarity score) for any two faces, and has many applications including identifying look-alike pairs from large scale face datasets. Additionally, match experimentation was carried out to demonstrate the effect of highly similar faces on both commercial off the shelf and academic machine learning based facial recognition tools and identify potential look-alike identities. Using an experimental threshold determined from the mean identical twin non-mated match score, potential look-alike pairs were extracted from the populations used in this study through match experimentation. 

Several conclusions can be drawn from the results of this work. First, look-alike pairs are quite rare in the populations contained within the datasets used for this study. This is confirmed by both the match experiments carried out using both a commercial off the shelf matcher, an academic machine learning based matcher, and the proposed similarity network. Due to the relatively low frequency of look-alikes contained within the data sets used, an accurate estimate of the frequency of look-alike pairs in general was not carried out, however the results of the proposed similarity network show that the network can identify similar face pairs from any face dataset. This result has applications in several spheres including: finding suitably similar face for morph face generation, determining the difficulty of any given dataset by the number of look-alike identities contained within, and further investigation into the relationship between comparison score and perceived similarity of any two faces. Second, the result show that the comparison score returned by a commercial off the shelf facial recognition tool are not directly correlated with the perceived visual similarity determined by the proposed network. This indicates that the commercial off the shelf facial recognition tool and the neural network-based tool used in the match experimentation in this work may not determine the comparison score between two faces directly from a measure of facial similarity, and instead determine its comparison scores via some feature set designed to maximize facial recognition performance. Also, comparison scores derived from ArcFace, ElasticFace, MagFace loss-based networks and similarity scores have little correlation. Because similarity scores and comparison scores are generated from different networks. Similarity scores come from FaceNet network and comparison scores come from COTS or FaceNet or ResNet-100 or Mobile FaceNet network.  Finally, the difficulty that both identical twin pairs and look-alike pairs pose to facial recognition systems has again been demonstrated, as it has in previous works. This conclusion is drawn from the initial match experimentation carried out on identical twin pairs, and the high comparison scores returned from non-twin look-alike pairs falling above the experimental threshold determined via the identical twin match experimentation. 

\subsection{Future Work}

There are several topics of interest for future studies in this area. The first is an adaptation of the birthday paradox to the occurrence of look-alikes in a population. Much like the birthday paradox seeks to calculate the probability of two people in a population sharing a birthday, calculating the probability of two unrelated individuals having high facial similarity based on the number of identities in a dataset would be a useful measure as face datasets continue to grow. This network would allow for the determination of such a measurement. The second is an application of the proposed similarity network in the generation of morphed faces. As the proposed network can identify similar faces from any dataset, using these look-alike pairs as suitably similar faces for morphed face pairs could further confirm the accuracy of the determination of facial similarity by the similarity network. Finally, the proposed network could be used to further investigate the role that facial similarity plays in the determination of comparison score by popular facial recognition tools. As shown in preliminary results in this work correlating the comparison score returned by two facial recognition tools and the similarity score returned by the proposed network, the comparison score returned by the facial recognition tools does not seem to be directly correlated with the similarity score returned by the proposed network. This experimentation could be extended to other commercial off the shelf facial recognition tools, or even machine learning based facial recognition tools, to better understand the role that facial similarity plays in the determination of a comparison score between two faces.

\section*{Acknowledgement}
This material is based upon work supported by the Centre for
Identification Technology Research and the National Science
Foundation under Grant No. 1650474.
\section*{Conflict of Interest}

The author declares that there is no conflict of interest that
could be perceived as prejudicing the impartiality of the
research reported.
\section*{Data Availability Statement}

The data that support the findings of this study are available
from the corresponding author upon reasonable request.

\bibliographystyle{unsrtnat}
\bibliography{template}  






\end{document}